\documentclass[final,1p,times]{elsarticle}

\usepackage{amssymb}
\usepackage{amsmath}
\usepackage{mathrsfs}
\usepackage{amsthm}
\usepackage{cancel}
\usepackage{color}
\newcommand{\add}[1]{\textcolor{black}{{#1}}}
\newcommand{\blue}[1]{\textcolor{black}{{#1}}}

\newcommand {\bp} {\bcancel{p}}
\newcommand {\bphi} {\bcancel{\phi}}
\newcommand {\bs} {\mbox{\boldmath $s$}}
\newcommand {\bt} {\mbox{\boldmath $t$}}
\newcommand {\bu} {\mbox{\boldmath $u$}}
\newcommand {\br} {\mbox{\boldmath $r$}}
\newcommand {\bz} {\mbox{\boldmath $z$}}
\newcommand {\bx} {\mbox{\boldmath $x$}}

\newcommand{\E}[1]{\mathbb{E}\left[{#1}\right]}

\newcommand {\cS} {\mathcal{S}}

\journal{Physics Reports}

\begin{document}

\begin{frontmatter}



\title{Quantifying relevance in learning and inference}

\author[ICTP]{Matteo Marsili}
\address[ICTP]{The Abdus Salam International Centre for Theoretical Physics, 34151 Trieste, Italy}
\author[NTNU]{Yasser Roudi\corref{cor1}}
\address[NTNU]{Kavli Institute for Systems Neuroscience and Centre for Neural Computation, Norwegian University of
Science and Technology (NTNU), Trondheim 7030, Norway}
\ead{yasser.roudi@ntnu.no}
\cortext[cor1]{Corresponding author.}

\begin{abstract}
Learning is a distinctive feature of intelligent behaviour. High-throughput experimental data and Big Data promise to open new windows on complex systems such as cells, the brain or our societies. Yet, the puzzling success of Artificial Intelligence and Machine Learning shows that we still have a poor conceptual understanding of learning. These applications push statistical inference into uncharted territories where data is high-dimensional and scarce, and prior information on ``true'' models is scant if not totally absent. 
Here we review recent progress on understanding learning, based on the notion of "relevance". The relevance, as we define it here, quantifies the amount of information that a dataset or the internal representation of a learning machine contains on the generative model of the data. This allows us to define maximally informative samples, on one hand, and optimal learning machines on the other. These are ideal limits of samples and of machines, that contain the maximal amount of information about the unknown generative process, at a given resolution (or level of compression). 
Both ideal limits exhibit critical features in the statistical sense: Maximally informative samples are characterised by a power-law frequency distribution (statistical criticality) and optimal learning machines by an anomalously large susceptibility. 
The trade-off between resolution (i.e. compression) and relevance distinguishes the regime of noisy representations from that of lossy compression. These are separated by a special point characterised by Zipf's law statistics. This identifies samples obeying Zipf's law as the most compressed loss-less representations that are optimal in the sense of maximal relevance.
Criticality in optimal learning machines manifests in an exponential degeneracy of energy levels, that leads to unusual thermodynamic properties. This distinctive feature is consistent with the invariance of the classification under coarse graining of the output, which is a desirable property of learning machines.
This theoretical framework is corroborated by empirical analysis showing {\em i)} how the concept of relevance can be useful to identify relevant variables in high-dimensional inference and {\em ii)} that widely used machine learning architectures approach reasonably well the ideal limit of optimal learning machines, within the limits of the data with which they are trained.
\end{abstract}




\end{frontmatter}

\tableofcontents

\newpage

\begin{quote}
Yet in truth there is no form that is with or without features;\\ 
he is cut off from all eyes that look for features. \\
With features that are featureless he bears a featured body,\\
 and the features of living beings with their featured bodies are likewise.\\
\centerline{\em the Immeasurable Meanings Sutra}~\cite{soka2009lotus}
\end{quote}

\rightline{\add{In memory of Miguel Virasoro.}}

\section{Introduction}
\label{sec:intro}

\add{To date there is no theory of learning -- the ability to ``make sense" of hitherto unseen raw data \cite{barlow1989unsupervised} -- that is independent of what is to be learned, and how and/or why it is to be learned. Both in supervised or unsupervised settings of statistical inference and machine learning, what ``making sense" means is defined at the outset, by encoding the task (regression, classification, clustering, etc.) in an objective function \cite{barberBRML2012}. This turns learning and inference into an optimisation problem\footnote{\add{Silver {\em et al.}~\cite{silver2021reward} suggest that indeed maximisation of reward is all that is needed to account for intelligent behaviour, including learning. Even if this were true, the key problem is that, in many interesting cases, the reward function is unknown to start with, and it is precisely what one would like to learn from data.}}.} \blue{In this perspective relevance of a dataset or a representation is a relative concept, that is, with reference to the task and the corresponding optimisation problem at hand. For example, for an experimental biologist, a gene expression dataset might be relevant for identifying genes important for cancer, but not for other diseases.}


As observed by Wigner~\cite{Wigner}, identifying {\em a priori} what features are important in each situation has been essential in order to design experiments that can reveal the mechanisms that govern natural systems. The {\em a priori} assumption about how relevant information is coded in the data may be derived from analysing mathematical models or through what is often a long history of trials and errors. Regardless of how it is obtained, this prior knowledge makes it possible to \add{precisely define the inference task, thereby allowing us to} extract meaningful information (about e.g. gravitational waves~\cite{gravitationalwaves}) even from extremely noisy data (e.g. on the imperceptible motion of interferometers). \add{But does it have to be the case that the usefulness or relevance of a dataset can only be defined in terms of what the data is {\em a priori} assumed to be useful for?}


\add{In this review, we argue that it is, \blue{indeed} possible to assign a quantitative measure of relevance to datasets and representations.}\blue{ Furthermore, this quantitative measure allows us to rank datasets and representations according to a universal notion of "relevance" that we show to shed light on problems in a number of settings ranging from biological data analysis to artificial intelligence.}
We shall argue that information theory can be used to tell us when a dataset contains interesting information, even if we don't know what this information is about, or to distinguish a system that learns about its environment, from one that does not, even without specifying {\em a priori} what is learned. \add{Just like maximum entropy codifies Socrates's ``I know that I know nothing'' in precise mathematical terms, the principle of maximal relevance can be used to characterise an information rich state of knowledge. As we shall see, ``relevance'' is recognisable by our naked eye: 
paraphrasing Justice Potter Stewart in {\em Jacobellis v. Ohio},  "[we] know it when [we] see it."} 


\add{This perspective on learning and inference based on an intrinsic notion of relevance has been developed in a series of recent works~\cite{MMR,statcrit,Odilon}. The aim of this review is to provide an unified picture of this approach. 
An approach based on an intrinsic notion of relevance allows us to define {\em maximally informative samples} without the need of specifying what they are informative about, and {\em optimal learning machines}, irrespective of \blue{what the data they are trained with is}. These will be the subjects of Sections~\ref{sec:inference} and~\ref{sec:learning}, respectively. Before that, Section~\ref{sec:general}
lays out the main framework and it introduces the concept of relevance for a statistical model and \blue{for} a sample. 
Section \ref{sec:peculiar} compares the statistical mechanics of optimal learning machines to that of physical systems. We conclude with a discussion of avenues of possible further research in the final Section. The rest of the introduction {\em i)} provides further motivation for the introduction of the notion of relevance, on the basis of general arguments as well as of some examples, {\em ii)} it clarifies its relation with the entropy as a measure of information content and how this notion fits into the growing literature on statistical learning, and {\em iii)} it summarises the results that are discussed in the rest of the review.}

\subsection{Understanding learning in the under-sampling regime requires a notion of relevance}

When the number of samples is very large with respect to the dimensionality of the statistical learning problem, then the data can speak for itself. This is the classical domain of statistics that we shall refer to as the {\em over-sampling} regime. 
The term ``dimension'' refers either to the number of components of each data point or to the number of parameters of the model used to describe the data. 
Classical statistics usually considers the asymptotic regime where the number of samples diverge, while the dimension of the data and/or of the model are kept constant. In this classical regime, the trial and error process of iteration between data and models easily converges. When the data is so abundant, relevant features emerge evidently from the data and the information content can be quantified in terms of the entropy of the inferred distribution. A quantitative notion of relevance is redundant in the over-sampling regime.

\add{Modern applications of statistical learning address instead the ``high-dimensional'' regime where the number of data points in the sample is smaller or comparable to the dimension of the data or of the model.} Single cell gene expression datasets, for example, are typically high-dimensional because the number of samples \add{(cells) is often smaller than the dimension (the number of tracked genes)\footnote{The number of genes ranges in the tens of thousands, which is generally larger than the number of cells, apart from special cases~\cite{park2020cell}.}.} Models used in statistical inference on protein sequences, as those of Morcos {\em et al.}~\cite{morcos}, can have millions of parameters \add{(see footnote~\ref{protparam})}. This is a high-dimensional inference problem because the number of available sequences is usually much smaller than that. 
\add{These are examples of situations where the data is scarce and high-dimensional, and where there is no or \blue{little} clue on what the generative model is. This is what we call the {\em under-sampling} regime of statistical learning.} In this regime, even the estimate of the entropy can be problematic~\cite{entropy_est}, let alone extracting features or estimating a probability density. 

In the under-sampling regime, a quantitive measure of "relevance" becomes useful for several reasons. First, \add{classical statistical methods, such as clustering or principal component analysis, are based on {\em a priori} assumptions on statistical dependencies. They often assume low order (e.g. pairwise) statistical dependencies, even when there is no \blue{compelling} reasons to believe that no higher order interaction should be there. Likewise, regularisation schemes need to be introduced to \blue{e.g. avoid overfitting or to} tame the fluctuations of inferred model parameters \cite{barberBRML2012}.} In these situations, \add{it may be hard to control the unintended biases that such \blue{assumptions} impose on the inference process.}
The measure of relevance we shall introduce depends only on the data and it is model-free. Relevance based inference techniques, such as those of~\cite{Grigolon,RyanMSR}, overcome these difficulties.

A second reason is that an absolute notion of relevance is crucial to understand learning in the under-sampling regime, which is, for example, the regime where deep learning operates. Quantifying learning performance is relatively easy in the over-sampling regime, where features and patterns emerge clearly from noise \add{and there is relatively small model uncertainty\footnote{See Section~\ref{sec:stat_inf} for a more detailed discussion.}.} 
There are no easily identifiable features that can be used to quantify learning in the under-sampling regime. \add{Indeed all measures of learning performance are based on task-dependent quantities. This makes it hard to} understand (deep) learning in the under-sampling regime in a way which is independent of what is learned and on how it is learned.

A third reason is that a quantitative measure of relevance allows us to define \add{the ideal limit of} {\em maximally informative samples}. These are samples for which the relevance is maximal, i.e. which potentially contain a maximal amount of information on their (unknown) generative process. Likewise, 
the notion of relevance, \add{applied to a statistical model,} allows us 
to define \add{the ideal limit of} {\em optimal learning machines}, as those statistical models with maximal relevance. Assuming a {\em principle of maximal relevance} as the basis of learning and inference, brings with it a number of predictions, much like the principle of maximal entropy dictates much of thermodynamics.
These prediction can be tested and used to identify relevant variables in high dimensional data or to design optimal learning machines. 

Also, an absolute notion of relevance allows us to better understand learning and inference in biological systems.
Learning is a distinguishing feature of living systems and evolution \blue{is likely to reward organisms that can detect significant cues from their environment on as little data as possible compared to those that don't.
For example, bacteria that rely on trial and error, or on measuring concentration gradients to arbitrary precision, before mounting the appropriate response, will, likely, not be successful in the struggle for survival}. Also, information processing impinges on the energy budget of living organisms because it requires them to maintain out-of-equilibrium states. The data that evolution dispenses us with on genetic sequences, or that we can collect from experiments that probe biological functions, necessarily reflects these constraints and calls for a more fundamental understanding of statistical inference in the under-sampling regime. \add{A deeper understanding of learning, based on the concept of relevance,} can suggest \blue{novel approaches} to distinguish a system that learns from one that does not, i.e. to tell apart living systems from inanimate matter \cite{Davies}.

\add{In spite of the spectacular successes of artificial learning machines, we are still very far from a satisfactory understanding of learning.} \blue{The performance of these learning machines is often very far from that of our brain and expert systems with performances comparable to those of human intelligence (e.g. in automatic translation) are limited to very specific tasks and require energy costs orders of magnitude greater than those employed by the human brain~\cite{energyML}}. In addition, the data on which machine learning algorithms are trained is abundant, \add{whereas even infants are capable of learning from few examples~\cite{vinyals2016matching}. A quantitative notion of relevance may offer hints on how the gap between artificial and human performance in learning can be filled.} 

\subsection{Relevance in some concrete examples}
\label{examples}

The previous subsection advocated for a quantitative measure of relevance in abstract and generic terms. In order to get a better sense of what we mean by relevance, we mention three examples: understanding how amino-acid sequences encode the biological function of a protein, how neurons encode and transmit information, and how deep neural networks form internal representations.
\begin{itemize}
\item Protein sequences in databases such as UniProt~\cite{uniprot2015uniprot} have been generated by evolution. These sequences contain information on the biological function of a protein, because features that are essential to carry out a specific function in a given organism have been protected in the evolutionary process. One of these features is the three dimensional structure of the folded protein. The structure is maintained by molecular forces acting between pairs of amino-acids that are in close \add{spatial} proximity in the \add{folded} structure, but that may be far away along the sequence. The mutation of one of the two amino acids needs to be compensated by a mutation on the other, in order to preserve the contact. 
Hence contacts between amino acids can be detected by analysing the co-evolutionary traces left in a large database of sequences for that protein across evolutionarily distant organisms~\cite{morcos}. {Contacts primarily involve pairs of amino-acids, so one might consider pairwise statistical correlations as the relevant variables and pairwise statistical models as the appropriate tool for making statistical inferences and predictions about the physical properties of proteins, e.g. their folding properties. Within this framework,} all other information contained in the data is irrelevant. Yet evolution operates on the whole protein sequence in order to preserve a biological functions that includes, but is not limited to contacts between amino acids\footnote{\add{The biological function of a protein is also related to its interaction with other proteins and molecules inside the cell, see e.g.~\cite{bitbol2016inferring}.}}. Hence, evolution likely leaves traces beyond pairwise statistics in a dataset. {Interestingly, even for protein contact prediction, pseudo-likelihood methods that exploit information on the whole sequence have a superior performance compared to statistical inference of pairwise models~\cite{Aurell}}. Detecting these traces may shed light on what is {\em relevant} for implementing a specific biological function. Understanding what that unknown function is requires us to focus on what is deemed relevant by a protein, {\em i.e.} to consider {\em relevance as an intrinsic property.}

\item {It is a general practice in neuroscience to record the activity of neurons during a particular task while certain external behavioural variables
~are measured. One will then try to find whether certain causal or non-causal correlations exist between the recorded activity and the measured covariate. The research that lead to the discovery of grid cells~\cite{Mosers} is a good example. Grid cells are specialised neurons in the Medial Entorhinal Cortex (MEC) believed to be crucial for navigation and spatial cognition. They were identified by analysing the correlations between the activity of neurons in the MEC of a rat roaming in a box and the position of the animal. This analysis revealed that a grid cell in MEC fires\footnote{The activity of neurons generally consists of electric discharges across the synaptic membrane. These events are called {\em spikes}, and they occur over time-scales of the order of one millisecond. In neuroscience's jargon, a neuron ``fires'' when a spike occurs. A sequence of spikes is called a {\em spike train}.} at particular positions that together form a hexagonal pattern. This suggests that grid cells are important for spatial navigation and cognition. Not all neurons in the MEC, however, correlate significantly with spatial correlates, or any other measured covariate. However, this does not imply that their activity is irrelevant for spatial cognition, or for some other aspect of the behaviour that the animal exhibits. Whatever these neurons may encode, their information is passed on to neurons in higher cortical regions for further processing. How does one of these neurons ``decide'' which neuron in the MEC to ``listen'' to in order to perform a task? Notice that such upper stream neurons do not have access to spatial covariates, like the experimentalist does. Their decision has to be based on the neural spike trains alone. Hence, all the information on which neuron should be listened to or not,  must be contained in the neural spike trains. {\em It should be possible to define relevance just from the data}, in a model-free manner, because that is what seems neurons are capable of doing, perhaps to different degrees, during development, learning and performance of a task. We expand further on these subjects in Section~\ref{sec:neuro}.}

\item {"Deep neural networks" is a generic term for artificial neural networks used for many data processing tasks. They are formed by several layers of units (called neurons) stacked one on top of the others. Neurons in one layer integrate inputs from the neurons in the layer below, and pass \add{them} to neurons in the layer above. Deep neural networks involve millions of parameters that determine how the activity of one neuron affects that of another. These parameters can be set using one of a large repertoire of algorithms. Deep neural network can discriminate images (e.g. dogs from cats), exploiting information in the training set which goes beyond the features which distinguish dogs from cats for us. One such feature could be, e.g., that} dogs are more frequently portrayed outdoor than cats. So the background of the picture turns out to be a relevant feature. Among features that deep learning machines find relevant, some resemble noisy patterns to us~\cite{AleksanderMadry}. 
Machines that only use the features which are reasonable for humans perform worse than those that use all features, in terms of generalisation~\cite{AleksanderMadry}. Features are just statistically significant patterns and their relevance is determined by the relative frequencies with which they occur in the dataset. Ultimately, {\em relevance is frequency}.
\end{itemize}

\subsection{Relevance defined}

Relevant features, intuitively speaking, are what makes it possible to discriminate different objects. Similar objects instead differ by irrelevant details, {that is usually denoted as} noise. In statistics and information theory, this intuitive notion of noise can be made precise by the maximum entropy principle -- that characterises a state of maximal ignorance -- and the asymptotic equipartition property~\cite{CoverThomas}. The latter implies that in a simple probabilistic model the logarithm of the probability of typical outcomes varies in a very narrow range. Likewise, a structureless dataset should feature a small relative variation in the frequency with which different outcomes are observed\footnote{\add{This description also applies to sets of weakly dependent random variables, such as (Ising) models of ferromagnets and error correcting codes~\cite{sourlas1989spin}, for example. Also in these cases the log-probability of typical outcome satisfies concentration principles, i.e. it exhibits a small variation.}}. Hence what distinguishes two outcomes with similar probabilities or frequencies may be fortuitous~\cite{HM}, but outcomes that occur with very different frequencies in a dataset or that have very different probabilities must differ by some relevant detail. 
In the under-sampling regime, when no other information is available, the frequency is the only quantity that can be used to distinguish outcomes.
We shall then define relevance as an information theoretic measure of the variation of the log-probability in a probabilistic model, and of the frequency in a dataset. In this way, relevance provides a measure of the discriminative power of models and of datasets.

The relevance captures the fact that a dataset that exhibits a high variation of the frequency across outcomes, or a model with a large variation of the log-probability, likely exhibit a significant variation of relevant features\add{ (those that allow a machine to discriminate between inputs)}. This applies independently of whether the relevant features are known {\em a priori} or not.

\subsection{Relevance is not resolution}
\label{sec:RelnotRes}
\add{From the point of view of coding theory, learning amounts to summarising a ``source'' (i.e. a dataset or a probability distribution) into a compressed representation. This transformation is called a {\em code}, and it associates to each outcome (a "word") a codeword which is a string of bits. The length of the codeword is the {\em coding cost}, i.e. the number of bits needed to represent that outcome. Efficient codes are those that achieve maximal compression of a long sequence of outcomes drawn from the source, i.e. a sample. This means that efficient codes minimise the total coding cost. The main insight in efficient coding is that short codewords should be assigned to frequent outcomes and longer codewords to rarer ones. In quantitative terms, the optimal {\em coding cost} of an outcome is equal to minus the logarithm of its probability. Hence, the optimal average coding cost is the entropy, which is Shannon's bound \blue{\cite{CoverThomas}}.}

The coding cost is a natural measure of information content and is the basis of several information theoretic approaches in data analysis, such as the Infomax principle~\cite{linsker1988self} or the information bottleneck~\cite{IB}. It is important to explain why relevance is not a measure of the average coding cost, but it is a measure of its variation (across a sample or across typical outcomes of a distribution). 

First, the coding cost is not an absolute measure, because it depends on how the variables are defined. The same dataset of images can be described at different resolutions and genes can be defined in terms of their genomic sequences, or of the sequence of amino-acids of the proteins they code for. 
This is why, in this review, the average coding cost is considered as a measure of {\em resolution} \blue{and not of relevance}. In high-dimensional data, resolution can be tuned by dimensional reduction techniques or by varying the number of clusters in a data-clustering task, and in a neural network the resolution of the internal representation can be adjusted varying the number of hidden nodes. 
Relevance quantifies instead the dynamical range that the data or the representation spans at a given level of resolution, and it is an intrinsic property. \footnote{\add{It is possible to express the variation of the frequency in different ways, e.g. with the variance of the coding cost (i.e. the heat capacity), as done e.g. in Ref.~\cite{retina,MoraBialek}. A definition of variation in terms of the second moment implicitly assumes \blue{that the Gaussian distribution, which is indeed uniquely defined by the first and second moments, is a sensible description of the distribution of} the coding cost. Our definition of relevance in terms of the entropy\blue{, instead,} does not make any assumption on the distribution of the coding cost}.} 
As the resolution varies, the relevance attains different values. The trade-off between relevance and resolution is, in our opinion, a key element to understand learning in the under-sampling regime.
%

\add{Secondly, in the under-sampling regime the data gives little clues on what the generative model can be, and attempts to estimate a model are doomed by over-fitting. The average coding cost does not capture the uncertainty on the generative model. The relevance instead quantifies this uncertainty in the sense that data with higher relevance have smaller indeterminacy on the underlying generative model}. The strategy that we shall follow in the under-sampling regime is to consider the average coding cost -- i.e. {\em resolution} -- as an independent variable\footnote{\blue{In practice, and as alluded to before, the resolution can be varied in different ways, e.g. through clustering or merging data points, and how this is done depends on the setup as will be clarified in more details in examples discussed in the following sections.}}, and the {\em relevance} -- i.e. the variation of the coding cost -- as the dependent variable, or the objective function that should be optimised in order to obtain maximally informative representations.

\subsection{Relevance and statistical learning}

Before proceeding, it may help to discussed the relation of the material discussed in this review with mainstream approaches in statistical learning. This will necessarily be a biased and incomplete account, that serves the main purpose of orienting the reader and providing a perspective. 

Statistical and machine learning borrows much of its conceptual basis from the classical statistics approach, in which statistical inference is turned into the optimisation problem of an error or a likelihood function over a set of parameters. The advent of high-throughput experiments and Big Data provided us with a wealth of data on complex systems we know very little about, such as cells~\cite{marx2013} or the brain~\cite{sejnowski2014putting}, our economies~\cite{varian2014big} and societies~\cite{Lazer721}. \add{This is an uncharted territory for statistical learning, that we refer to as the under-sampling domain, because of the high-dimensionality of the data and of the lack of theoretical models.} Machine learning has given up on the presumption that the models used have anything to do with an hypothetical ``true'' model. Learning machines have their values in their ability to capture statistical dependencies and to generalise the data they have been trained with. This has led to spectacular successes in automated tasks such as voice recognition and image classification~\cite{lecun2015deep}. The emphasis on optimal performance in specific tasks and for specific data, has led to a classification of learning into different domains (supervised, unsupervised, reinforcement learning) and sub-domains (regression, classification, clustering, etc). By contrast, this review takes the perspective of studying learning as a general phenomenon, much in the same way as statistical mechanics is a general theory of thermodynamic equilibrium in physics. Because of this, the material of this review is admittedly far from the most advanced applications of machine learning.

This review also departs from the statistical mechanics approach to learning (see e.g.~\cite{engel2001statistical}). This has led to several interesting insights on the free energy landscape of learning machines~\cite{tubiana2017emergence,decelle,zecchina,hennig,biroli} and in powerful results in the teacher-student case~\cite{engel2001statistical} or on learning and signal detection in extremely noisy regimes~\cite{zdeborova2016statistical}. Statistical learning is the inverse problem of statistical mechanics. While the former aims at inferring models from data, the latter describes the properties of data generated by a given model; \blue{this relationship between statistical mechanics and statistical learning is illustrated, for example, by the case of Ising models and their inverse inference problems \cite{hertz2013ising,nguyen2017inverse}}. A statistical mechanics approach needs to assume at the outset what the data and the task of learning is. This defies at the outset the goal of defining what relevance is, in a way that is independent of the data and of the task.

Machine learning is the ideal test-ground for any definition of relevance, and this is the way it will be used in this review. Our starting point is that learning machines ``know'' what relevance is, because with no prior knowledge and on the basis of the data alone, well trained machines have the power to generate realistic instances of pictures or text, for example. Therefore relevance should be encoded in their internal structure in a precise and quantifiable manner. 

\subsection{Summary of main results}

This review discusses two complementary perspectives on statistical learning, from the point of view of a sample and from the point of view of a statistical model. We shall specifically discuss models that corresponds to the internal representation of a learning machine. 

\add{Loosely speaking,} for a sample, the relevance is defined as the amount of information that the sample contains on the generative model of the data~\cite{MMR,statcrit}. For a learning machine, the relevance quantifies the amount of information that its internal representation extracts on the generative model of the data it has been trained with~\cite{Odilon}. An informative sample can be thought of as being a collection of observations of the internal states of a learning machine trained on some data. This correspondence puts the two perspectives in a dual relation, providing consistency to the overall framework.

\subsubsection{Relevance bounds}

\blue{As will be discussed in Section \ref{sec:inference}}, the significance of the relevance for learning can be established through information theoretic bounds. On one side we shall see that the relevance {\em upper bounds} the amount of information that a sample contains on its generative model. This allows us to derive an upper bound on the number of possible parameters of models that can be estimated from a sample, within Bayesian model selection. Even though this bound is approximate and tight only in the \blue{extreme} under-sampling regime \blue{see Section \ref{sec:bound}}, it is the only result of this type that we're aware of.

For learning machines, we review the results of Ref.~\cite{Odilon} that show that the relevance {\em lower bounds} the mutual information between the internal state of the machine and the ``hidden'' features that the machine extracts. This provides a rationale for the principle of maximal relevance, \blue{namely the principle that the internal representation of a learning machine must have maximal relevance}, because it guarantees that learning machines extract at least an amount of information equal to the relevance, about the hidden features of the data they're trained with.

\subsubsection{Maximal relevance, physical criticality and statical criticality} 

In statistical physics, the term ``critical'' refers to those particular states of matter where anomalous fluctuations are observed, as a consequence of the physical system being poised at a critical point of a continuous phase transition. In many cases, a symmetry of the system is spontaneously broken across the phase transition. 
Statistical criticality, on the other hand, refers to the \add{ubiquitous} occurrence of anomalous fluctuations and broad distributions in \blue{various statistics collected from} systems as diverse as language~\cite{Zipf,i2005variation,baixeries2012exponent}, biology~\cite{Immune,retina,MoraBialek,Mora5405,Hidalgo,beggs2008criticality} and economics~\cite{city}, and it has attracted a  great deal of attention in the physics community~\cite{SOC,newman2005power,sornette2006critical,clauset2009power,MSN,ACL,munoz2018colloquium}. 

The notion of relevance discussed in this review, provides a link between these notions of criticality in statistical physics with what "being critical" means in common language, that is the ability to discriminate what is relevant from what is not. Indeed we show that maximal discrimination of what is relevant from what is not in a learning machine implies criticality in the statistical physics sense. At the same time,\blue{we demonstrate that} a sample that has maximal power \add{in discriminating between competing models} exhibits statistical criticality. 
We shall establish this link from the two complementary perspectives discussed above: that of a sample and that of the statistical model of the internal representation of a learning machine. The principle of maximal relevance identifies {\em Maximally Informative Sample} (MIS) and {\em Optimal Learning Machine} (OLM) as those samples and models, respectively, that maximise the relevance at a given level of resolution. 
As we shall see, statistical criticality arises as a general consequence of the principle of maximal relevance~\cite{statcrit}. 
The ideal limit of MIS describes samples that are expressed in terms of most relevant variables and, as a result, they should exhibit statistical criticality. The ideal limit of OLM describes learning machines that extract information as efficiently as possible from a high-dimensional dataset with a rich structure, at a given resolution. We show that well trained learning machines should exhibit features \add{similar to those that appear in physical system at the critical point of a second order phase transition.} 
The connection between criticality and efficiency in learning has a long tradition~\cite{langton1990computation} and has become a general criterium in the design of recurrent neural networks (see e.g. \cite{bertschinger2004real}) and reservoir computing (see e.g.~\cite{livi2017determination}). 
Our definition of an absolute notion of relevance provides a very general rationale for the emergence of criticality in maximally informative samples and efficient learning machines. The principle of maximal relevance that we suggest lies at the basis of efficient representations, and it is consistent with the hyperbolic geometric nature that has been observed in several neural and biological circuits (see \cite{sharpee2019argument}).

\subsubsection{How critical?}

In statistical mechanics, criticality emerges when a parameter is tuned at a critical point, beyond which a symmetry of the system is spontaneously broken. What is this parameter and which is the symmetry that is broken in efficient representations? We review and expand the results of Ref.~\cite{RyanMDL} that suggest that the critical parameter is the resolution itself and the symmetry which is broken is the permutation symmetry between outcomes of a sample. This picture conforms with the idea that efficient representations are maximally compressed\blue{: the phase transition} occurs because further compression cannot be achieved without altering dramatically the statistical properties of the system. In particular the symmetry broken phase is reminiscent of the mode collapse phenomenon in generative adversarial networks~\cite{goodfellow2014generative}, whereby learned models specialise on a very limited variety of the inputs in the training set.

\subsubsection{The resolution relevance tradeoff \add{and Zipf's law} }

The approach to learning discussed in the next pages, reveals a general tradeoff between resolution and relevance in both MIS and OLM. This identifies two distinct regimes: at high resolution, efficient representations are still noisy, in the sense that further compression brings an increase in relevance that exceeds the decrease in resolution. At low levels of resolution the tradeoff between resolution and relevance is reversed: not all the information that is compressed away is informative on the underlying generative process. The rate of conversion between resolution and relevance corresponds to the exponent that governs the power law behaviour of the frequency distribution in a sample. This allows us to attach a meaning to this exponent and it sheds light on its variation (e.g. in language~\cite{i2005variation}). In particular, this tradeoff identifies a special point where the rate of conversion of resolution into relevance is exactly one. This corresponds to a maximally compressed lossless representation and it coincides with the occurrence of Zipf's law~\cite{Zipf} in a sample\footnote{Zipf's law is an empirical law first observed by Zipf~\cite{Zipf} in language. It states that, in a large sample, frequency of the $r^{\rm th}$ most frequent outcome is proportional to $1/r$. It is equivalent to the statement that the number of outcomes observed $k$ times in the sample is proportional to $k^{-2}$.}. 
The occurrence of Zipf's law in systems such as language~\cite{Zipf}, the neural activity in the retina~\cite{retina} and the immune system~\cite{Immune,Mora5405} suggests that these are maximally compressed lossless representations. From the perspective of learning machines, the optimal tradeoff between resolution and relevance identifies representations with optimal generative performance, as discussed in~\cite{SMJ}.

\subsubsection{Statistical mechanics of optimal learning machines}

The principle of maximal relevance endows OLM with an exponential density of states (i.e. a linear entropy-energy relation \add{in the statistical mechanics analogy of Ref.~\cite{MoraBialek}}). This in turn determines very peculiar statistical mechanics properties, as compared to those of typical physical systems. OLM can be discussed as a general optimisation problem and their properties can be investigated within an ensemble of optimisation problems where the objective function is drawn from a given distribution, as in Random Energy Models~\cite{REM}. This analysis, \add{carried out in Ref.~\cite{xie2021random}, reveals that\add{, within this approach,} information can flow across different layers of a deep belief network only if each layer is tuned to the critical point.} So in an ideal situation, all the different layers should have an exponential density of states. It is tempting to speculate that this should be a general property of learning: experts learn only from experts. 
The general optimisation problem studied in Section~\ref{sec:peculiar} \add{following Ref.~\cite{marsili2019peculiar},} reveals an interesting perspective on OLM in the limit where the size of the environment they interact with (the heat bath or the dimensionality of the data) diverges. Indeed OLM sit at the boundary between physical systems, whose state is largely independent of the environment, and unphysical ones, whose behaviour is totally random. Only when the density of states is a pure exponential the internal state of the sub-system is independent of the size of the environment. This suggests that an exponential density of states is important to endow a learning machine with an invariance under coarse graining \add{of the input} that allows it to classify data points (e.g. images) in the same way, irrespective of their resolution (in pixels). 


\section{General framework and notations}
\label{sec:general}

Consider a generic complex system whose state is defined in terms of a vector $\vec x\in \mathbb{R}^d$\add{, where $d$ is the dimensionality of the data}. 
Formally, we shall think of $\vec x$ as a draw from an unknown probability distribution $\bp(\vec x)$ \blue{.This distribution is called the {\em generative model}}, because it describes the way in which $\vec x$ is generated. Here and in the rest of the paper, backslashed symbols refer to unknown entities. For example, if $\vec x$ is a digital picture of a hand written digit, a draw from the generative model $\bp$ is a theoretical abstraction for the process of writing a digit by a human.

Contrary to generic random systems, typical systems of interest have a specific structure, 
they perform a specific function and/or they exhibit non-trivial behaviours. Structure, functions and behaviours are {\em hidden} in the 
statistical dependencies between variables encoded in the unknown $\bp(\vec x)$. Strong statistical dependencies suggest that typical 
values of $\vec x$ are confined to a manifold whose dimensionality -- the so-called {\em intrinsic dimension}~\cite{ansuini2019intrinsic} --  is much smaller than $d$.

\subsection{Resolution as a measure of the coding cost of a learning machine}

Learning amounts to finding structure in the data\footnote{\add{In this review, we do not discuss reinforcement learning, where learning occurs while interacting with an environment, with the objective of maximising a reward function.}}. In unsupervised learning, this is done without using any \blue{external} signal on what
the structure could be. More precisely, learning amounts to searching a mapping $p(\vec x|\bs)$ that associates to each $\vec x$ a compressed representation in terms of a discrete variable $\bs\in\cS$, so that $p(\vec x|\bs)$ describes ``typical objects'' of type $\bs$. \blue{For example, in the case of unsupervised clustering of the data $\vec x$, the variable $\bs$ may indicate the label of the clusters.}

Training on a dataset of observations of $\vec x$ induces a distribution $p(\bs)$ in the internal states of the learning machine, such that the generating model
\begin{equation}
\label{pxdef}
p(\vec x)=\sum_{\bs\in\mathcal{S}}p(\vec x|\bs)p(\bs),
\end{equation}
is as close as possible to the unknown distribution $\bp$, within the constraints imposed by the architecture of the learning machine used and the available data. Similar considerations apply to supervised learning tasks that aim to reproduce a functional relation 
$\underline{x}_{\rm out}=f(\underline{x}_{\rm in})$ between two parts of the data 
$\vec x=(\underline{x}_{\rm in},\underline{x}_{\rm out})$, where 
$\bp(\vec x)=\bp(\underline{x}_{\rm in})\delta\left( \underline{x}_{\rm out} - f(\underline{x}_{\rm in})\right)$. In this case, marginalisation on the internal states as in Eq.~(\ref{pxdef}) generates a probabilistic association $p(\underline{x}_{\rm in},\underline{x}_{\rm out})$ between the input and the output. 

For example, in Restricted Boltzmann Machines (RBM) $\bs=(s_1,\ldots, s_n)$ is a vector of binary variables that corresponds to the state of the hidden layer, whereas $\vec x=(x_i,\ldots,x_m)$ is the data vector in input, which correspond to the so-called visible layer\add{\footnote{\add{Soft clustering is a further example, whereby each datapoint $\vec x$ is associated to a distribution $p(\bs|\vec x)$ over a discrete set of labels. The case of hard clustering, when $p(\bs|\vec x)$ is a singleton, is discussed in Section~\ref{sec:dataclustering}.}}}. The distributions $p(\bs)$ and $p(\vec x|\bs)$ are obtained by marginalisation and conditioning, from the joint distribution
\begin{equation}
\label{eq:RBM}
p(\vec x,\bs)=\frac 1 Z \exp\left\{\sum_{i=1}^m a_i x_i+\sum_{j=1}^n b_j s_j +\sum_{i,j} x_i w_{i,j}s_j\right\}\,,
\end{equation}
where $Z$ is a normalisation constant. In unsupervised learning, the parameters $\theta=\{a_i,b_j,w_{i,j}\}_{i=1,n}^{j=1,m}$ are adjusted during training in order to maximise the likelihood of a dataset $\hat x=(\vec x^{(1)},\ldots,\vec x^{(N)})$ of $N$ observation, as discussed e.g. in~\cite{hinton2012practical}. In supervised learning instead, the parameters $\theta$ are adjusted in order to minimise the distance between the labels $\underline{x}_{\rm out}^{(i)}$ and the predictions 
$f(\underline{x}_{\rm in}^{(i)})$, for each datapoint $\vec x^{(i)}$. For more complex architectures (e.g. Deep Belief Networks~\cite{bengio2017deep}) that involve more than one layers of hidden units, we think of $\bs$ as the state of one of the hidden layers.


In this review, we abstract from details on the objective function employed or on the algorithm used, and we focus on the properties of the learned representation, i.e. on $p(\bs)$. For this reason, the dependence on the parameters $\theta$ will be omitted, assuming that they are tuned to their optimal values. 
Both $p(\bs)$ and $p(\bx)$ in Eq.~(\ref{pxdef}) are proper statistical models, as opposed to $\bp$ which is a theoretical abstraction.

Learning can be naturally described in terms of coding costs. The logarithm of the probability of state $s$
\begin{equation}
E_{\bs}=-\log p(\bs)
\end{equation}
is the coding cost of state \blue{$\bs$, i.e. the number of nats}\footnote{We shall use natural logarithms throughout, and nats as a measure of information.} used by the machine to represent \blue{$\bs$  \cite{CoverThomas}}. The average coding cost is the entropy
\begin{equation}
\label{eq:resolution_model}
H[\bs]=\mathbb{E}[E_{\bs}]=-\sum_{\bs\in\mathcal{S}} p(\bs)\log p(\bs),
\end{equation}
where henceforth $\mathbb{E}[\cdots]$ denotes the expectation value. The entropy measures the information content of the representations $p(\bs)$, that can be seen as the amount of resources (measured in nats) that the learning machine employs to represent the space of inputs $\vec x$. 
More detailed representations have larger values of $H[\bs]$ than coarser ones.
\blue{Recalling the discussion in section \ref{sec:RelnotRes}, we shall denote $H[\bs]$ as} {\em resolution} of the representation because it 
quantifies its level of compression. For example, in RBMs the resolution can be adjusted by varying the number $n$ of hidden units. Typically $H[\bs]$ will be an increasing function of $n$ in RBMs. 

\subsection{Relevance of a representation as the informative part of the average coding cost}

Part of the $H[\bs]$ nats is relevant information and part of it is noise, in the sense that it does not provide useful information on how the data has been generated. Irrespective of how and why a part of the data turns out to be uninformative, information theory allows us to derive a quantitative measure of noise in nats. Using this, Section \ref{sec:learning} will argue that 
the amount of information that the representation contains on the generative model $\bp$ is given by the {\em relevance}, which is defined as
\begin{equation}
\label{HErel}
H[E]=\mathbb{E}[-\log p(E)]\,,
\end{equation}
where 
\begin{equation}
\label{eq:pE}
p(E)=\sum_{s\in\mathcal{S}} p(s)\delta\left(E+\log p(s)\right)=W(E) e^{-E}
\end{equation}
is the distribution of the coding cost, and $W(E)$ is the number of states with coding cost $E_{\bs}=E$, that we shall call the density of states. 
Since $E_{\bs}$ is the coding cost, the relevance $H[E]$ coincides with the entropy of the coding cost. Given that $E_{\bs}$ is a function of $\bs$, we have that~\cite{CoverThomas}
\begin{equation}
\label{eq:noise_model}
H[\bs]=H[E]+H[\bs|E]\,,
\end{equation}
where $H[\bs|E]=\mathbb{E}\left[\log W(E_{\bs})\right]$ quantifies the level of noise in the representation\footnote{For ease of exposition, we focus on the case where both $E$ and $\bs$ are discrete variables. When $E$ is a continuous variable, Eq.~(\ref{HErel}) yields the differential entropy of $E$~\cite{CoverThomas}. Since $p(E|\bs)=\delta(E-E_{\bs})$ is a delta function, the differential entropy of $E$ conditional \blue{on} $\bs$ diverges to $-\infty$. This divergence can be removed if the entropy is defined with respect to a finite precision $\Delta$, as explained e.g. in~\cite{CoverThomas}. 
We refer to Ref.~\cite{Odilon} for a discussion of the general case.}. 

\begin{figure}[ht]
\centering
\includegraphics[width=0.5\textwidth,angle=0]{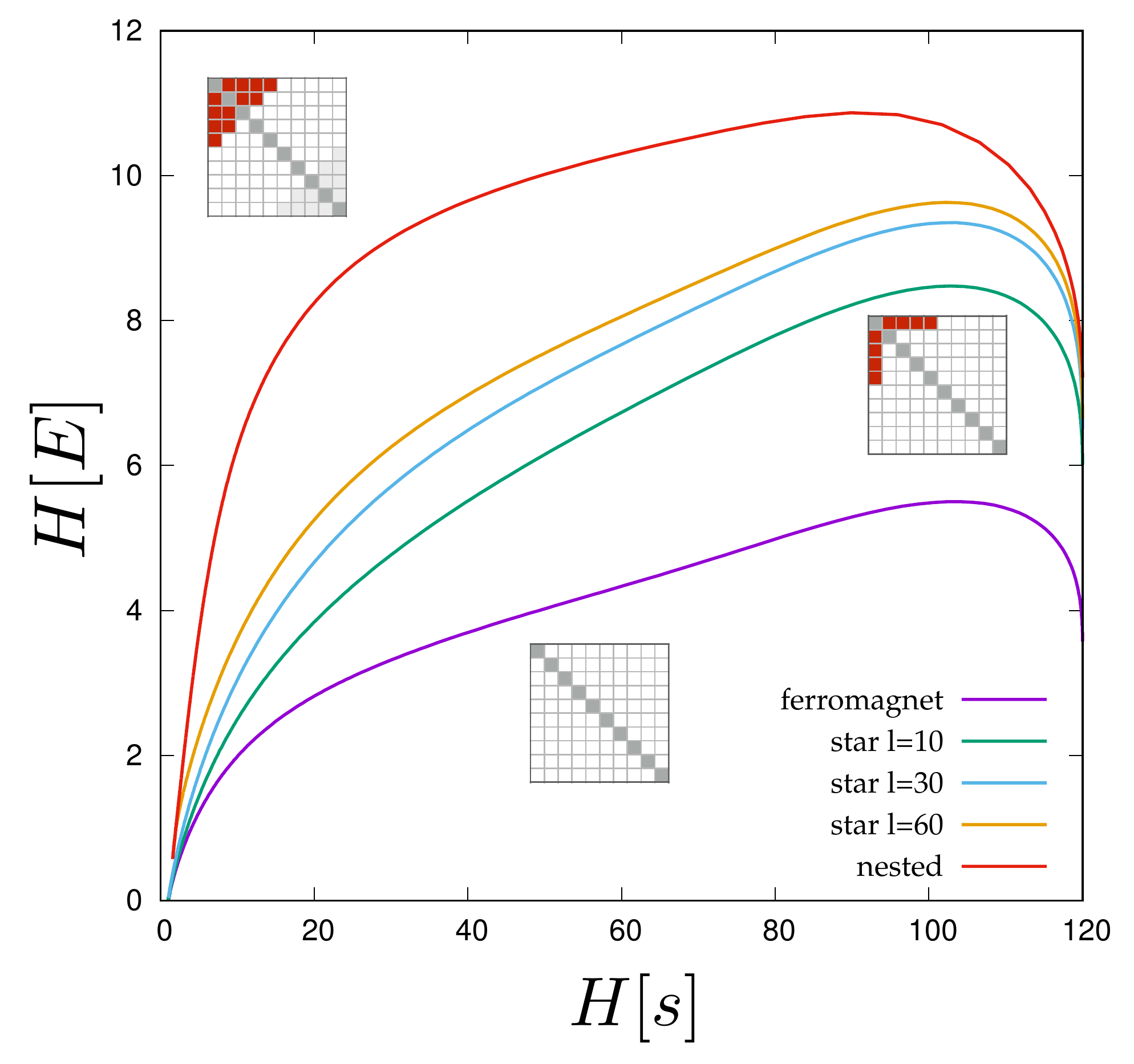}
\caption{\label{FigSpin} Relevance as a function of resolution for the fully connected pairwise models of $n=120$ spins, with $J_{i,j}=\pm J$ \add{discussed in Ref.~\cite{Odilon}}. The signs of the matrix $J_{i,j}$ is schematically depicted for the different models. From bottom to top, ferromagnetic model ($J_{i,j}>0$ for all $i\neq j$), star model ($J_{1,j}<0$ for $j\le l$ and $J_{i,j}>0$ otherwise) for $l=10, 30$ and $60$, nested model ($J_{i,j}<0$ for $i+j\le n/2$ and $J_{i,j}>0$ otherwise). \add{For each value of $J$, $H[\bs]$ and $H[E]$ can be computed for each of these models and the curves are obtained varying $J$ (see Ref.~\cite{Odilon} for more details)}.}
\end{figure}

\add{The relevance depends on the statistical dependencies between the variables $\bs$. As an example, Fig.~\ref{FigSpin} reports the dependence of the relevance $H[E]$ on the resolution $H[\bs]$ for the $p(\bs)$ that corresponds to different spin models\add{ where $\bs=(\sigma_1,\ldots,\sigma_n)$ is a string of $n$ variables $\sigma_i=\pm 1$}. As this figure shows, the relevance depends on the arrangement of couplings, in this case.} 

If $H[E]$ quantifies the information learned on $\bp$, then optimal learning machines (OLM) correspond to 
maximally informative representations at a given resolution $H[\bs]$. These are the solutions of the optimisation problem
\begin{equation}
\label{maxrel1}
\{E_{\bs}^*\}\in{\rm arg}\max_{E_{\bs}:~\mathbb{E}[E]=H[\bs]} H[E].
\end{equation}
\add{As we shall see, this optimisation principle dictates what the density of states $W(E)$ of an OLM should be, whatever the data $\hat x$ with which it has been trained.}

Section \ref{sec:learning} reviews the arguments of Ref.~\cite{Odilon} that show that {\em the relevance lower bounds the mutual information between the representation and the hidden features that the learning machine extracts from the data}. In addition, it reviews the evidence in support of the conclusion that, {\em within the constraints imposed by the architecture and the available data, real learning machines maximise the relevance}. 

\subsection{The relevance and resolution of a dataset}

Now imagine that \add{we're interested in a complex system described by a vector of variables $\vec x$, but} we only observe a sample $\hat \bs =(\bs^{(1)},\ldots,\bs^{(N)})$ of $N$ observations of a variable $\bs$ that probes the state $\vec x$ of the system. Now both $\vec x$ and the generative process $\bp(\bs)$ are unknown\footnote{\add{In the setting of the previous section, this would correspond to the situation where we observe a state $\bs^{(i)}$ of the hidden layer of \blue{a learning machine, for input} $\vec x^{(i)}$ of the visible layer. However inputs $\vec x^{(i)}$ are not observed and the model $p(\vec x,\bs)$ is also unknown.}}. 
Take for example a dataset $\hat \bs$ of amino acid sequences $\bs$ of a protein that performs a specific function (e.g. an ion channel or a receptor). In each organism, the functionality of the protein depends on many other unknown factors\blue{, $\vec x$,} besides the protein sequence $\bs$. These likely include the composition of the other molecules and proteins the protein interacts with in its specific cellular environment. 
An independent draw from the unknown generative process $\bp(\bs)$, in this case, is a theoretical abstraction for the evolutionary process by which organisms that perform that specific biological function efficiently have been selected. Hence the dataset contains information on the structure and the function of the protein that we informally identify with the amount of information that the sample contains on its generative process.



The problem of quantifying the information content of a sample $\hat \bs =(\bs^{(1)},\ldots,\bs^{(N)})$ is discussed in \blue{detail in} Section \ref{sec:inference}.
We assume that each observation $\bs$ belongs to a set $\mathcal{S}$. 
We make no assumption on $\mathcal{S}$, which may not be even fully known, as when sampling from an unknown population\footnote{Indeed, $\mathcal{S}$ could be countably infinite in cases where the sample could potentially be extended to arbitrarily large values of $N$.}. Each $\bs^{(i)}$ is an empirical or experimental observation carried out under the same conditions \blue{meaning that} it can be considered as an independent draw from the same\blue{, unknown} probability distribution $\bp(\bs)$. 
The empirical distribution $\hat p_{\bs}=k_{\bs}/N$ provides \blue{an estimate} of $\bp(\bs)$, where \add{the frequency} $k_{\bs}$, defined as
\begin{equation}
\label{eq:ks}
k_{\bs}=\sum_{i=1}^N\delta_{\bs^{(i)},\bs}\,,
\end{equation}
counts the number of times a state $\bs$ is observed in the sample. The entropy of the empirical distribution $\hat p_{\bs}$ 
\begin{equation}
\label{eq:resolution_sample}
\hat H[\bs]=-\sum_{\bs}\frac{k_{\bs}}{N}\log \frac{k_{\bs}}{N}\,.
\end{equation}
provides a measure of the information content of the sample, because this is the \blue{minimum} number of nats necessary to \blue{encode a data point from} the sample. 
Here and in the following, we shall denote with a hat $\hat{~}$ quantities that are estimated from the sample $\hat \bs$. In analogy with Eq.~\eqref{eq:resolution_model} \blue{and recalling the discussion in Section \ref{sec:RelnotRes}}, we shall henceforth call $\hat H[\bs]$ {\em resolution}.

\blue{It} is important to stress that we take \blue{the resolution,} $\hat H[\bs]$, as a quantitative measure of the coding cost \blue{of the specific sample $\bs$} and not as an estimate of the entropy $\bcancel{H}[\bs]$ of the unknown distribution $\bp(\bs)$. It is well known that $\hat H[\bs]$ is a biased estimator of the entropy $\bcancel{H}[\bs]$ of the underlying distribution $\bp(\bs)$~\cite{miller1955note,entropy_est}. As an estimator, $\hat H[\bs]$ is particularly bad specially in the under-sampling regime~\cite{entropy_est,NemenmanCoinc}. This is immaterial for our purposes, because our aim is not to estimate $\bcancel{H}[\bs]$ but rather to give a precise quantitative measure of the coding cost of a specific sample $\hat \bs$.

Section \ref{sec:inference} gives several arguments to support the conclusion that an upper bound of the amount of information that the sample $\hat \bs$ contains on its generative model is given by the {\em relevance}
\begin{equation}
\label{eq:relevance_sample}
\hat H[k]=-\sum_k\frac{km_k}{N}\log \frac{km_k}{N}\,,
\end{equation}
where
\begin{equation}
\label{eq:mk}
m_k=\sum_s\delta_{k_s,k}\,.
\end{equation}
is the state degeneracy, i.e. the number of states that are observed $k$ times. To the best of our knowledge, the relevance $\hat H[k]$ was first introduced as an ancillary measure -- called the degenerate entropy -- within an information theoretic approach to linguistics~\cite{naranan1992information,balasubrahmanyan2002algorithmic}. 

\begin{figure}[ht]
\centering
\includegraphics[width=0.5\textwidth,angle=0]{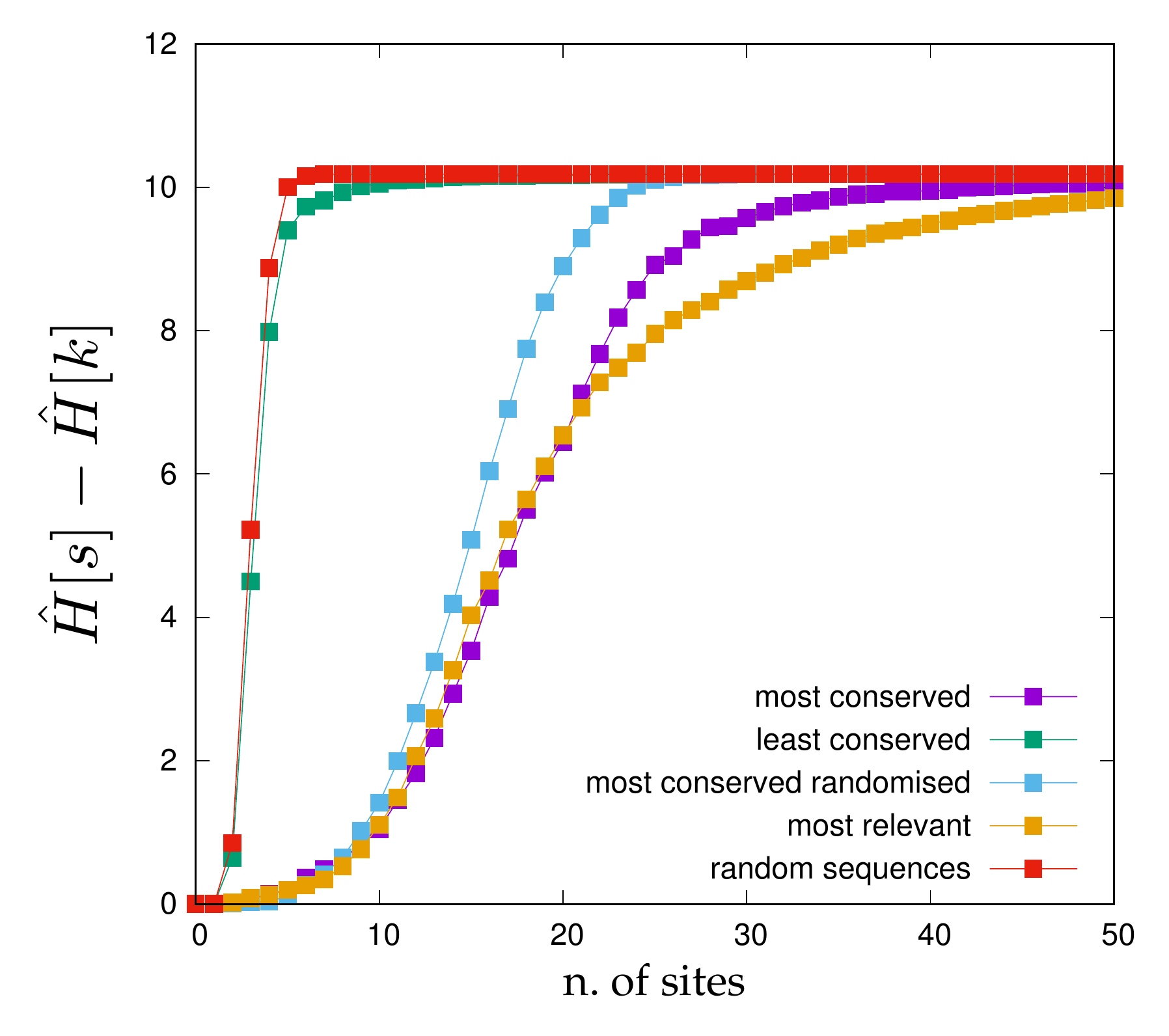}
\caption{\label{FigProt} Noise level $\hat H[\bs]-\hat H[k]$ in subsequences of $n$ amino acids in the multiple sequence alignment of the receptor binding protein domain used in Ref.~~\cite{Grigolon} (PF00072 of {\tt www.pfam.org}). \add{A multiple sequence alignment, gathers the sequences $a_1^{(i)},\ldots,a_L^{(i)}$ of amino acids for the same (orthologous) protein in different species $i=1,\ldots,N$. Each amino acid $a_\ell^{(i)}$ can take $20$ values. A subsequence is a subset $\mathcal{I}\subseteq\{1,\ldots,L\}$ of the sites of the protein and $\bs^{(i)}=\{a_\ell^{(i)},~\ell\in\mathcal{I}\}$. The most (least) conserved sites are those with smaller (larger) site entropy $\hat H[a_\ell]$. The most relevant sites are derived using the algorithm of Ref.~\cite{Grigolon}. The plot also shows the results for a randomised dataset where the amino acids that occur at a particular site are randomised across sequences. The noise level of the most conserved sites is significantly higher than that of the non-randomised dataset. This shows that statistical dependencies among the most conserved sites are significant.}}
\end{figure}

Section \ref{sec:inference} will also argue that the difference 
\begin{equation}
\hat H[\bs]-\hat H[k]\equiv \hat H[\bs|k]
\end{equation}
gives a lower bound on the number of non-informative nats in the sample $\hat \bs$, \add{and hence can be taken as a quantitative measure of the noise level}. Fig.~\ref{FigProt}, for example, reports $\hat H[\bs|k]$ for different subsets of amino acids in a database of sequences for a receptor binding domain. This shows that the noise level for the $n$ most conserved sites is smaller than that of $n$ randomly chosen sites,  or of the $n$ least conserved sites. The noise level can be further reduced by using the relevance itself to select the sites, as done in Ref.~\cite{Grigolon}.

\add{For ease of exposition, we shall first discuss the notion of relevance for a sample in Section~\ref{sec:inference}, and then turn to the analysis of the relevance for statistical models in Section~\ref{sec:learning}, following the opposite order with respect to the one 
in which the relevance has been introduced in this Section.}

\section{Statistical inference without models}
\label{sec:inference}

In Eq.~(\ref{eq:resolution_sample}) we defined the relevance of a sample. The relevance provides an upper bound to the amount of information that a sample contains on its generative model. The first argument in support of this conclusion, \blue{advanced in ~\cite{MMR}}, is that the maximum likelihood estimate that a sample provides of the unknown generative model is given by the frequency, i.e. 
\begin{equation}
\blue{\bp(\bs)}\approx \frac{k_s}{N}.
\end{equation}
\add{This suggests} to take \add{the entropy of the frequency} $\hat H[k]$ as a quantitative measure of the amount of information that a sample contains on \blue{$\bp(\bs)$.}
Cubero {\em et al.}~\cite{statcrit} refined this argument further. \add{First, on the basis the Information Bottleneck method~\cite{IB}, they observe that the relevance $\hat H[k]$ can be taken as an estimate of the mutual information between $\bs$ and $\bp$. Second, they derive the same conclusion from} the maximum entropy principle. In brief, the resolution \blue{$\hat H[\bs]$} quantifies the number of nats needed to \blue{code} a single datapoint of the sample. Using textbook relations between the joint and the conditional entropy~\cite{CoverThomas}, we observe that the resolution also equals the joint entropy $\hat H[\bs,k]=\hat H[\bs]+\hat H[k|\bs]$, because the conditional entropy $\hat H[k|\bs]=0$, in view of the fact that $k_s$ is a function of $\bs$. 
\blue{Using the same textbook relation, we can also write}
\begin{equation}
\hat H[\bs]=\hat H[\bs,k]=\hat H[\bs|k]+\hat H[k]\,.
\end{equation} 
Here, \blue{$\hat H[\bs|k]$ is the conditional entropy of the distribution $\hat p(\bs|k)=\frac{1}{km_k}\delta_{k_s,k}$}, and it measures the information content of subsamples of observations that occur the same number of times. \add{On the basis of the information given only by the sample,} we cannot differentiate between any of the states that are observed the same number of times\footnote{\add{In some cases, additional information is available in the way the variable $\bs$ is defined. For example, when $\bs$ is a protein (i.e. a sequence of amino acids) the distance between sequences provides further useful information. The definition of $\bs$ does not provide any further information when $\bs$ is just a label as, for example, the index of a cluster in data clustering.}}. Thus no information on the generative model can be extracted from these sub-samples. More formally, the distribution $\hat p(s|k)$ is a distribution of maximal entropy, that corresponds to a state of maximal ignorance. 
Therefore $\hat H[s|k]$ measures the amount of irrelevant information (or noise) in the sample. By exclusion, the information that the sample $\hat \bs$ contains on its generative model cannot exceed $\hat H[k]$. The next section offers a further derivation of this fact, drawing from an analogy with statistical inference for parametric models.

\subsection{Relevance and statistical inference}
\label{sec:stat_inf}

In order to establish a connection between the notion of relevance and the classical domain of parametric statistics, let us recall few basic facts of statistical inference. We consider the classical setup where $\hat \bs$ are $N$ independent draws from a parametric model $f(s|\theta)$, 
with $\theta\in \mathbb{R}^r$ a vector of $r$ parameters, in the limit $N\to\infty$ with $r$ fixed (the over-sampling regime).
This will serve as a reference for venturing into the case where $N$ is not so large and the model is unknown. In order to establish a precise connection, we shall ask the same question: how much information does a sample $\hat \bs$ delivers on its generative model? 

Since the model $f(\bs|\theta)$ is (assumed to be) known and what is unknown are only the parameters, the question should be rephrased as: how much information does the sample $\hat\bs$ delivers on the parameters? Suppose that, before seeing the data, our state of knowledge over the parameters is encoded by a {\em prior} distribution $p_0(\theta)$. Upon seeing the sample, the state of knowledge can be updated incorporating the likelihood of the data with Bayes rule, to obtain the {\em posterior} distribution $p(\theta|\hat\bs)$. 
The amount of information that can be gained from the sample on its generative model $f(s|\theta)$ can then be quantified by in terms of the Kullback-Leibler divergence $D_{KL}$ between the posterior and the prior distribution of $\theta$. When $N\gg r$, as we will show in~\ref{app:param}, this is asymptotically given by
\begin{eqnarray}
D_{KL}\left[p(\theta|\hat \bs)||p_0(\theta)\right]&=& \int d\theta p(\theta|\hat \bs) \log \frac{p(\theta|\hat \bs)}{p_0(\theta)}\nonumber \\ 
&\simeq& \frac{r}{2}\log\frac{N}{2\pi e}+\frac 1 2 \log{\rm det}\, \hat L(\hat \theta)-\log p_0(\hat\theta)+O(1/N.)
\label{eq:MI}
\end{eqnarray}
{where $\hat\theta$ is the value of $\theta$ that maximises the likelihood of the data $f(\hat \bs|\theta)$, i.e. the {\em maximum likelihood} estimate for the parameters, and $\hat L(\hat \theta)$ is the Hessian of the log-likelihood per data point, at this maximum likelihood estimate. The choice of $D_{KL}$, as opposed to another measure of distance can be justified by observing that the expected value of $D_{KL}\left(p(\theta|\hat \bs)||p_0(\theta)\right)$ over the marginal distribution of the data $p(\hat \bs)=\int d\theta f(\hat \bs|\theta)p_0(\theta)$, coincides with the mutual information 
\begin{equation}
\label{IDKL}
I(\hat \bs,\theta) =\E{D_{KL}\left[p(\theta|\hat \bs)||p_0(\theta)\right]}
\end{equation}
between the data and the parameters $\theta$. 

Note that the total information contained in the sample\footnote{\add{I.e. the minimal number of nats needed to encode the data.}}, $N\hat H[\bs]$, is proportional to $N$, but the leading term, $\frac r 2 \log N$, in Eq.~(\ref{eq:MI}) grows only logarithmically with $N$. Hence only a very small fraction of the information contained in the sample is informative about the generative model. In addition, \add{the leading term} only depends on the model through the number of parameters and it is independent of the sample. So this term gives no information on whether the sample is informative or not. If the prior is chosen in a judicious manner\footnote{In order to see this, we need to discuss the content of the sub-leading terms in Eq.~(\ref{eq:MI}). The first and the second term account for the reduction in the  uncertainty $\delta\theta$ on the parameters. The first states that the reduction is of order $1/\sqrt{N}$ and the second that it depends on how sharply peaked around its maximum $\hat \theta$ the posterior distribution on $\theta$ is. The second term may be small when small eigenvalues of the Hessian occur, which correspond to directions in parameter space where the posterior is flat -- the so-called {\em sloppy modes}~\cite{sloppy}. A broad posterior distribution of $\theta$ is a signature of overfitting and it suggests that the modeller didn't do a good job in choosing the model. The last term informs us that we learn more if the parameters {\em a posteriori} turn out to attain values $\hat\theta$ that are very unlikely {\em a priori}. 
A reasonable modeller would not choose a prior such that values $\hat \theta$ for which the statistical errors are small are very unlikely. Indeed, the sum of the last two terms is a constant in principled approaches to statistical inference in which the model is an exponential family and the prior is uninformative (see e.g.~\cite{Myung}). We note in passing, that Ref. \cite{mastromatteo} remarks that when the model is chosen appropriately, the second term in Eq.~\eqref{eq:MI} should be large. This corresponds to parameters $\hat\theta$ which are close to a {\em critical point} in a statistical physics analogy, in agreement with the general relation between efficient representations and criticality that we shall discuss in the sequel.}, one can argue that also the sum of the second and third terms of Eq.~(\ref{eq:MI}) is independent of the sample. This leads us to the conclusion that all typical samples are equally informative in the over-sampling regime, and then $D_{KL}\left[p(\theta|\hat \bs)||p_0(\theta)\right]\simeq I(\hat\bs,\theta)$.

In order to move towards settings which are closer to the under-sampling regime, let us address the same question as above, in the case where the model is unknown. Then one can resort to Bayesian model selection~\cite{Myung}, in order to score the different models according to their probability conditional to the observed data. If each model is a priori equally likely, models can be scored according to their log-evidence, which is the probability of the data according to model $f$ (see~\ref{app:param}). In the asymptotic regime for $N\to\infty$, this reads
\begin{equation}
\label{eq:BMS}
\log P(\hat\bs|f)\simeq \log f(\hat \bs|\hat\theta) - D_{KL}\left[p(\theta|\hat \bs)||p_0(\theta)\right] -\frac r 2.
\end{equation}
\add{This equation states that} the log-likelihood, $\log f(\hat \bs|\hat\theta)$, of a model should be more penalised the more information the data delivers on the parameters. 
In other words, among all models under which the data is equally likely, those that deliver the least information on the parameters should be preferred. This agrees with the minimum description length principle~\cite{MDL} that prescribes that the model that provides the shorter description of the data should be preferred. 

When the information about the generative model is totally absent, one should in principle compare all possible models. This is practically unfeasible, apart from special cases~\cite{HM,MCM}, because the number of models that should be considered grows more than exponentially with the dimensionality of the data. {In addition, if the number $N$ of observations is not large enough, the posterior distribution over the space of models remains very broad. Even in the case when the data were generated from a parametric model $f$, Bayesian model selection may likely assign a negligibly small probability to it. In the under-sampling regime, the posterior on the space of models will be dominated by models that are simpler than the true model, whose number of parameters is smaller than that of the true model. 

Still, whatever the most likely model may be, we can rely on Eq.~(\ref{IDKL}) in order to estimate information learned from the data, precisely because it does not depend on the data. In addition, for samples generated from parametric models, the information content of the sample can be traced back to a set of relevant variables, the {\em sufficient statistics} $\phi(\bs)$. These are observables such that the values $\hat\phi$ \add{that the empirical averages of $\phi$} take on a sample $\hat \bs$ are sufficient to estimate the model's parameter. In other words, all the information a sample provides on the parameters is contained in $\hat\phi$, i.e. $I(\hat \bs,\theta)=I(\hat\phi,\theta)$~\cite{CoverThomas}. 

In addition, whatever the most likely model is, and whatever its sufficient statistics are, the frequencies $\vec k=\{k_{\bs},~\bs\in\mathcal{S}\}$ serve as sufficient statistics. {Indeed, under the assumption that $\bs^{(i)}$ are drawn independently from  a distribution $\bp(\bs)$, $\vec k$ satisfies the Fisher-Neyman factorisation theorem~\cite{CoverThomas}, because the conditional distribution of $\hat \bs$, given $\vec k$, is independent of $\bp$. So even if the model and, as a consequence the sufficient statistics, are unknown, we have that $I(\hat \bs,\theta)=I(\vec k,\theta)$.
This brings us to the chain of relations
\begin{equation}
\label{bound1}
I(\hat \bs,\theta)=I(\vec k,\theta)\le H[\vec k]\le NH[k]\approx N \hat H[k].
\end{equation}
The first inequality results from the definition $I(\vec k,\theta)\le H[\vec k]-H[\vec k|\theta]$ and the fact that $H[\vec k|\theta]\ge 0$. In the second, $H[k]$ is the entropy of the probability $P\{k_{\bs^{(i)}}=k\}$ that a randomly chosen point $s_i$ of \add{a random} sample \add{of $N$ points} occurs $k$ times in $\hat \bs$. 
The inequality derives from the fact that frequencies of different sampled points are not independent variables\footnote{In addition, a frequency profile $\vec k=\{k_{\bs},~\bs\in\mathcal{S}\}$ typically corresponds to more than one frequency sequence $\hat k=(k_{\bs_1},\ldots, k_{\bs_n}\}$, so $H[\vec k]\le H[\hat k]\le NH[k]$.}. Both inequalities in Eq.~(\ref{bound1}) are not tight at all in the over sampled regime $N\gg r$, because Eq. (\ref{eq:MI}) implies that $I(\hat \bs,\theta)$ increases only logarithmically with $N$. Eq.~(\ref{bound1}) is informative in the under-sampling regime where the number of sampled points $N\sim r$ is of the order of the number of parameters. 

Finally, the approximation $H[k]\approx \hat H[k]$ estimates the entropy of $k$ from a single sample. {It is well known~\cite{miller1955note,entropy_est} that the entropy of the empirical distribution is a biased estimate of the entropy of the true distribution. Due to convexity, $\E{\hat H[k]}\le H[k]$ with a bias 
$H[k]-\E{\hat H[k]}$ which is of order $1/N$~\cite{entropy_est}. Statistical errors on $\hat H[k]$ may be much larger than the bias, which confers to the bound~(\ref{bound1}) an unavoidable approximate status. Several methods have been proposed to improve entropy estimation in the under-sampling regime~\cite{entropy_est,NemenmanCoinc}. These could be used to improve the bound (\ref{bound1}) for quantitative analysis. This goes beyond the scope of this Section, which is that of illustrating the significance of the relevance $\hat H[k]$ in relation to statistical inference. 
In this respect, the interpretation of  Eq.~(\ref{bound1}) is that {\em samples with a small value of $\hat H[k]$ likely come from simple models, whereas samples with a large value of $\hat H[k]$ may be generated by models with a rich structure}.

\subsection{An approximate bound on the number of inferable parameters}
\label{sec:bound}

Taking $N\hat H[k]$ as an upper bound on the information $I(\hat \bs,\theta)\simeq D_{KL}\left(p(\theta|\hat \bs)||p_0(\theta)\right)$
that a sample delivers on the parameters of the generative model, and combining it with Eq.~(\ref{eq:MI}), allows us to derive an upper bound to the complexity of models that can be inferred from a dataset $\hat \bs$. In particular, keeping only the leading term $I(\hat \bs,\theta)\approx \frac r 2 \log N$, leads to an approximate upper bound
\begin{equation}
\label{boundd}
r\le 2\frac{\hat H[k]}{\log N}N
\end{equation}
on the number of parameters that can be estimated from the sample $\hat \bs$. It is important to stress that this is not a bound on the number of parameters of the unknown generative model $\bp$ but on the number of parameters of the model that best describes the sample in a Bayesian model selection scheme. Ref.~\cite{HM} discusses in detail Bayesian model selection in the class of Dirichelet mixture models and it shows that the number of parameters\footnote{\add{A Dirichelet mixture model decomposes the set $\mathcal{S}$ of states into $Q$ subsets $\mathcal{Q}_q$ of equiprobable states, i.e. $p(\bs)=\mu_q$ for all $\bs\in\mathcal{Q}_q$, where $\mu_q\ge 0$ are the parameters of the Dirichelet's model. Because of the normalisation constraint $\sum_{q=1}^Q|\mathcal{Q}_q|\mu_q=1$, the number of parameters is $Q-1$.}} of the optimal model increases with the relevance $\hat H[k]$. This class of models provides a simple intuition for the relation between the relevance and the number of parameters that can be estimated. In brief, if two states $\bs$ and $\bs'$ occur in the sample with very different frequencies $k_{\bs}$ and $k_{\bs'}$, then models that assume they occur with the same frequency $\bp(\bs)=\bp(\bs')$ will very unlikely be selected in Bayesian model selection. Conversely, if $k_{\bs}=k_{\bs'}$ the most likely models will be those for which $\bp(\bs)=\bp(\bs')$. As a consequence, the larger is the variation of $k_{\bs}$ in the sample, as measured by $\hat H[k]$, the larger will the number of parameters be. 

The bound (\ref{boundd}) is not very informative in the over sampled regime. For example, in the data used in Ref.~\cite{BialekUSSC} on the voting behaviour of the nine justices of the US Supreme Court ($N=895$), Eq.~(\ref{boundd}) bound predicts $r\le 626$ which is an order of magnitude larger than the number of parameters of a model with up to pairwise interactions among the judges ($r=45$), that Lee {\em et al.}~\cite{BialekUSSC} argue describes well the dataset. Instead, in the case of contact predictions from datasets of protein sequences, models based on pairwise interactions, such as those in Ref.~\cite{morcos}, may need orders of magnitude more parameters than what the bound (\ref{boundd}) allows\footnote{\label{protparam} For the Sigma-70 factor (Pfam ID PF04542), using all sequences available on {\tt http://pfam.org} ($N=105709$ as of Feb. $15^{\rm th}$ 2021), Eq.~(\ref{boundd}) prescribes $r\le 20659$, which is much smaller than the number ($r\approx 3\cdot 10^7$) of parameters in a pairwise interacting model on sequences of $L=394$ amino acids, each of which may be of $20$ different types~\cite{morcos}.}. 

Let us comment on the extreme under-sampling regime, where each outcome $\bs$ is observed at most once, and\footnote{Note that this is the limit where the approximation $\hat H[k]\approx H[k]$ becomes uncontrollable, as discussed e.g. by Nemenman~\cite{NemenmanCoinc}.} $\hat H[k]=0$. In the absence of prior information, Eq.~(\ref{boundd}) predicts that the only generative model consistent with the data is the one with no parameters, $d=0$. Under this model, all outcomes are equally likely. Conversely, in the presence of {\em a priori} information, complex models can be identifies even when $\hat H[k]=0$. For example, the literature on graphical model reconstruction~\cite{nguyen2017inverse} shows that when $\bs=(\sigma_1,\ldots,\sigma_n)$ is a vector of spin variables \hbox{$\sigma_i=\pm 1$} that is known to be generated from a model with low order interactions, it is possible to infer the network of interactions with very limited data. If the class of models is sufficiently constrained, Santhanam and Wainwright show that $N\sim\log n$ data points may be sufficient to identify the generative model~\cite{santhanam2012information}. 

One way to deal with the extreme under-sampling regime where $\hat H[k]=0$ is to resort to dimensional reduction schemes. These correspond to a transformation $\bs\to\bs'$ of the variables (e.g. coarse graining) such that the new sample $\hat\bs'$ has a lower resolution $\hat H[\bs']$, so that $\hat H[k]>0$. The relevance $\hat H[k]$ can then provide a quantitative measure to score different dimensional reduction schemes. Section~\ref{sec:dataclustering} discusses the example of data-clustering. As a further example, in the case where $\bs=(\sigma_1,\ldots,\sigma_n)$ is the configuration of $n$ discrete variables $\sigma_i$, the resolution can be reduced by considering only a subset $\mathcal{I}$ of the variables, i.e $\bs'=(\sigma_i~i\in \mathcal{I})$. Then the corresponding relevance $\hat H_{\mathcal{I}}[k]$ can be used to score the selected variables $i\in\mathcal{I}$. This idea was explored in Grigolon {\em et al.}~\cite{Grigolon} in order to identify relevant positions in protein sequences (see Fig.~\ref{FigProt}). In principle, by considering small sets $\mathcal{I}$ of variables, it may be possible to estimate efficiently a statistical model; \blue{see e.g. \cite{dunn2013learning, battistin2017learning}}. This suggests that by considering all possible subsets $\mathcal{I}$ of variables, it may be possible to reconstruct the whole generative model. The problem with this approach is that, in general, this procedure does not generate consistent results, unless the sufficient statistics of the generative model satisfy special (additivity) properties\add{, as discussed in Ref.}~\cite{shalizi2013consistency}.



\subsection{The relevance and the number of distinguishable samples}

For the same value of $\hat H[\bs]$, samples with higher relevance, are those in which more states have differing frequencies and thus are more distinguishable in this sense.  In order to provide further intuition on the meaning of relevance, it is useful to take an ensemble perspective, as e.g. in~\cite{tikochinsky1984alternative}. 

There are three levels of description of a sample. The finest is given by the sample $\hat \bs$ itself, which specify the outcome $\bs^{(i)}$ of the $i^{\rm th}$ sample point for all $i=1,\ldots,N$. A coarser description is given by the frequency profile $\vec k=\{k_{\bs},~\bs\in\mathcal{S}\}$, which specifies the number of times an outcome $\bs$ occurs in the sample. It is a coarser description because it neglects the order in which the different outcomes occur, which is irrelevant information in our setting. An even coarser description is given by the degeneracy profile $\vec m=\{m_k,~k>0\}$, that specifies the number $m_k$ of outcomes $\bs$ that occur $k$ times. This is equivalent to describing a sample by its frequency sequences $\hat k=(k_{\bs_1},\ldots,k_{\bs_N})$. In this description, the distinction between outcomes $\bs$ and $\bs'$ that occur the same number of times ($k_{\bs}=k_{\bs'}$) is lost. 
Only the distinction of outcomes by their frequency is considered relevant.

When all samples $\hat \bs$ of $N$ observations are {\em a priori} equally likely, intuition can be gained from simple counting arguments. Let us denote the total number of samples with a given frequency profile $\vec k=\{k_{\bs},~\bs\in\mathcal{S}\}$ by $W_{\bs}$. This is given by the number of ways in which the variables $\bs^{(i)}$ can be assigned so that Eq.~(\ref{eq:ks}) can be satisfied 
\begin{equation}
W_{\bs}=\frac{N!}{\prod_{\bs} k_{\bs}!}=\frac{N!}{\prod_k(k!)^{m_k}}\sim e^{N\hat H[\bs]}
\end{equation}
where the last step, that relates $W_{\bs}$ to the resolution, holds for large $N$, and it relies on a trite application of Stirling's formula $n!\simeq n^n e^{-n}$. The observation of $\hat \bs$ allows us to discriminate between $W_{\bs}$ possible samples. 

Likewise, the number of frequency sequences $\hat k=(k_{\bs_1},\ldots,k_{\bs_N})$ with the same degeneracy profile $\vec m$ is given by 
\begin{equation}
W_{k}=\frac{N!}{\prod_k(km_k!)}\sim e^{N\hat H[k]}\,,
\end{equation}
which is the number of ways to assign $k_{\bs^{(i)}}$ for each $i$, such that Eq.~(\ref{eq:mk}) is satisfied.
The observation of a frequency sequence $\hat k=(k_{\bs_1},\ldots,k_{\bs_N})$ allows us to discriminate between $W_{k}$ possibilities. 
The number of samples with the same frequency sequence $\hat k$ is the ratio between these two numbers 
\begin{equation}
W_{\bs|k}=\frac{W_{\bs}}{W_k}=\prod_k\frac{(km_k)!}{(k!)^{m_k}}\sim e^{N\hat H[\bs|k]}
\end{equation}
where again the last relation holds for large $N$. These samples cannot be distinguished based on the sequence of frequencies $\hat k$. At fixed $\hat H[\bs]$, $W_{\bs|k}$ clearly decreases when $\hat H[k]$ increases. In other words, for the same value of $\hat H[\bs]$, 
samples with higher relevance, are those for which the observed frequencies have a higher discriminative power. 

\begin{figure}[ht]
\centering
\includegraphics[width=0.8\textwidth,angle=0]{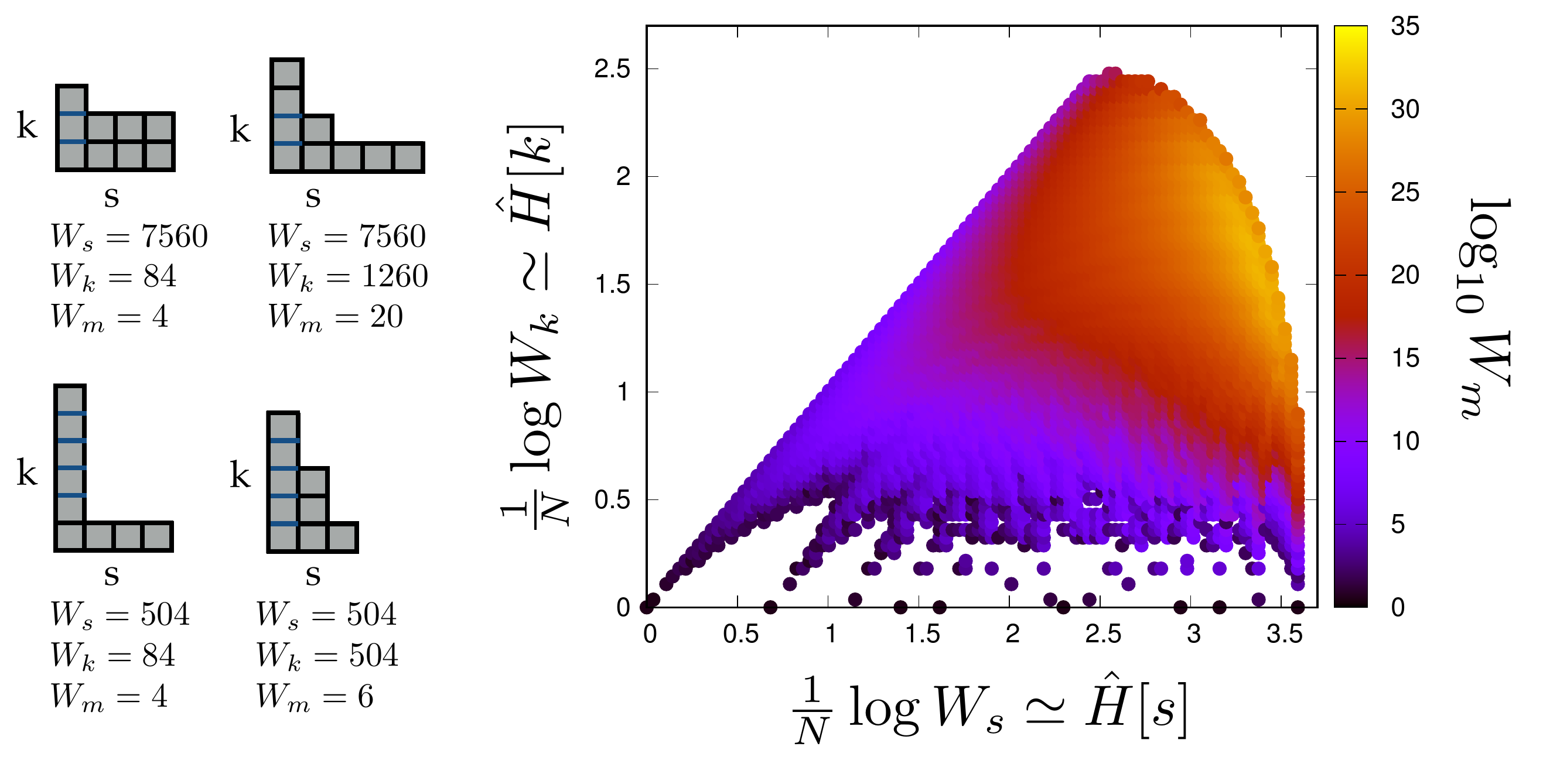}
\caption{\label{fig:n_dist} Number $W_m$ of different distributions $\vec k$ as a function of $\hat H[\bs]$ and $\hat H[k]$. Left: examples of frequency profiles corresponding to the same number of samples ($W_{\bs}$) with different values of $W_k$ and $W_m$ for $N=9$. In both the top and the bottom cases, a higher value of $W_k$ is associated with a larger number ($W_m$) of possible distributions. Right: density plot of $W_m$ as a function of resolution and relevance, for $N=100$.}
\end{figure}

For given values $\hat H[\bs], \hat H[k]$ of the resolution and of the relevance, there can be a number $W_m$ of different profiles $\vec k$ consistent with them. By observing a sample $\hat \bs$ we effectively discriminate between these, hence $\log W_m$ provides a quantitative measure of the uncertainty on the generative model $\bp(\bs)\approx k_{\bs}/N$ that the sample resolves. 
For small $N$, $W_m$ is simply given by the number of ways the $M=\sum_k m_k$ sampled states can be distributed in the frequency classes
\begin{equation}
W_m=\frac{M!}{\prod_k m_k!}.
\end{equation}
Fig.~\ref{fig:n_dist}(left) illustrates the relation between $W_{\bs},W_k$ and $W_m$ in some examples with $N=9$. This figure suggests that, at equal resolution ($W_{\bs}$), higher relevance ($W_k$) is associated with a larger number of distributions $W_m$. For large $N$, many degeneracy profiles $m_k$ correspond to the same values of $W_{\bs}$ and $W_k$. This makes the calculation of $W_m$ more complex. In brief, different degeneracy profiles $m_k$ correspond to different integer partitions of $N$. It is possible to generate all integer partitions with efficient algorithms \cite{nijenhuis2014combinatorial} for moderate values of $N$, and to compute $W_{\bs},W_k$ and $W_m$ for each partition. Fig.~\ref{fig:n_dist}(right) shows the number $W_m$ of different distributions, for different values of $\hat H[\bs]$ and $\hat H[k]$, for $N=100$. This suggests that,  in the under-sampling regime where $\hat H[\bs]$ is close to $\log N$, large values of $\hat H[k]$ at the same resolution correspond to regions with a high density of distinguishable frequency distributions. A sample in this region has a higher discriminative power, because it can single one out of a larger number of distributions. 

This analysis is similar in spirit to that of Myung {\em et al.}~\cite{Myung}, who count the number of distinguishable distributions for parametric models. In that context, the number of distributions that can be distinguished based on a sample (which would be the analog of $W_m$) is related to the stochastic complexity of the model. Loosely speaking, this suggest that we can think of the relevance as an easily computable proxy for the complexity of the generative model, in the case where the latter is unknown.

\subsection{Resolution, relevance and their trade-off: the case of data clustering}
\label{sec:dataclustering}

In practice, one always choses the variable $\bs$ that is measured or observed. For instance, one can measure a dynamical systems at different temporal resolutions\blue{\footnote{An example is the case of spiking neurons that will be discussed in detail in Section \ref{sec:neuro}.}}. Each of these choices corresponds to a different level of resolution $\hat H[\bs]$. In other words, $\hat H[\bs]$ is an independent variable. At the highest resolution all sample points $\bs^{(i)}$ are different and \hbox{$\hat H[\bs]=\log N$}. At the lowest resolution they are all equal and \hbox{$\hat H[\bs]=0$}. For each choice, the value of $\hat H[k]$ is a property of the dataset, i.e. it is a dependent variable. When different levels of resolution can be chosen, the relevance traces a curve in the $(\blue{\hat H[\bs]},\hat H[k])$ plane that encodes the tradeoff between resolution and relevance.

As an illustrative example of the tradeoff between resolution and relevance, this Section discusses the case of data clustering.
Data clustering deals with the problem of classifying a dataset \blue{$\hat x=(\vec x^{(1)},\ldots,\vec x^{(N)})$} of $N$ points $\vec x^{(i)}\in \mathbb{R}^d$ into $K$ clusters. \add{Each point $\vec x^{(i)}$ is specified by the value of $d$ features. For example, the paradigmatic case of the iris dataset~\cite{fisher1936use} contains $N=150$ samples of iris flowers from three different species, each of which is represented by a vector  $\vec x$ of $d=4$ characteristics (the length and the width of the sepals and petals). We refer to Refs.~\cite{gan2020data,sikdar2020unsupervised} for further examples.}

The aim of data clustering is to group points that are ``similar" in the same cluster, distinguishing them from those that are ``different". Each cluster is identified by a label $\bs$ that takes integer values $\bs=1,\ldots,K$. 

Each data-clustering scheme needs to specify {\em a priori} a notion of similarity between points and between clusters of points. Given this, the data clustering algorithm assigns to each data point \blue{$\vec x^{(i)}$ a label $\bs^{(i)}$} that specifies the cluster it belongs to. This transforms the data into a dataset $\hat\bs$ of labels, where the frequency $k_{\bs}$ of an outcome corresponds to the size of the corresponding cluster $\bs$. The resulting cluster structure depends on the algorithm used and the similarity metrics that is adopted. A plethora of different algorithms have been proposed~\cite{gan2020data}, which raises the question of how the most appropriate algorithm for a specific dataset and task should be chosen. We shall see what the notion of relevance can contribute in this respect.

\begin{figure}[ht]
\centering
\includegraphics[width=\textwidth,angle=0]{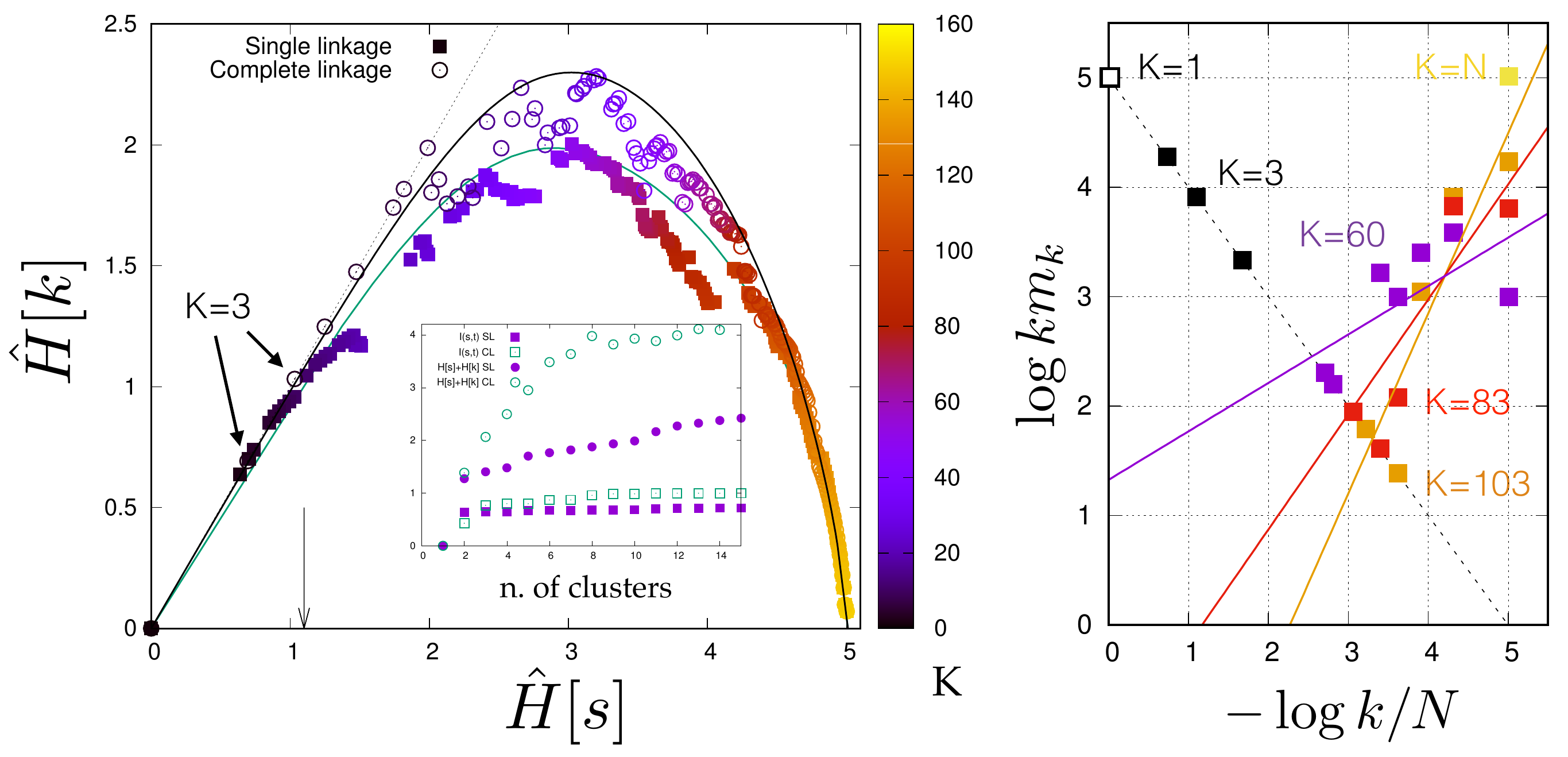}
\caption{\label{fig_iris} {\bf Left}: Relevance as a function of resolution for the classification of the iris dataset into different clusters. Each point corresponds to the cluster structure with a given number of clusters. Different symbols refer to single ($\blacksquare$) or complete  ($\odot$) linkage hierarchical clustering algorithms. In single (complete) linkage the distance between clusters $\bs$ and $\bs'$ is the minimum (maximum) euclidean distances between points in cluster $\bs$ and points in cluster $\bs'$. The arrow marks the resolution $\log 3$ of the true classification (three species of iris flowers, each present with $50$ samples). The black full line is a theoretical lower bound for the maximal $\hat H[k]$ as a function of $\hat H[s]$ (see \cite{HM}) and the green line corresponds to a random cluster distribution~\cite{statcrit}. Since $k_{\bs}$ is a function of $\bs$, the data processing inequality implies that $\hat H[k]\le \hat H[\bs]$. When $\hat H[k]= \hat H[\bs]$ all states are sampled a different number of times (i.e. $k_{\bs}\neq k_{\bs'}$ for all $\bs\neq \bs'$).
The inset reports the mutual information $I(\bs,\bt)$ between the cluster labels and the true classification $\bt$, as well as the total information $\hat H[\bs]+\hat H[k]$, as a function of $K$ for the two algorithms. {\bf Right}: Tradeoff between the degeneracy $\log(k m_k)$ and the precision $\log (k/N)$ in the cluster structures obtained by the complete linkage algorithm for the iris dataset. Cluster structures for $K=103,83,60$ and $3$ clusters are shown with a colour code corresponding to the left panel. Linear fits of the relation $\log(k m_k)\simeq \mu \log (k/N)+c$ are also shown. The values of the slopes are $\mu\simeq 1.65,1.05$ and $\mu=0.44$ for $K=103,83$ and $60$ respectively.}
\end{figure}

For each value $K$ of the number of clusters, the values of $\hat H[\bs]$ and $\hat H[k]$ can be computed from $k_{\bs}$, using Eqs.~(\ref{eq:resolution_sample},\ref{eq:relevance_sample}). The resolution $\hat H[\bs]$ can be adjusted by varying the number of clusters $K$. 
Fig.~\ref{fig_iris} illustrates the tradeoff between resolution and relevance for the case of the iris dataset~\cite{fisher1936use}. It shows the trajectories that the point $(\hat H[\bs],\hat H[k])$ traces as $K$ varies from $N$ to $1$, for two popular data-clustering algorithms (see caption).

The over sampled regime corresponds to the region on the extreme left of the graph, where the number of clusters is small and the number of points in each cluster is large ($k_{\bs}\propto N$). This is the region of interest in most applications of data clustering. Indeed,  the purpose of data clustering is to provide an interpretation of how the data is organised  or to reveal the structure of an underlying ground truth. At low resolution we expect the cluster labels to ``acquire'' a meaning that reflects recognisable properties. The meaning of labels fades as we move into the under sampling regime. On the right side of the plot, cluster labels are purely featureless theoretical constructs, with no specific meaning.

The tradeoff between resolution and relevance can be discussed assuming that the labels $\bs$ have been drawn from an unknown distribution $\bp$. As we move to lower an lower values of $\hat H[\bs]$ we acquire more and more information on $\bp$, and this can be quantified by $\hat H[k]$. In the under-sampling regime a decrease by one bit in resolution results in a gain in relevance that is related to the slope of the curve in Fig.~\ref{fig_iris}. The total information $\hat H[\bs]+\hat H[k]$ has a maximum at the point where the tangent to the curve has slope  $-1$. Further compression (i.e. decrease in $\hat H[\bs]$) beyond this point comes with an increase in $\hat H[k]$ that does not compensates for the loss in $\hat H[\bs]$. 
Considering clustering a way of compressing the data, this point marks the boundary between the lossless and the lossy compression regimes. 

The same tradeoff can be discussed at the level of the cluster structures $\mathcal{C}$, refining arguments from~\cite{MMR,SMJ}. Let us take the iris dataset example: 
At the coarsest resolution ($\hat H[\bs]=0, K=1$) all the $\vec x_i$ are just iris flowers, no information that distinguishes $\vec x_i$ from some other sample point is retained in $\mathcal{C}$. At the other extreme ($\hat H[\bs]=\log N, K=N$) each $\vec x_i$ is distinguished as being a different iris flower. At this level, $\log N$ nats are needed to identify each point within the sample. Of this information, at intermediate resolution $\hat H[\bs]$ ($1<K<N$), the information that is retained about a point $\vec x_i$ that belongs to cluster $\bs$ is $-\log({k_{\bs}}/{N})$ nats. The rest is assumed to be noise, i.e. irrelevant details, by the cluster algorithm. The classification $\mathcal{C}$ retains more details about points in small clusters ($-\log ({k_{\bs}}/{N})> \hat H[\bs]$) than about those in larger clusters ($-\log ({k_{\bs}}/{N})< \hat H[\bs]$). The way in which the cluster algorithm allocates points $\vec x_i$ to different levels of detail $k$ depends on the algorithm used and on the dataset. Abstracting from the specific algorithm, we expect that 
in the under-sampling regime small clusters should be more numerous than large clusters. In order to see this, let us analyse how a cluster algorithm should assign different points to different frequency classes $k$ (i.e. cluster sizes). We first observe that $-\log(k/N)$ measures the variability of points in clusters of size $k$. Points $\vec x_i$ and $\vec x_j$ belonging to \add{the same} small cluster differ more than points that belong to larger ones. Put differently, points in larger clusters share similar features with more other points than points in smaller clusters. This distinction should be reflected in the number $\log (k m_k)$ of nats needed to identify a point that belongs to a cluster of size $k$, which can be taken as a measure of the noise. Therefore we expect a positive dependence between noise, quantified by $\log (k m_k)$, and variability, quantified by $-\log(k/N)$. 
Fig.~\ref{fig_iris} supports this argument. Approximating this dependence with a linear behaviour, we see that the slope $\mu$ of this relation decreases as one approaches the over sampled regime. In the over sampled regime, when all clusters have different sizes ($m_k\le 1~\forall k$), all clusters align on the line where $\log (km_k)-\log(k/N)=\log N$. On this line the allocation of nats into different classes of precision $k$ is such that, for each point $\vec x_i$ the precision $-\log(k/N)$ equals the initial information content of each point $(\log N)$ minus the noise $\log (km_k)$.

\begin{figure}[ht]
\centering
\includegraphics[width=0.8\textwidth,angle=0]{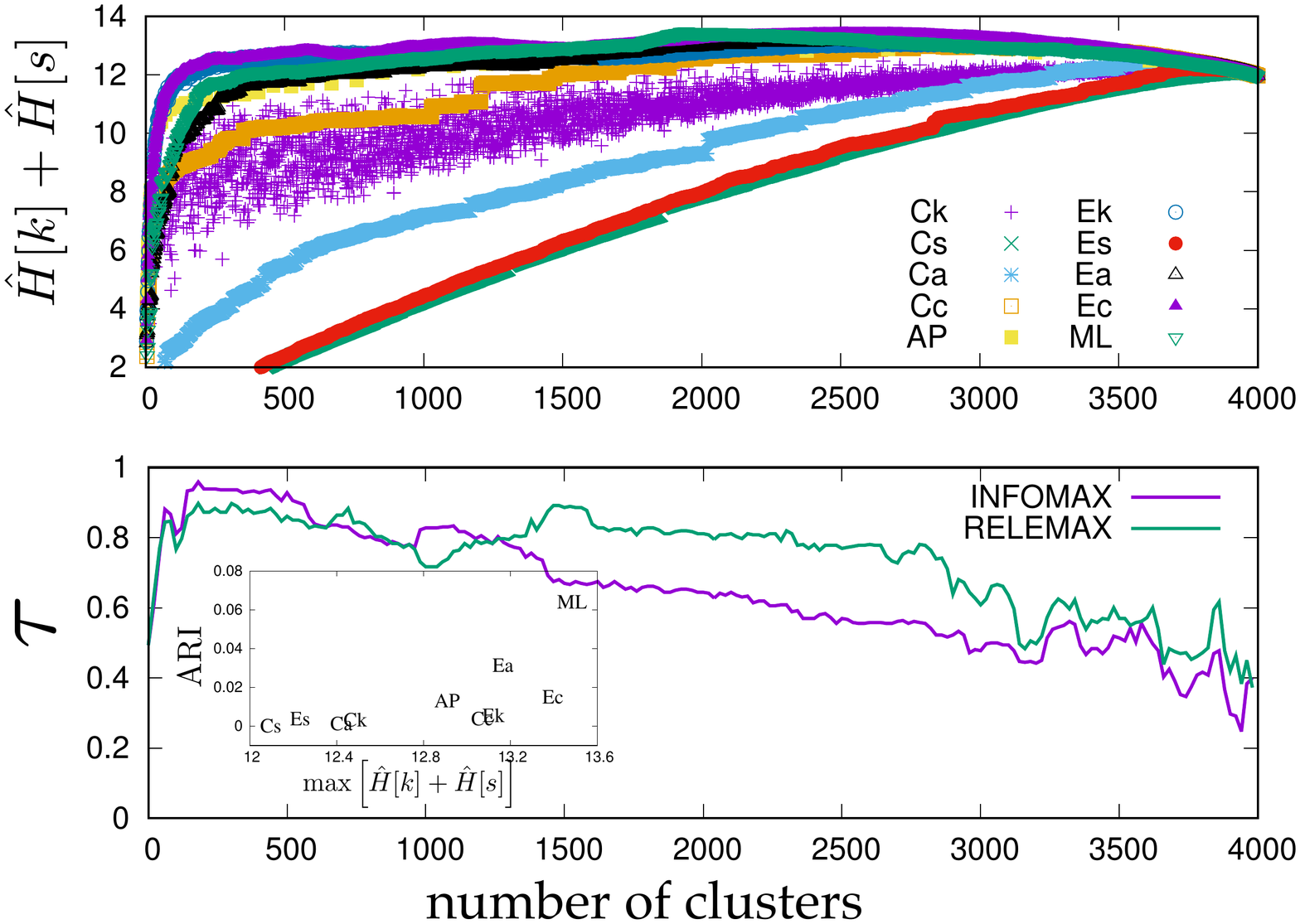}
\caption{\label{cluster_stocks} Data clustering of $N=4000$ stocks in the New York stock market, based on $d = 2358$ daily returns in the period 1 January 1990 to 30 April 1999. The data and the algorithms are the same used in Ref.~\cite{sikdar2020unsupervised}. The algorithms include single (s), average (a) and complete (c) linkage hierarchical algorithms based on $L_2$ (E) or $L_1$ (C) metric (e.g. Es correspond to single linkage with $L_2$ metric), Affinity Propagation (AP)~\cite{frey2007clustering} and the algorithm of Ref.~\cite{giada2001data} (ML).
Top: $\hat H[\bs]+\hat H[k]$ as a function of the number of clusters $K$ for the different algorithms. Bottom: Kendall-$\tau$ correlation between the distance of the different cluster structures with $K$ clusters to the ground truth and the ranking of algorithms based on $\hat H[\bs]$ (INFOMAX) and on $\hat H[\bs]+\hat H[k]$ (RELEMAX). The distance to the ground truth, as in Sikdar {\em et al.}~\cite{sikdar2020unsupervised}, is based on the majority ranking of three different cluster \add{metrics} (purity, Adjusted Rand index and normalised mutual information, see \cite{sikdar2020unsupervised} for more details). Inset: Adjusted Rand index between the ground truth and the cluster structure with $K^*={\rm arg}\max_K \hat H[\bs]+\hat H[k]$ clusters, for the different algorithms, as a function of the maximal value of $\hat H[\bs]+\hat H[k]$. Note that each cluster algorithm attains its maximum of $\hat H[\bs]+\hat H[k]$ at different values of $K^*$. The Adjusted Rand index is a measure that allows for a comparison between distances in these cases. The positive dependence suggests that the algorithm that reaches closest to the ground truth at $K=K^*$ is the one that achieves a maximal value of $\hat H[\bs]+\hat H[k]$.}
\end{figure}

This suggests that relevance can be used to compare different data clustering algorithms in a  unsupervised manner. In the example of Fig.~\ref{fig_iris} this conclusion can be validated by comparing the cluster structures to the ground truth classification of the flower specimen in the three different species. In this case, the algorithm with a higher relevance curve generates the classification in $K=3$ clusters which is closest to the ground truth. However the situation is more complex than the case discussed in Fig.~\ref{fig_iris} suggests. Indeed, Sikdar {\em et al.}~\cite{sikdar2020unsupervised} found that the best predictor of the algorithm that generates a cluster structure which is closer to the ground truth is not the relevance $\hat H[k]$. Analysing a large variety of datasets and algorithms, Sikdar {\em et al.}~\cite{sikdar2020unsupervised} concluded that the best way to rank algorithms in an unsupervised manner is by their resolution $\hat H[\bs]$. Specifically, the algorithm that generates a cluster structure with a number of clusters $K$ equal to that of the ground truth, which is closest to the ground truth is (most likely) the one with the largest value of $\hat H[\bs]$. This agrees with Linsker's INFOMAX\footnote{Some intuition on this result can be gained by considering a clustering algorithm as a translator of the data in an alphabet of $K$ letters. The algorithm that generates the richest description (i.e. largest $\hat H[\bs]$) is the one which retains more information on the ground truth.} principle~\cite{linsker1988self}, which is also at the basis of other dimensional reduction schemes such as Independent Component Analysis~\cite{bell1995information}. In loose terms, INFOMAX\footnote{In Linsker's words, this organizing principle favours architectures that ``maximize the amount of information that is preserved when signals are transformed at each processing stage''.} rewards algorithms that compress the data as slowly as possible~\cite{linsker1988self}. 

Yet, it is important to remark that this result is based on comparing cluster structures with the same number of clusters $K$ of the ground truth and that the typical values of $K$ in all cases fall in the over-sampled regime. In this regime $\hat H[k]$ is not informative because in most cases it is constrained by the data processing inequality to be close to $\hat H[\bs]$. 

Fig.~\ref{cluster_stocks} suggests that the situation is different when the comparison between cluster algorithms is done in the under-sampling regime.
Fig.~\ref{cluster_stocks} reports the behaviour of ten different clustering algorithms on a dataset of financial returns of $N=4000$ different stocks (see caption). The lower panel reports the same analysis as in~\cite{sikdar2020unsupervised}, where the comparison with the ground truth is performed for different values of $K$ (see caption). This suggests that when clustering algorithms are compared in the under-sampling regime (large $K$), the best predictor of the optimal algorithm is $\hat H[\bs]+\hat H[k]$. This captures not only the information content of the representation ($\hat H[\bs]$) but also the uncertainty on the underlying generative model ($\hat H[k]$). In loose words, $\hat H[\bs]+\hat H[k]$ reward algorithms that, while compressing the data in the under-sampling regime, translate as much of the $\hat H[\bs]$ nats as possible into relevant information.

\subsection{Statistical criticality of maximally informative samples}
\label{sec:stat-crit}

Taking $\hat H[k]$ as a quantifier of the amount of information that a sample $\hat \bs$ contains on its generative model, allows us to define {\em maximally informative samples} (MIS). These are samples that attain a maximal value of the relevance at a fixed resolution, i.e.
\begin{equation}
\label{MIS}
\hat \bs^*\in{\rm arg}\max_{\hat\bs:\hat H[\bs]=H_0}\hat H[k].
\end{equation}
Eq.~(\ref{MIS}) can be turned into the maximisation problem
\begin{equation}
\max_{\hat\bs,\mu}\left\{\hat H[k]+\mu\left(\hat H[\bs]-H_0\right)\right\}
\label{HkOptmu}
\end{equation}
where the Lagrange multiplier $\mu$ can be adjusted to achieve the desired level of resolution. For large $N$, MISs exhibit {\em statistical criticality}, in the sense that the number of states $\bs$ that are observed $k$ times in $\hat \bs$ follows a power law behaviour, \add{
\begin{equation}
\label{powlawmk}
m_k\sim k^{-\mu-1}
\end{equation}
where $\mu$ is the Lagrange multiplier in Eq.~(\ref{HkOptmu}). 
As shown in Ref.~\cite{MMR}, this is easily seen if $m_k$ is treated as a continuous variable. 
The maximisation of $\hat H[k]$ over the integers is a complex problem. Yet, it is important to remark that this solution would not be representative of samples that are obtained as random draws from an unknown probability $\bp(\bs)$. In order to account for stochastic sampling effects, Haimovici and Marsili~\cite{HM} analyse the case where the degeneracies $m_k$ are Poisson random variables, as in models of sampling processes~\cite{crane2016ubiquitous}. They provide an analytical estimate of the expected value $\E{\hat H[k]}$ over the distribution of $m_k$ and maximise it with respect to $\E{m_k}$. In this way, Ref.~\cite{HM} confirms Eq.~(\ref{powlawmk}). An example of the curve so obtained} is shown in Fig.~\ref{fig_iris} for $N=150$. The same plot also reports the value of $\E{\hat H[k]}$ for a random sample drawn from a uniform distribution $p(\bs)=1/|\mathcal{S}|$~\cite{statcrit}. 

Statistical criticality has attracted considerable attention, because of its ubiquitous occurrence in natural and artificial systems~\cite{newman2005power,sornette2006critical,clauset2009power,MoraBialek,munoz2018colloquium,roli2018dynamical,sharpee2019argument}. The quest for a general theory of statistical criticality has identified several mechanisms~\cite{newman2005power,sornette2006critical}, from the Yule-Simon sampling processes~\cite{simon1955class} and multiplicative processes~\cite{sornette1998multiplicative}, to Self-Organised Criticality~\cite{SOC}. The latter, in particular, is based on the observation that power law distributions in physics occur only at the critical point of systems undergoing a second order phase transition. This raised the question of why living systems such as cells, the brain or the immune system should be poised at criticality~\cite{MoraBialek}. While there is mounting evidence that criticality enhances information processing and computational performance~\cite{roli2018dynamical}, it is fair to say that the reason why statistical criticality is so ubiquitous has so far remained elusive~\cite{sorbaro2019statistical}. 

A general explanation of statistical criticality which encompasses both the frequency of words in language~\cite{Zipf} and the organisation of the immune system of zebrafish~\cite{Mora5405}, to name just two examples, cannot be rooted in any specific mechanism. The fact that statistical criticality emerges as a property of maximally informative samples provides an information theoretic rationale encompassing a wide variety of phenomena. In loose words, statistical criticality certifies that the variables $\bs$ used to describe the system are informative about the way the data has been generated. For example, the fact that the distribution of city sizes follows a power law, whereas the distribution by ZIP code does not, suggests that the city is a more informative variable than the ZIP code on \add{where} individuals decide to live (i.e. on the generative process). Likewise, the fact that the distribution of gene abundances in genomes or words in books exhibit statistical criticality~\cite{LEGO} is consistent with the fact that these are the relevant variables to understand how evolution works or how books are written. 

It is worth to remark that Zipf's law $m_k\sim k^{-2}$ corresponds to the point $\mu=1$ where $\hat H[\bs]+\hat H[k]$ is maximal. 
Indeed the exponent $\mu$ is related to the slope of the $\hat H[k]$ versus $\hat H[\bs]$ curve~\cite{MMR,statcrit}. Hence, the exponent $\mu$ encodes the tradeoff between resolution and relevance: a decrease of one \add{nat} in resolution affords an increase of $\mu$ \add{nats} in relevance. Hence Zipf's law emerges at the optimal tradeoff between compression and relevance, that separates the noisy regime ($\mu>1$) from the lossy compression regime ($\mu<1$). 

\subsubsection{A digression in Quantitative Linguistics}

The finding that samples that satisfy the optimisation principle (\ref{MIS}) exhibit statistical criticality was first derived in the information theoretic analysis of texts, by Balasubrahmanyan and Naranan~\cite{naranan1992information,balasubrahmanyan2002algorithmic}, who introduced $\hat H[k]$ in this context, under the name of {\em degenerate entropy}.  Leaving aside many details, for which we refer to~\cite{balasubrahmanyan2002algorithmic}, we note that the main gist of the arguments given in~\cite{naranan1992information} agrees with the main thesis which is presented in this review. In brief, the framework \add{of Refs.~\cite{naranan1992information,balasubrahmanyan2002algorithmic}} builds on the assumption that a language is efficient if shorter words are more frequent than longer ones. This principle, that Zipf termed the {\em principle of least effort}, was later formalised in coding theory~\cite{CoverThomas}, which states that the length of the codeword of a word $\bs$ that occurs $k_{\bs}$ times in a text of $N$ words, should be of $\ell_{\bs}=-\log_2\frac{k_{\bs}}{N}$ bits. If $m_k$ is the number of words that occur $k$ times in a text, then 
\[
\mathcal{W}=\prod_k\frac{(km_k)!}{(k!)^{m_k}}\sim e^{N\hat H[\bs|k]}
\]
is the number of texts that can be obtained from it by scrambling the worlds with the same word frequency (or codeword length) in all possible ways. A text which is randomly scrambled in this way will retain less and less of the meaning of the original text, the larger is $\mathcal{W}$. For example, random scrambling will have no effect on a text with $m_k\le 1$ for all $k$, because $\mathcal{W}=1$, but it will most likely produce a meaningless text if all words occur just once (i.e. $m_1=N$). The capacity to preserve meaning is a statistical feature of the language, that can be measured by $\log \mathcal{W}$. This leads to the definition of {\em 
Optimum Meaning Preserving Codes}~\cite{naranan1992information,balasubrahmanyan2002algorithmic}, as those that minimise\footnote{In fact, Naranan and Balasubrahmanyan~\cite{naranan1992information} define their information theoretic model for language as the solution of the maximisation of $\hat H[\bs]$ at fixed $\hat H[k]$. This is only apparently different from the definition given here, because it is equivalent to the maximisation of $\hat H[k]$ at fixed resolution $\hat H[\bs]$, which is Eq.~(\ref{MIS}). Eq.~(\ref{MIS}) in turn is equivalent to the minimisation of $\hat H[\bs|k]$ at fixed $\hat H[\bs]$.} $\hat H[\bs|k]\simeq \frac 1 N \log\mathcal{W}$ at fixed coding cost $\hat H[\bs]$. 

Although the general idea of the origin of Zipf's law is not new, our derivation gives a precise meaning to the exponent $\mu$ in terms of the tradeoff between resolution and relevance. Indeed the exponent $\mu$ is related to the trade-off between resolution and relevance, in the sense that it is related to the slope of the $\hat H[k]$ versus $\hat H[\bs]$ curve~\cite{MMR,statcrit}. 

\add{This interpretation of Zipf's law in terms of compression, leads to the prediction that}
a modern translation of an ancient text should have a lower $\mu$ than that of the earlier versions. This is because modern translations can be thought of as compressed version of the older ones. In the case of the Holy Bible, Ref.~\cite{mehri2017variation} reports a value $1/\mu=0.938$ and $1/\mu=0.969$ for Hebrew and Coptic, respectively, and larger values for Latin ($1.065$), German ($1.191$) and English ($1.258$), and Benz {\em et al.}~\cite{bentz2014zipf} report a value $1/\mu=1.03$ for the Old English Bible  which is significantly lower than that for the Modern English version ($1/\mu=1.22$). Cancho-i-Ferrer~\cite{i2005variation} report values of $\mu>1$ for fragmented discourse in schizophrenia, whereas texts in obsessive patients, very young children and military combat texts \add{exhibit statistical criticality} with exponents $\mu<1$. This is typical of repetitive and stereotyped patterns, also in agreement with the tradeoff between resolution and relevance discussed above.} 

\subsubsection{The tradeoff within a sample}

The relation $m_k\sim k^{-2}$ can also be interpreted in terms of the tradeoff between precision and noise, within the sample. As we discussed in the example of clustering (see Fig.~\ref{fig_iris}), the distinction between data points in different states $\bs$ is {\em a priori} arbitrary whereas the distinction between points in frequency is not, since it depends on the data. The relation $m_k\sim k^{-2}$ corresponds to an optimal allocation of discriminative power. To understand this, we first remind that $-\log(k/N)$, is the number of nats needed to identify a point $i$ in the sample given its value of $\bs^{(i)}$ (with $k_{\bs^{(i)}}=k$), whereas $\log(km_k)$ is the number of nats needed to identify a point $i$ based on the frequency of occurrence of the state $\bs^{(i)}$ it belongs to. Then the relation $m_k\sim k^{-2}$ implies that the number of nats needed to identify a point of the sample in terms of $\bs$ matches the number of nats needed to identify it in terms of its frequency $k$, up to a constant, across all frequencies, i.e.
\begin{equation}
\log (km_k)\simeq -\log(k/N)+c\,.
\end{equation}
In a MIS with $\mu>1$, the representation in terms of frequency is more noisy on the low frequency part of the spectrum, than on the high frequency part. This situation describes a noisy sample where poorly sampled points are resolved more efficiently in terms of the arbitrary labels $\bs^{(i)}$ than in terms of their frequency $k_{\bs^{(i)}}$. Conversely, for $\mu<1$ the variable $\bs$ \add{of} well sampled states carry relatively more information than $k$, with respect to poorly sampled states. 

\bigskip

Summarising, the observation of Zipf's law indicates that the variables that appear in the data are relevant, because they deliver an efficient allocation of the available information budget across sample points. The occurrence of Zipf's law in the frequencies of words~\cite{Zipf}, in the antibody repertoire of the immune system~\cite{Immune,Mora5405} and in firing patterns of the neurons of the retina~\cite{retina}, is \add{consistent with the view that these systems are efficient representations (of concepts, antigens and images, respectively).}
It is worth to remark that, these are three examples of efficient representations that are generic and featureless. Generic means that they process data drawn from a broad ensemble of inputs. Featureless means that the representation does not depend on specific features. In this sense, Zipf's law is a statistical signature of the most compressed, generic and featureless efficient representations. 

The fact that statistical criticality emerges in maximally informative samples does not explain why and how it occurs in a specific system. Statistical criticality merely serves as a certificate of significance of the data, that call for further analysis. Fig.~\ref{fig_iris} shows, for example, that the relevance allows us to distinguish interesting data from pure noise (corresponding to a random cluster structure). \add{Also} statistical criticality does not \add{require} that the system should necessarily be poised at a special (critical) point or at the ``edge of chaos''. In some system, this could be the outcome of the process searching the most relevant variables, by trial and error.

It is worth stressing that the notion of efficient representation that emerges from the maximisation of the relevance is independent of what the sample represents. 
This contrasts with other notions of efficient representations, such as those based on the information bottleneck method~\cite{IB}, which are defined with respect to a predefined input-output relation, e.g. between the past and the future of a process as in~\cite{EffRepBialek,Chalk186}. An absolute notion of relevance separates the two aspects of how information is efficiently encoded in a sample and of how the data should be interpreted, or decoded. This opens the way to unsupervised methods to identify relevant variables in high dimensional datasets. We believe that this is one of the most promising avenues of future research. An example will be discussed in Section~\ref{sec:neuro} in more detail.

\subsection{Maximal relevance and criticality in efficient coding and statistical learning}
\label{sec:maxrel-crit}

The above discussion implies that when samples are generated in order to be maximally informative, they should exhibit statistical criticality. 
This hypothesis is corroborated by the theory of optimal experimental design~\cite{atkinson2007optimum}. This prescribes that experiments should be designed in order to maximise the (determinant or the trace of the) Fisher information, in order to maximise the expected accuracy of parameter estimates. The Fisher information in parametric models is maximal close to critical points (see e.g. \cite{mastromatteo}). Hence samples generated from optimally designed experiments are expected to exhibit critical features. 

Natural and artificial learning systems offer a further test for the criticality hypothesis. The hypothesis that the brain operates in a critical state has been advanced by many authors (see \cite{beggs2008criticality} and references therein). 
The observation that neural networks have enhanced computational efficiency when they are tuned at the edge of chaos dates back at least 30 year~\cite{langton1990computation}. Tuning a network at a critical point is an accepted criterium for optimality in reservoir computing~\cite{bertschinger2004real,livi2017determination}. Sharpee~\cite{sharpee2019argument} has argued that statistical criticality in many sensory circuits and in the statistics of natural stimuli arises from an underlying hyperbolic geometry, that she argues is consistent with the principle of maximal relevance discussed here.

Rule {\em et al.}~\cite{hennig} and Song {\em et al.}~\cite{SMJ} have shown that the internal representation of well trained learning machines such as Restricted Boltzmann Machines (RBM) and Deep Belief Networks (DBN) exhibits statistical criticality. In particular, Song {\em et al.}~\cite{SMJ} have tested the theoretical insights based on the maximum relevance principle. They found that the relevance of different layers in DBNs approaches closely the maximum theoretical value. Furthermore, they confirmed that the frequency with which a state $\bs$ of the internal representation occurs obeys a power law distribution with an exponent $\mu$ that decreases with the depth of the layer. The layer which approaches most a Zipf's law behaviour ($\mu\simeq 1$) is the one with best generative performance. Shallower layers with $\mu>1$ generate noisy outputs, whereas deeper ones ($\mu<1$) generate stereotyped outputs that do not reproduce the statistics of the dataset used in training. This is reminiscent of the phenomenon of {\em mode collapse} observed in generative adversarial networks~\cite{goodfellow2014generative}, which refers to the situation where the learned model ``specialises'' to 
generate only a limited variety of the inputs with which it has been trained.

If  statistical criticality is the signature of maximally informative samples, we should find it in the theory of optimal codes, which deals precisely with compressing data generated by a source in an efficient way. Cubero {\em et al.}~\cite{RyanMDL} have addressed this issue within Minimum Description Length (MDL) theory~\cite{MDL}. In brief, MDL deals with the problem of optimally compressing samples generated as independent draws from a parametric distribution $f(\bs|\theta)$, with unknown parameters $\theta$. MDL finds that the optimal code for a sample $\hat \bs$ has a minimal length of $-\log \bar P(\hat\bs)$ nats, where
\begin{equation}
\label{NML}
\bar P(\hat\bs)=e^{-\mathcal{R}}\prod_{i=1}^Nf\left(\blue{{\bs}^{(i)}}|\hat\theta(\hat\bs)\right)\end{equation}
\add{is called the {\em normalised maximum likelihood}. In Eq.~(\ref{NML}), $\hat\theta(\hat\bs)$ is the maximum likelihood estimate of the parameters $\theta$ for the sample $\hat\bs$ and 
\begin{equation}
\mathcal{R}=\log\sum_{\hat \bs}
\prod_{i=1}^Nf\left(\blue{{\bs}^{(i)}}|\hat\theta(\hat\bs)\right)
\end{equation}
is the stochastic complexity of model $f$}. Cubero {\em et al.}~\cite{RyanMDL} show that typical samples generated from $\bar P(\hat \bs)$, for different models, feature values of the relevance that are close to the maximal attainable one, at the corresponding level of resolution. Correspondingly, the frequency distributions exhibit statistical criticality. \add{Xie and Marsili~\cite{xie2021random} reach a similar conclusions concerning the origin of statistical criticality of the internal representation of well trained learning machines observed in Refs.~\cite{SMJ,hennig}. }

\subsubsection{Statistical criticality and the mode collapse phase transition} 

Does statistical criticality emerges because the system is poised at a critical point? And if so, what is the order parameter and what is the conjugate variable associated with it that needs to be fine tuned? What is the symmetry that is broken across the transition?

Cubero {\em et al.}~\cite{RyanMDL} show that, studying the large deviation of MDL codes it is possible to answer these questions in a precise manner. They consider the large deviation of the coding cost $\hat H[\bs]$ (i.e. the resolution) on the ensemble of samples defined by $\bar P(\hat\bs)$. This entails studying the \blue{tilted} distribution~\cite{CoverThomas}
\begin{equation}
\label{ldt_samp}
P_\beta(\hat \bs)=\frac{1}{Z_\beta} \bar P(\hat \bs)e^{\beta N\hat H[\bs]}
\end{equation}
which \blue{permits exploring the properties of the} atypical samples $\hat \bs$ that exhibit anomalously large or small values of $\hat H[\bs]$, with respect to the typical value $\E{\hat H[\bs]}$, \blue{where the expectation $\E{\cdots}$ is over the distribution $\bar P(\hat\bs)$ of samples}. For $\beta>0$ ($\beta<0$) the distribution $P_\beta$ reproduces large deviations with coding cost higher (lower) than the typical value. For $\beta>0$ the distribution $P_\beta$ has support on all samples. For $\beta<0$ the distribution instead  concentrates on samples with identical outcomes $\bs^{(1)}=\bs^{(2)}=\ldots =\bs^{(N)}$, an instance of the mode collapse phenomenon discussed above. The transition to the "mode collapse" phase is sharp and it occurs at the critical point $\beta_c=0$ that coincides exactly with MDL codes $\bar P$. The symmetry that is broken, therefore, is the permutation symmetry among sample outcomes $\bs$: in the ``disordered'' phase for $\beta<\beta_c$ all sample outcomes $\bs$ occur in the sample, whereas for $\beta>\beta_c$ one enters the ``ordered'' asymmetric phase where one particular state $\bs$ occurs disproportionally often in the sample. 

This analysis identifies both the order parameter -- the resolution $\hat H[\bs]$ -- and the associated critical parameter $\beta_c=0$ that defines the phase transition. In hindsight, the fact that the MDL code $\bar P$ is critical is not surprising, given that $\bar P$ is related to the optimal compression of samples generated from $f(\bs|\theta)$, with unknown parameters $\theta$. Criticality arises as a consequence of the fact that samples generated from $\bar P$ are {\em incompressible}. 

\begin{figure}[ht]
\centering
\includegraphics[width=\textwidth,angle=0]{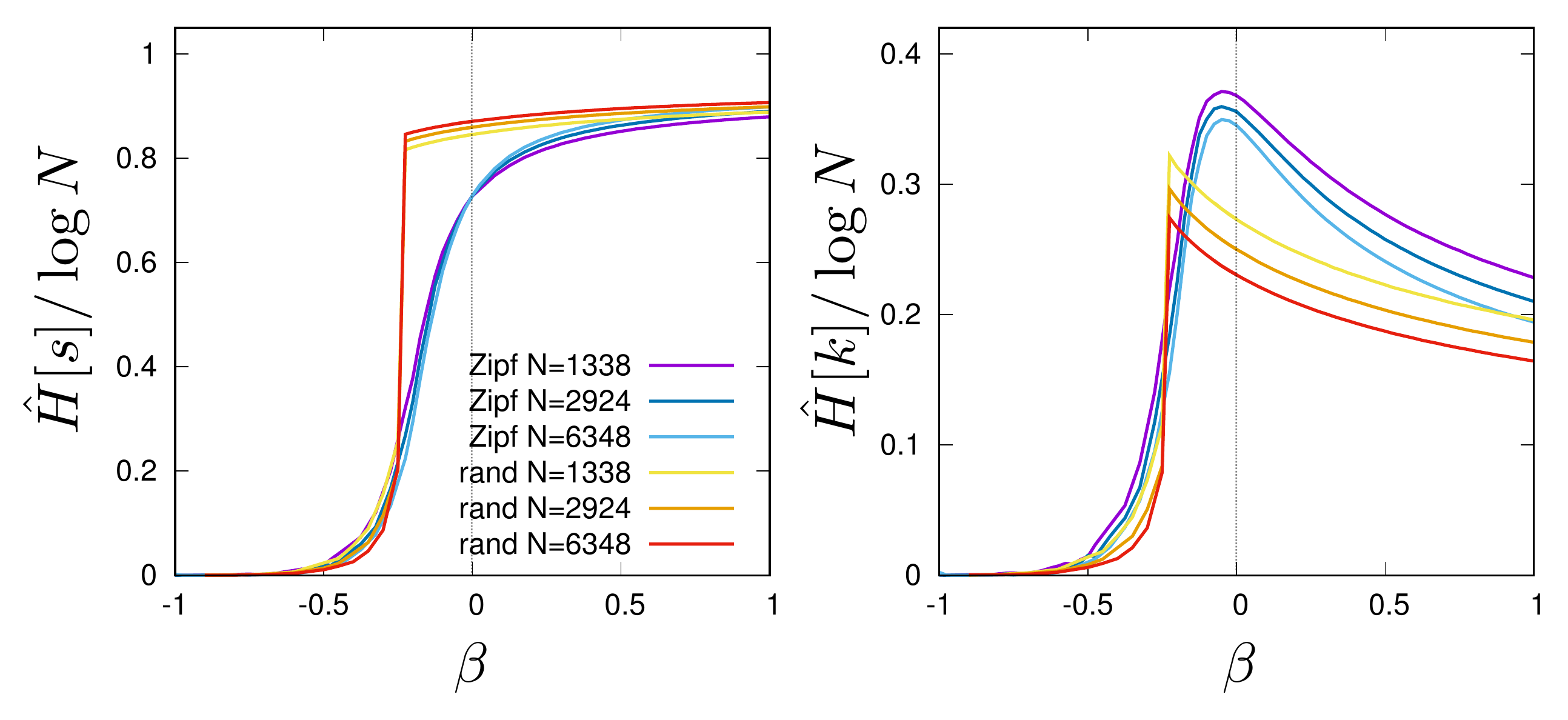}
\caption{\label{fig_LDT} Resolution (left) and relevance (right) of large deviations samples drawn from Eq.~(\ref{ldt_samp}), for a typical sample obeying Zipf's law or for a random sample. In both cases, the set of states has $|\mathcal{S}|=2 N$, where $N=1338,2924$ and $6348$ are the number of samples. The values of $\hat H_q[\bs]$ and $\hat H_q[k]$ are averaged over $10^4$ independent samples obtained by \blue{Markov Chain Monte Carlo}.}
\end{figure}

Does the relation between between criticality and incompressibility extend further than MDL? 
This hypothesis can be tested in the following way. 

\blue{Assume that, the probabilities $\bp(\bs)$, are a priori Dirichelet distributed, namely
\begin{equation}
P_0(\bp)=\Gamma \left (\sum_{\bs} a_{\bs} \right) \prod_{\bs} \frac{\bp(\bs)^{a_{\bs}-1}}{\Gamma(a_{\bs})} \delta \left ( \sum_{\bs} \bp(\bs) -1 \right)
\end{equation}
prior with parameters $a_1,a_2,\cdots$ are the parameters of the Dirichelet prior and with $\Gamma$ is the gamma function. After observing the frequencies $\hat k=\{k_{\bs}\}$ of the observed states in the sample $\hat \bs$, and taking $a_1,a_2,\cdots=a$ to reflect the a priori symmetry of the states, we can, as in Refs.~\cite{HM,NemenmanCoinc}, use the Bayes law to write the posterior over the  the probabilities $\bp(\bs)$ as
\begin{equation}
P(\bp|\hat k)=\frac{\Gamma(N+aM)}{\prod_{\bs}\Gamma(k_{\bs}+a)}\prod_{\bs}\bp(\bs)^{k_{\bs}+a-1}\delta\left(\sum_{\bs}\bp(\bs)-1\right)
\end{equation}
where $M$ is the number of states.}


\add{In order to explore large deviations of $\hat H[\bs]$,
we should consider\blue{, as was done for the MDL codes in Eq.\ \eqref{ldt_samp}, samples generated from the tilted} distribution
\[
\bp_\beta(\hat \bs')=A\prod_{i=1}^N\bp(\bs^{(i)})e^{\beta N\hat H_q[\bs]},\qquad
\hat H_q[\bs]=-\sum_{\bs'}\frac{q_{\bs'}}{N}\log \frac{q_{\bs'}}{N}
\]
where $q_{\bs'}$ is the number of times the state $\bs'$ occurs in the sample $\hat \bs'$ and $A$ is a normalising constant. Marginalising this distribution over the posterior $P(\bp|\hat k)$, yields the distribution of samples $\hat \bs'$ that realise large deviations of the coding cost
\begin{equation}
P_\beta(\hat \bs'|\hat k)=\frac{1}{Z_\beta}\prod_{\bs'}\frac{\Gamma(k_{\bs'}+q_{\bs'}+a)}{\Gamma(k_{\bs'}+a)}e^{\beta N\hat H_q[\bs']},
\end{equation}
where $Z_\beta$ is a normalising constant. Fig.~\ref{fig_LDT} shows the properties of samples drawn from this distribution by \blue{Markov Chain Monte Carlo}, as a function of $\beta$. It compares the results obtained when the initial sample $\hat \bs$ is chosen to obey Zipf's law (violet and blue lines) and for samples $\hat \bs$ drawn from a uniform distribution $\bp(\bs)=1/M$ (yellow and red lines).} Large deviations of samples that obey statistical criticality exhibit the same mode collapse phase transition discussed above, at $\beta_c=0$. By contrast the large deviations of $\hat H_q[\bs]$ for a random sample do not exhibit any singularity at $\beta=0$. Rather they feature a discontinuous phase transition for some $\beta_c<0$. The nature of the phase transition for a Zipf distributed sample is markedly different. The relevance shows a maximum for $\beta\approx 0$, signalling that the frequency distribution is maximally broad at $\beta=0$. This behaviour is \add{reminiscent of} the maximum in the ``specific heat'' discussed in Ref.~\cite{MoraBialek,retina}.
This large deviation analysis corroborates further the hypothesis that the phase transition associated to generic statistical criticality may have a phenomenology similar to mode collapse, and that the resolution $\hat H[\bs]$ plays the role of an order parameter.

\subsection{{An application to neuroscience}}
\label{sec:neuro}

As mentioned in the Introduction, the notion of relevance that we defined in Eq.\ \eqref{eq:relevance_sample} as the entropy of the observed frequencies, can be useful for ranking how useful datasets are in real life. In this section\add{, following Ref.~\cite{RyanMSR},} we briefly discuss one such application in the case of neural coding which was mentioned in section \ref{examples}.

Research in neural coding aims at understanding how neurons in a neural circuit represent information, how this information can be read off by researchers through recording neural activity, or can be transmitted and used by other parts of the brain. It also aims at using answers to these question for understanding the properties of neural circuits, e.g. the distribution of outputs and inputs, or the type of neurons representing certain information in a brain region. All this is done through analysing experimental data as well as theoretical models of information representation\add{s}. 

\begin{figure}[ht]
\centering
\includegraphics[width=0.8\textwidth,angle=0]{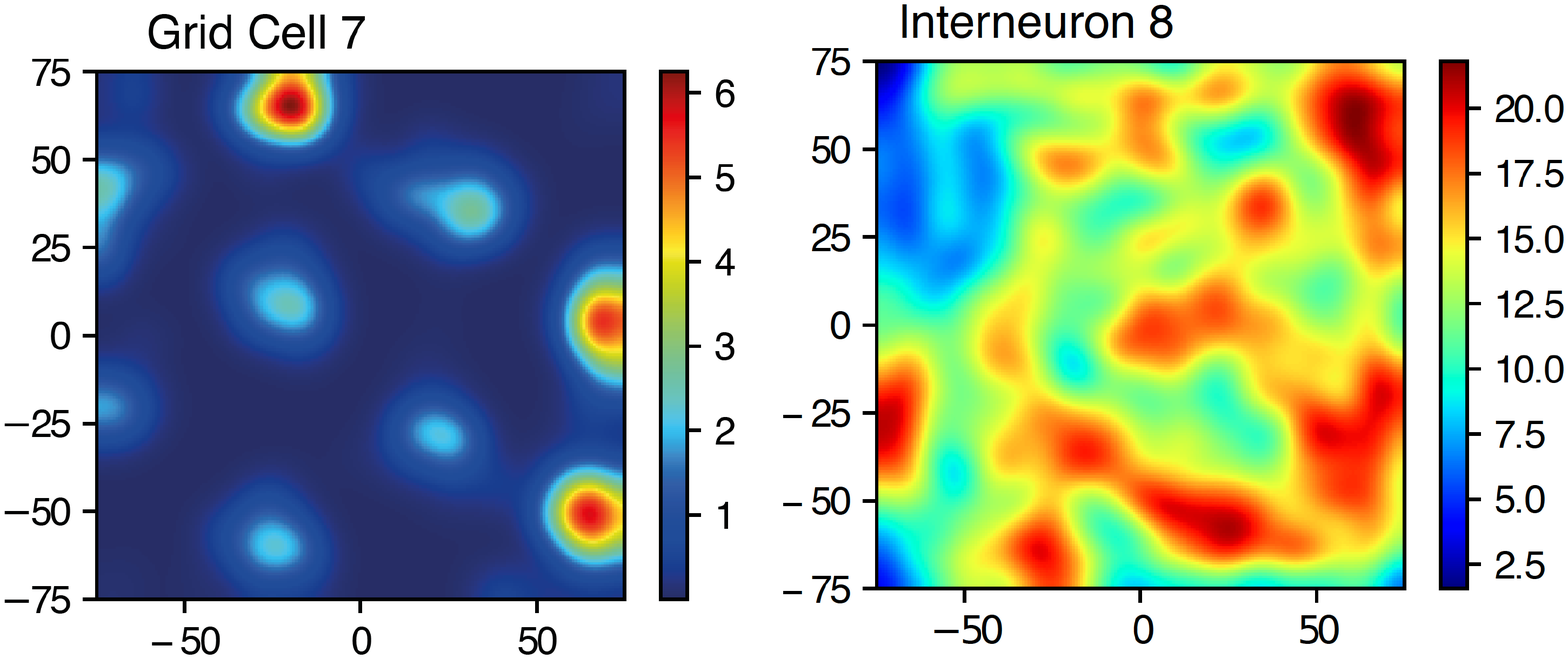}
\caption{\label{Gridcells} Rate maps of two neurons showing their firing rates as a function of a the spatial position of a rat freely moving in a 1.5m x1.5 m sqaure box. The cell on the left panel is a grid cell with firing fields clearly organised as a hexagonal lattice, while the one on the right is an interneuron; figure adapted from~\cite{RyanMSR}}
\end{figure}

\blue{This approach is clearly exemplified in research on understanding neural coding of spatial navigation which has seen a huge advancement in the past two decades.} In this line of research, the hippocampus, an area in the mammalian temporal lobe, and the nearby area of the Medial Entorhinal Cortex (MEC) play a signifiant role; see \cite{rowland2016ten,moser2014grid}. The hippocampus is involved in a variety of tasks, ranging from episodic memory (memory of specific events happened in a specific time and specific place, e.g. a particular wedding party) and regulating emotions, to spatial navigation; 
\blue{see e.g. \cite{eichenbaum2004hippocampus,eichenbaum2018spatial} for review}\add{. The hippocampus} receives its major input from MEC. A rather interesting feature of MEC is the presence of the so called grid cells~\cite{Mosers}. As a rat freely moves in a familiar box, a grid cell fires at specific positions, called grid fields, and these fields together form a hexagonal pattern; see Fig.\ \ref{Gridcells} (left). Different grid cells have different grid firing patterns, differing from each other in grid spacing, orientation and spatial phase. The firing of a grid cell, in addition to its spatial selectivity, might be modulated by other factors, or so called {\em co-variates}: head direction, reward, speed and perhaps others. The degree of this modulation highly depends on where exactly the grid cell is recorded from. In addition to grid cells, MEC includes other cells, e.g. the neurons in Fig.\ \ref{Gridcells} (right), some selective for spatial covariates, and some not, at least not obviously, or in fact to any measured covariates. All in all, the diversity of neural responses, the rich structure of the firing of neurons in the MEC and its proximity to the hippocampus has lead to the conclusion that it is a particularly important region for spatial navigation and cognition. 

What neurons in the MEC code for was discovered essentially by a process that involved a large degree of trial and error. Anatomical arguments and previous research had led to the suspicion that this area is likely to include neurons that represent spatial information. This suspicion led to efforts for recording from neurons in this area while a rat is foraging in a box. What is sometimes forgotten in the history of the discovery of grid cells is that initially the grid structure was not visible as the recording was done in small boxes where the repetitive patterns of grid firing could not be observed \cite{fyhn2004spatial}, and it was only later that the repetitive pattern was observed. 

Although the remarkable spatial selectivity of grid cells or other cells in MEC are striking, one should note that neurons that receive this information do not have any information about the spatial covariates of the animal to start with. So they cannot correlate the neural activity they receive with such covariates and detect spatially informative cells like grid cells, or cells that carry information about other covariates: they are like an experimentalist that has the data on the spikes of the neuron, but no other covariates. And they need to decide which neurons in MEC they should listen to by only seeing the spike train of that neuron. But is there a way to decide this? The results in \cite{RyanMSR} and summarised here show that the answer is yes and that indeed one can use the notion of relevance defined in this section to define a measure to rank the activity of these neurons for this purpose.


\blue{Let us consider the spike train recorded from a neuron emitting $N$ spikes in a period of  duration $T$ as in Fig. \ref{MSRdef} A. This recording period can be devided into bins of size $\Delta t$. Assuming that $k_{\bs}$ spikes are emitted with the $\bs^{\rm th}$ time bin, we can define the resolution in this case as the entropy $\hat H[\bs]=-\sum_{\bs} k_{\bs}/N \log_N k_{\bs}/N$. With this definition, $\hat H[\bs]$ corresponds directly to the temporal resolution at which the neural activity is probed. As opposed to $\Delta t$, $\hat H[\bs]$ provides an intrinsic \add{and adimensional} measure \add{of time} resolution: for any recordings length $T$, $\hat H[s]=0$ for $\Delta t \geq T$ and $\hat H[\bs]=1$ for all values of $\Delta t$ small enough that the bins include at most one spike.}

\blue{Denoting by $m_k$ the number of times bins in which $k$ spikes were emitted, $k m_k/N$ is the fraction of spikes that fall in bins with $k$ spikes. We thus define relevance \add{in the usual way,} as $\hat H[k]=-\sum_k km_k/N \log_N k m_k/N.$ As can be seen in Fig. \ref{MSRdef} B, for small time bins, where each bin only includes a maximum of one spike, resolution $\hat H[\bs]$ is maximal, while $\hat H[k]=0$. Increasing the bin size, $\hat H[\bs]$ decreases monotonically until reaching zero for $\Delta t= T$, while, $\hat H[k]$, reaches a maximum at some values of $\Delta t$, before that, too, drops to zero.}

\begin{figure}[ht]
\centering
\includegraphics[width=0.8\textwidth,angle=0]{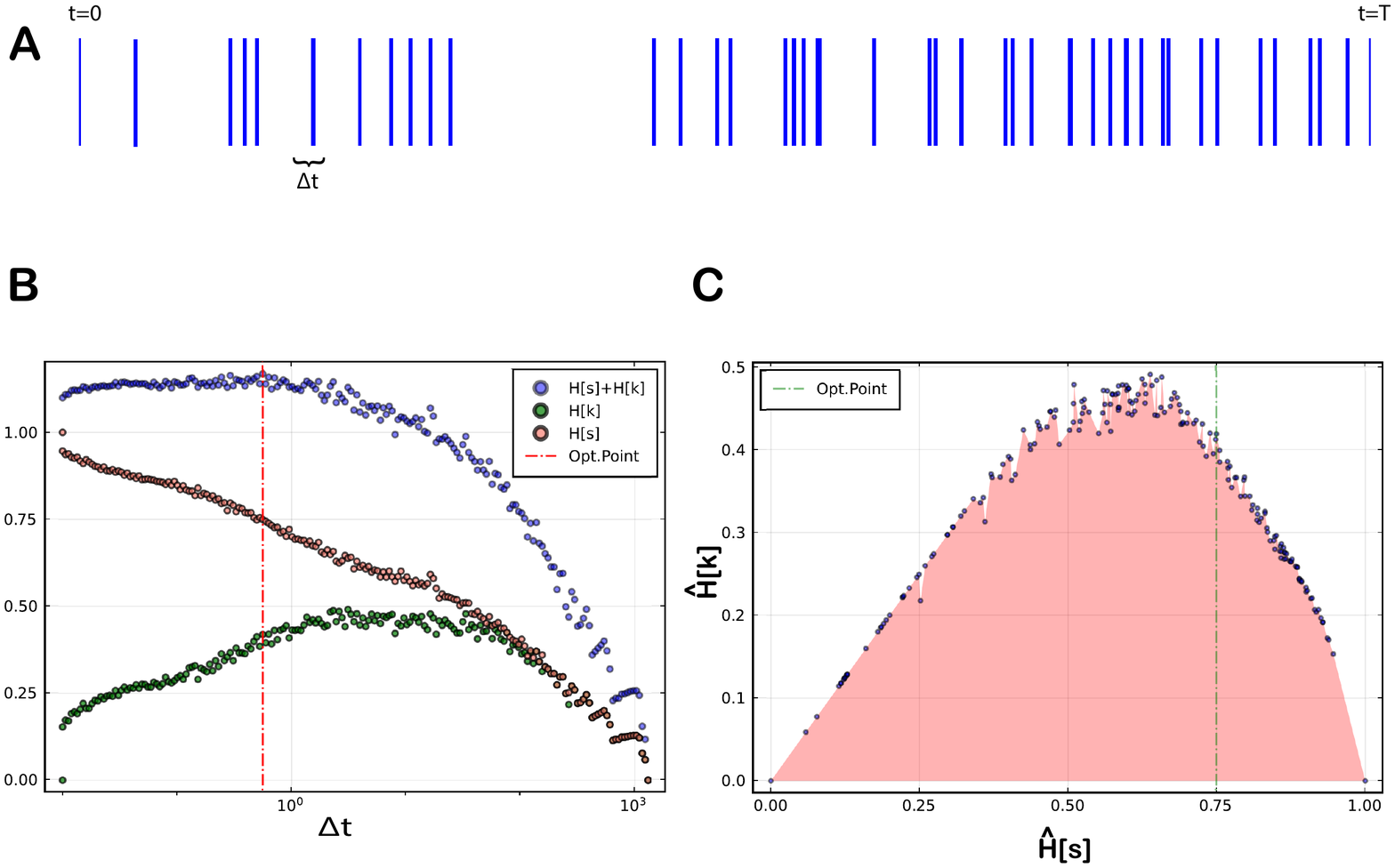}
\caption{\label{MSRdef}  \blue{(A) Spikes occurring at times shown by vertical lines in a recording period $T$. The recording period can be divided into time bins of size $\Delta t$ and the number of spikes falling in each time bin counted. (B) By varying $\Delta t$, resolution and relevance as defined in the text and the sum of the two vary, leading to resolution-relevance curve in (C). Data used in this figure are from a neuron recorded in the CA1 region of the hippocampus \cite{ledergerber2021task}}}
\end{figure}

\blue{A plot of $\hat H[k]$ versus $\hat H[\bs]$ is shown in Fig. \ref{MSRdef} C.The relevant time-scale of a neuron is typically unknown {\em a priori} and the same neuron can represent information across a range of time-scales, influenced by its physiological properties, the properties of the network of neurons it belongs to, and the behaviour of the animal. In order to take this into account, we thus consider the relevance at different (temporal) resolutions and define the Multi Scale-Relevance (MSR) of a neuron as the area under the $\hat H[k]\ va \ \hat H[\bs]$ curve \cite{RyanMSR}. As we discussed in Section \ref{sec:dataclustering}, an as indicated in Figs. \ref{MSRdef}B and C, an optimal point (time bin size $\Delta t$), in the sense of the tradeoff between resolution and relevance, can be defined as where the value of $\hat H[k]+\hat H[\bs]$ reaches its maximum, or equivalently where the slope of the $\hat H[k]\ vs\ \hat H[\bs]$, is equal to $-1$.
} 

How can this approach be useful in understanding neural coding? Fig.\ \ref{MSR-Info}A shows how \add{the MSR, so defined,} correlates with \blue{spatial information, calculated as the amount of information that the neuron carries, per spikes, a standard measure for characterising the spatial selectivity of neurons \cite{skaggs1993information}.} As can be seen in this figure, neurons that have a low value of the MSR do not carry any spatial information. On the other hand, all neurons that carry information on spatial covariates have a relatively high MSR value. High MSR neurons can have high or low mutual information. Fig.\ \ref{MSR-Info}B shows that using the 20 most spatially informative neurons and the 20 neurons with the highest MSR for decoding position leads to the same level of error.  

\begin{figure}[ht]
\centering
\includegraphics[width=0.65\textwidth,angle=0]{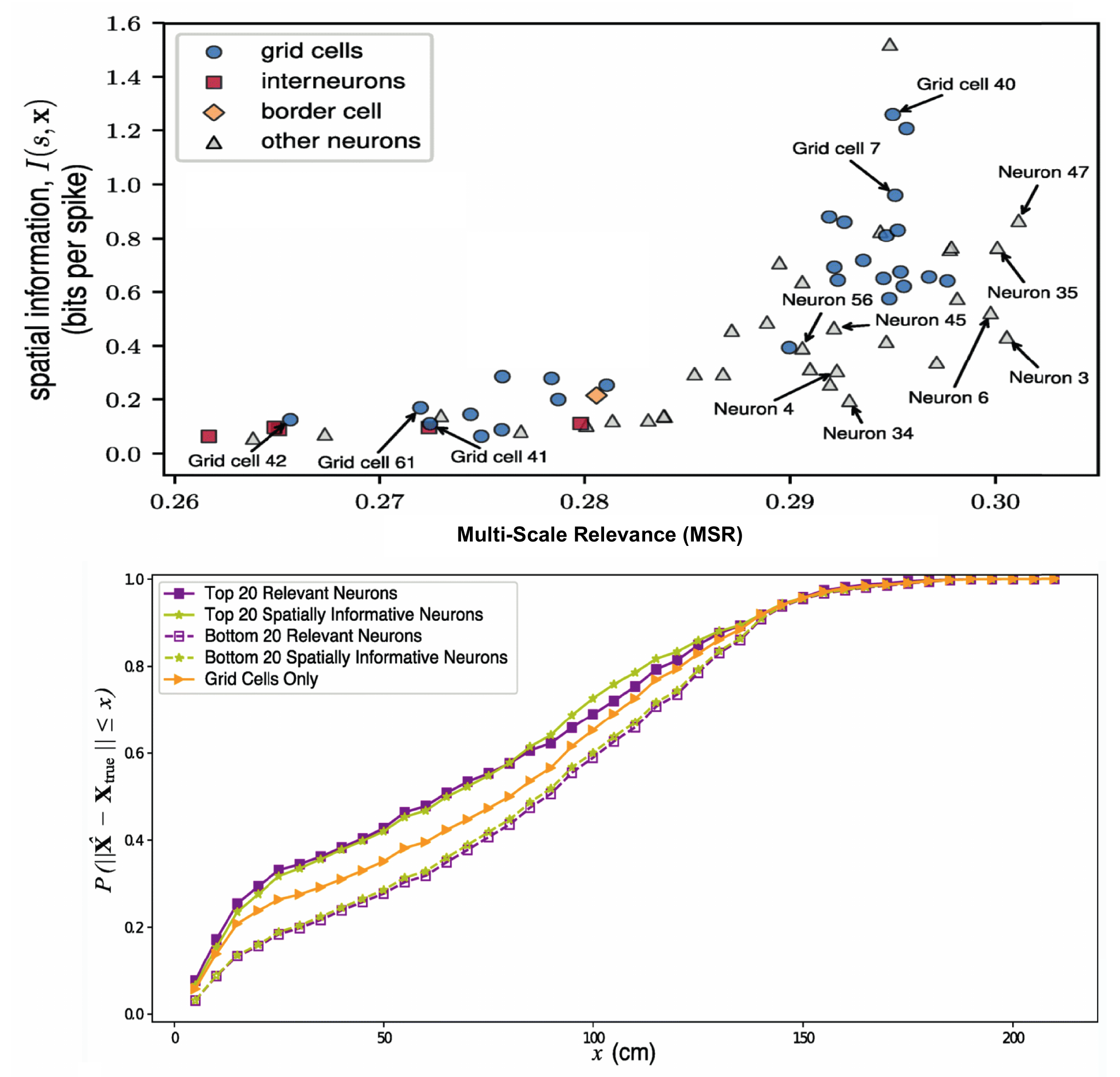}
\caption{\label{MSR-Info} (A) Spatial information (i.e. mutual information between the firing rate of a neuron and the position of the animal) versus Multi-Scale Relevance as defined in \cite{RyanMSR} for a group of 65 neurons recorded from the MEC of a rat freely moving in a 1.5m x 1.5 m box. (B) The cumulative distribution of the decoding error, defined as the distance between the actual position of the rat $X_{\rm true}$ and the decoded position $\hat X$. The decoding of the position was done using the activity of different populations of neurons as mentioned in the figure; see \cite{RyanMSR} for details where the figure us adapted from.}
\end{figure}

What does all these mean? Identifying informative neurons about space using measures such as spatial information \add{requires} to first know that these neurons are likely to be coding \add{for} space; this step is usually guided by knowledge of the function of nearby areas, anatomical considerations, lesion studies etc. One needs to measure the position of the animal and calculate an information measure between spatial position and the activity of neurons. MSR on the other hand, ranks neurons by just observing their spike trains. For an experimentalist analysing this data, the relationship shown in Fig.~\ref{MSR-Info} can thus be used to prune or discard neurons that are unlikely to be interesting. But more importantly, these results shows that, at least in principle, an upstream neuron which knows nothing about the content (e.g. spatial position, head direction etc) or form (e.g. over long time scale, short time scale) of information that these MEC neurons carry, can select which one to listen to and which one not to listen. 

\section{Statistical learning without data}
\label{sec:learning}

Let us now turn to the discussion of statistical models that are trained to represent complex datasets. We briefly remind the general setting and the main questions. Consider an unknown generating process $\bp(\vec x)$ of high dimensional data $\vec x$ such as digital pictures or gene expression profiles. We focus on the unsupervised learning task of approximating $\bp(\vec x)$ on the basis of a sample $\hat x=(\vec x^{(1)},\ldots,\vec x^{(N)})$ of $N$ observations, which are assumed to be independent draws from the unknown $\bp$. This is implemented by a generic learning machine 
\begin{equation}
p(\bs,\vec x)=p(\vec x|\bs)p(\bs)
\end{equation}
that projects the data $\vec x$ onto the internal states $\bs$ of the machine. Both $p(\vec x|\bs)$ and $p(\bs)$ depend on parameter that are adjusted in the training process, in order for the marginal distribution $p(\vec x)$ to reproduce as accurately as possible the sample $\hat x$, and hence the unknown generative process $\bp(\vec x)$. We shall abstract from details on the training process and focus exclusively on the properties of the learned representation $p(\bs)$. The way the internal representation is projected onto the data, which is encoded in $p(\vec x|\bs)$, clearly depends on the data. The main thesis of this section is that \add{the internal representation $p(\bs)$ satisfies universal statistical properties}, at least in an ideal limit. Hence they can be discussed without explicit reference to any dataset. More precisely, these properties originate from the postulate that when learning machines are trained on data with a rich structure, they approach the ideal limit of {\em optimal learning machines}\add{, i.e. of statistical models $p(\bs)$ whose relevance is maximal, for a given level of resolution, as} defined in Section~\ref{sec:general}. This thesis has been supported by different arguments by Cubero {\em et al.}~\cite{statcrit} and by Duranthon {\em et al.}~\cite{Odilon}, which we review below. 

The main prediction of the maximum relevance hypothesis is that learning machines should exhibit critical features, reminiscent of those that characterise critical phenomena in second order phase transitions. This prediction agrees with the observation that tuning machines to a critical point (or to the ``edge of chaos'') improves computational performance in generic learning tasks (see \cite{roli2018dynamical} for a review). This observation dates back  at least three decades~\cite{langton1990computation} and is so widely accepted that is often assumed as a design principle (see e.g.~\cite{livi2017determination}). 

We remark once again that our approach differs from attempts to understand learning in high-dimensions in the statistical mechanics literature (see e.g.~\cite{zdeborova2016statistical,tubiana2017emergence,zecchina,nguyen2017inverse}). These need to assume a models of the data at the outset \add{which}, apart from few examples~(see e.g.~\cite{mezard,goldt2019modelling,gherardi2020}), are structureless. Indeed learning machines differ substantially from systems described by statistical mechanics, such as inanimate matter, as we're going to see next.

\subsection{Statistical learning as inverse statistical mechanics}
\label{sec:statmech}
Statistical learning deals with the inverse problem of statistical mechanics. The latter aims at deriving the behaviour of a model that is specified by an Hamiltonian $\mathcal{H}[\bs]$. For a system in thermal equilibrium at inverse temperature $\beta$, all observables can be computed \add{as averages} on the Gibbs-Boltzmann distribution
\begin{equation}
\label{Gibbs}
p_{\rm eq}(\bs)=\frac{1}{Z(\beta)}e^{-\beta \mathcal{H}[\bs]}={\rm arg}\max_{p(\bs):~\E{\mathcal{H}}=\bar{\mathcal{H}}} H[\bs],
\end{equation}
where $Z(\beta)$ is the normalisation constant (the partition function). Eq.~(\ref{Gibbs}) can be derived from a principle of maximum entropy: 
the only information that is needed to compute $p_{\rm eq}(\bs)$ is the expected value~\add{$\bar{\mathcal{H}}$} of the energy, which determines $\beta$. Eq.~(\ref{Gibbs}) describes a system in thermal equilibrium with an environment (the heat bath) whose state $\vec x$ is largely independent of the state of the system\footnote{This is because the interaction term $\mathcal{V}[\bs,\vec x]$ in the Hamiltonian of the combined system $\mathcal{H}[\bs,\vec x]=\mathcal{H}[\bs]+\mathcal{H}[\vec x]+\mathcal{V}[\bs,\vec x]$ is small compared to $\mathcal{H}[\bs]$. Usually $\mathcal{V}$ is proportional to the surface whereas $\mathcal{H}[\bs]$ is proportional to the volume, hence the interaction is negligible in the thermodynamic limit.} $\bs$, i.e. $p(\bs|\vec x)\approx p_{\rm eq}(\bs)$. A system in equilibrium does not carry any information on the state of the environment it is in contact with, besides the value of the temperature which determines the average energy. Indeed, by the equivalence of the canonical and micro-canonical ensembles, all the statistical properties of a system in contact with its environment are asymptotically (for large systems) identical to those of an isolated system at the same (average) energy. As a result, the distribution of energy levels is sharply peaked around the mean value. States of matter where the energy features anomalously large fluctuations are atypical, and they require fine tuning of some parameters \add{to a critical point marking a continuous phase transition}.

Summarising, the distribution $p_{\rm eq}(\bs)$ applies to systems about which we have very rich prior information -- the Hamiltonian $\mathcal{H}[\bs]$ -- and that retains as little information as possible on its environment. \add{Critical behaviour is atypical.}

In the case of a learning system the situation is the opposite: The objective of a machine that learns is that of making its internal state $\bs$ as dependant as possible on the data $\vec x$ it interacts with, making as few prior assumptions as possible. The Hamiltonian of the learning machine is not given {\em a priori}. Rather\add{, during training,} it can \add{wander into} a large parametric space in order to adjust its energy levels in such a way as to reproduce the structure of the data. The Hamiltonian is itself the variable over which optimisation is performed. The objective function varies depending on the learning task, yet in all cases where the structure of the data is non-trivial, it needs to generate a statistical model $p(\bs)$ which differs markedly from the Gibbs-Boltzmann distributions studied in statistical mechanics. 

\add{We shall argue, following~\cite{statcrit,Odilon}, that} the relevance \add{provides} a measure of information capacity. This allows us to define an ideal limit of {\em optimal learning machines} whose properties can be studied without specific reference to a particular objective function, learning task or dataset. In this ideal limit, the internal representations $p(\bs)$ of the learning machines should satisfy the maximum relevance principle
\begin{equation}
\label{maxrel}
\max_{\{E_{\bs}\}:~\E{E_{\bs}}=H[\bs]} H[E],
\end{equation}
where $E_{\bs}=-\log p(\bs)$ is the coding cost of state $\bs$. The maximisation in Eq.~(\ref{maxrel}) is constrained to representations $p(\bs)$ with average coding cost given by the resolution $H[\bs]$ and that satisfy the normalisation constraint $\sum_{\bs}p(\bs)=1$. In words, an optimal learning machine is a statistical mechanics model with an energy spectrum which is as broad as possible, consistently with the resolution constraint.

In a real learning task, the optimisation~(\ref{maxrel}) is also constrained by the architecture of the learning machine, e.g. the form of $p(\vec x,\bs)$ as a function of the parameters (see e.g. Eq. \ref{eq:RBM} for RBMs), and by the available data~\cite{Odilon}. In this respect, over-parametrisation is clearly a desirable feature for general purpose learning machines, in order to make the search in the space of Hamiltonians as unconstrained as possible. In this regime, we conjecture that the behaviour of well trained learning machines depends weakly on the constraints imposed in a specific learning task and approaches the one described by the ideal limit of Eq.~(\ref{maxrel}). 

We shall first discuss the general properties of optimal learning machines and then relate their properties to those of maximally informative samples, that we discussed in the previous section. Finally, we will review the evidence that supports the hypothesis that this ideal limit is representative of the internal representations of real learning machines trained on data with a rich structure.

\subsection{The properties of Optimal Learning Machines}
\label{sec:OLMprop}

The principle of maximal relevance Eq.~(\ref{maxrel}) allows us to discuss the properties of learning machines independently of the data that they are supposed to learn, provided it has a non-trivial structure. In analogy with statistical physics, in this Section we refer to $E_{\bs}$ as the energy of state $\bs$, \add{setting the inverse temperature $\beta=1$.}
The relevant variable in the maximisation problem (\ref{maxrel}) is the degeneracy of energy levels $W(E)$, which is the number of internal states $\bs$ with energy $E_{\bs}=E$\add{, that we shall call the density of states}. Notice that, in statistical mechanics, the density of states $W(E)$ is determined {\em a priori} by the knowledge of the Hamiltonian. In this respect, statistical mechanics (Eq.~\ref{Gibbs}) and statistical learning (Eq.~\ref{maxrel}) can be seen as dual problems. 

As shown in Ref.~\cite{statcrit}, the solutions of Eq.~(\ref{maxrel}) feature an exponential density of states\add{\footnote{\add{As we're going to see in Section~\ref{sec:correspondance}, the exponential behaviour in Eq.~(\ref{WEexp}) is consistent with the power law behaviour of the frequency distribution introduced in the previous Section, 
and the parameter $\mu$ coincides with the exponent $\mu$ of the frequency distribution for a dataset $\hat\bs$ of states sampled from the distribution $p(\bs)$.}}} 
\begin{equation}
\label{WEexp}
W(E)=\sum_{\bs}\delta\left(E_{\bs}-E\right)=W_0 e^{\mu E}.
\end{equation}
We shall derive this result within a simple model below. Before doing that, let us mention that Eq.~(\ref{WEexp}) 
implies that the entropy $S(E)=\log W(E)$ of states at energy $E$ varies linearly with $E$. The slope of this relation equals
\begin{equation}
\label{internalmu}
\mu=\frac{dS}{dE}=\frac{d H[\bs|E]}{dH[\bs]}
\end{equation}
where $H[\bs|E]=H[\bs]-H[E]=\E{S(E)}$ measures the residual amount of \add{uninformative nats of the resolution $H[\bs]=\E{E_{\bs}}$. Indeed $H[\bs|E]=\E{\log W(E)}$ arises from  the degeneracy of states $\bs$ with the same value of $E$, and hence with the same probability.} \add{Hence, we shall take $H[\bs|E]$} as a measure of irrelevant information, or noise. As in the case of a maximally informative sample (Eq. \ref{MIS}), $\mu$ describes the tradeoff between resolution and relevance: Further compression (i.e. decrease in $H[\bs]$) removes $\mu$ bits of noise, and hence \add{lead to} an increase of $\mu-1$ bits in relevance $H[E]$. Notice that $H[s|E]=\E{\log W(E)}$ has the flavour of a Boltzmann entropy, whereas $H[\bs]$ is akin to the average energy. However, the relation between entropy and energy in optimal learning machines is convex (see Fig.~\ref{fig_OLM}), whereas in statistical mechanics it is concave. This is natural in learning machines, because the largest rate $\mu$ of conversion of noise into relevant information attains at high resolution, and $\mu$ is expected to be an increasing function of $H[\bs]$. This contrasts with the fact that the average energy is expected to decrease with $\beta$ (i.e. to increase with temperature) in statistical mechanics. Hence it is misleading to interpret $\mu$ as an inverse temperature. 

\begin{figure}[ht]
\centering
\includegraphics[width=0.8\textwidth,angle=0]{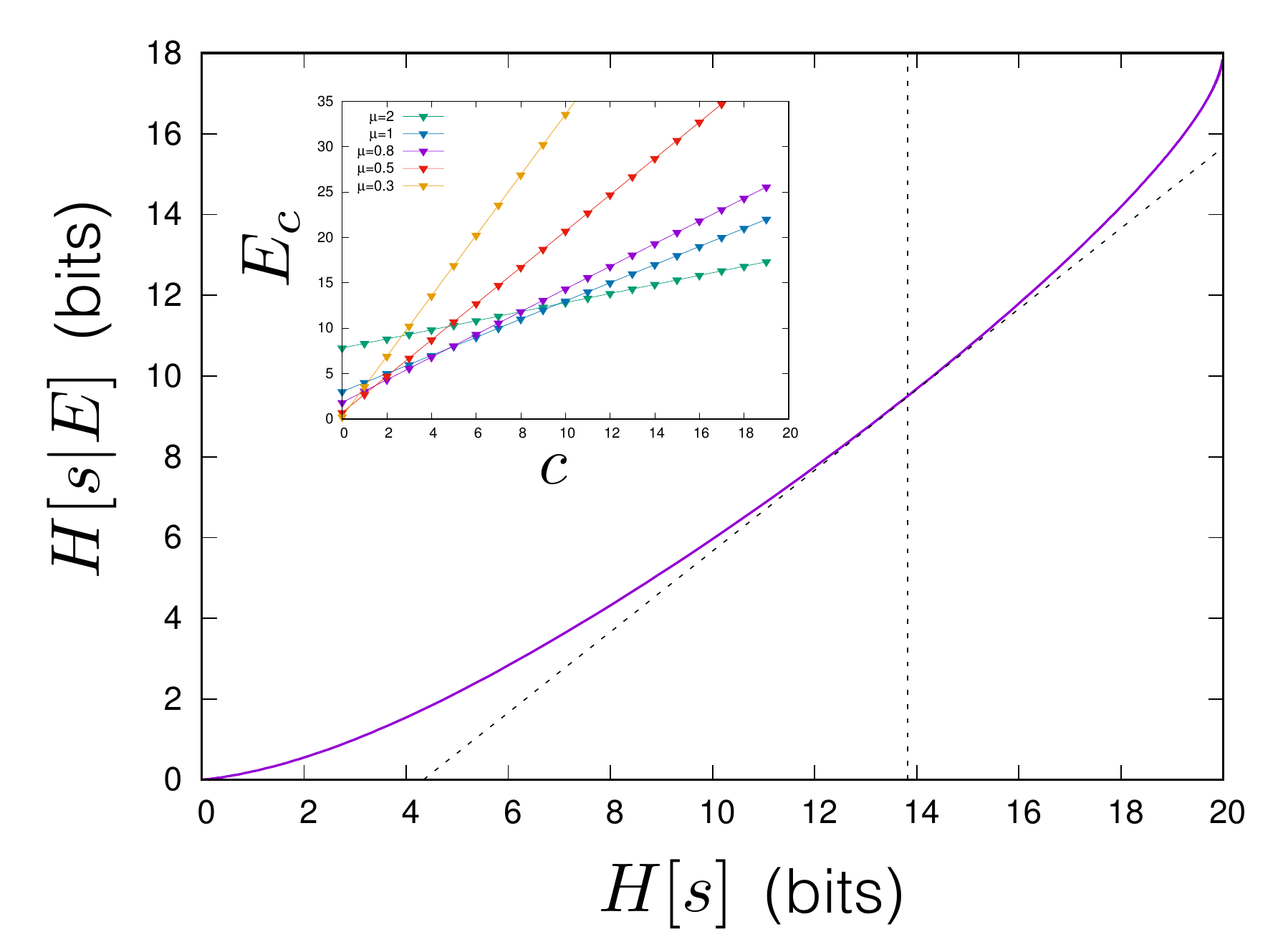}
\caption{\label{fig_OLM} Entropy ($H[\bs|E]$) versus energy ($H[\bs]$) relation for an optimal learning machine as the one discussed in the text, with $W_c=2^c$ and $c=0,1,\ldots,19$. The inset shows the energy levels for different values of $\mu$.}
\end{figure}

The linear dependence of $S(E)\simeq S_0+\mu E$ implies that the same tradeoff between resolution and relevance is at play across different energy levels of the learning machine. Eq.~(\ref{internalmu}) can be considered as a statement of {\em internal} thermodynamic equilibrium between different isolated systems at fixed energy $E$ (the coding cost). In loose words, the ``chemical potential'' $\mu$ regulates the exchange of bits between states. Let us see how this equilibrium is realised in a simple example. 

\subsubsection{\add{A toy model of an optimal learning machine}}

Imagine a machine whose states $\bs$ are grouped in classes $\mathcal{S}_c$ with $c=1,\ldots,C$, of preassigned sizes $|\mathcal{S}_c|=W_c$. All states in class $c$ have the same probability, i.e. $p(\bs)=e^{-E_c}$ for all $\bs\in\mathcal{S}_c$ and for all $c$. Therefore the distribution $p(\bs|\bs\in\mathcal{S}_c)=1/|\mathcal{S}_c|$ of $\bs$ in each class $\mathcal{S}_c$ is a state of maximal uncertainty, and its entropy $\log W_c$ provides a measure of the uncertainty of states $\bs\in\mathcal{S}_c$. In order to capture complex data, the machine should be designed so that the uncertainty $\log W_c$ spans a large enough range, so as to distinguish regular patterns from less regular ones, down to noisy data points.

How should the coding costs $E_c$ be allocated to different classes, such that the average coding cost is $H[\bs]$ and the average uncertainty 
\begin{equation}
H[\bs|E]=\E{\log W_c}=\sum_{c=1}^CW_c e^{-E_c}\log W_c
\end{equation}
is minimal?

Introducing Lagrange multipliers, to account for the constraints on resolution and on normalisation, this problem can be translated into the minimisation of the functional 
\begin{equation}
\mathcal{F}[E]=\sum_{c=1}^CW_c e^{-E_c}\log W_c-\mu\left[\sum_{c=1}^CW_c e^{-E_c}E_c-H[\bs]\right]-\nu\left[\sum_{c=1}^CW_c e^{-E_c}-1\right]
\end{equation}
with respect to $E_c$, $\mu$ and $\nu$. Differentiation with respect to $E_c$ yields the optimal coding costs $E_c=E_0+\mu^{-1}\log W_c$. This reveals that $\mu$ is the {\em chemical potential} associated to exchanges of bits across different classes. 

Since $\log W_c$ measures the uncertainty of states $\bs\in\mathcal{S}_c$, the linear behaviour $E_c=E_0+\mu^{-1}\log W_c$ states that more bits are used to code states that are more noisy, which is consistent with an efficient coding strategy. When $\mu=1$ the tradeoff between noise and coding cost is optimal, since the coding costs matches the uncertainty on all states, up to a constant. 

A linear behaviour $S(E)=S_0+\mu E$ of the entropy with the energy has been interpreted as a sign of criticality (see e.g.~\cite{MoraBialek}). This relies on the textbook definition of the specific heat as the inverse of the second derivative of $S$ with respect to $E$, and the fact that a linear relation between $S$ and $E$ implies an infinite specific heat. Yet the relation $S(E)=S_0+\mu E$ is a statement on the energy spectrum of the system, not on its thermodynamic properties. In order to discuss the thermodynamic properties, let us consider an optimal learning machine in thermal equilibrium with a heat bath at inverse temperature $\beta$. The Hamiltonian of the learning machine can then be derived from the coding cost $\mathcal{H}[\bs]=E_{\bs}/\beta$. At a different inverse temperature $\beta'$, the specific heat can be computed within the canonical ensemble in the usual way
\begin{equation}
C=k_B\left(\frac{\beta'}{\beta}\right)^2\left\langle (E-\langle E\rangle_{\beta'})^2\right\rangle_{\beta'}
\end{equation}
where averages $\langle \ldots\rangle_{\beta'}$ are computed on the distribution 
$p(\bs|\beta')=\frac{1}{Z(\beta')}W_ce^{-(\beta'/\beta)E_c}$ for all $\bs\in\mathcal{S}_c$.For an optimal learning machine with $W(E)=W_0 e^{\mu E}$, the specific heat $C$ has a maximum at inverse temperature $\beta^*=\mu\beta$, where the distribution of energies is as broad as possible. If we interpret the divergence of the specific heat as a signature of a critical system, then a learning machine at temperature $\beta$ is critical only if $\mu=1$. We shall discuss the nature of this phase transition in more detail in Section~\ref{sec:peculiar}. There we shall see that in the thermodynamic limit, the point $\mu=1$ marks a phase transition between a disordered phase -- that corresponds to noisy ($\mu>1$) representations -- and an ordered phase ($\mu<1$), that corresponds to the lossy compression regime. 

\add{The observation that the specific heat $C$ exhibit a maximum as $\beta'=\beta$ has been taken as an indication of criticality (i.e. Zipf's law, $\mu=1$) in the analysis of the neural activity of a population of neurons~\cite{retina}. Yet the maximal value of $C$ is obtained when the distribution $p(E)$ of energy levels is bimodal, as observed in~\cite{xie2021random}. Therefore, maximisation of the specific heat does not imply statistical criticality and it cannot be taken as a principle of optimality in learning. Indeed, Xie and Marsili~\cite{xie2021random} show evidence that, for a well trained RBM on the MNIST dataset, $C$ attains a maximum for a value $\beta'$ way larger than $\beta$, for which $p(E)$ develops a markedly bimodal character (see Fig.~4 in~\cite{xie2021random}). On the contrary, the relevance $H[E]$ is maximal for values $\beta'$ very close to $\beta$, corroborating the hypothesis that the principle of maximal relevance approximately holds for real learning machines.}

We also note that the distribution $p(\bs|\beta')$ also corresponds to an optimal learning machine at a different compression rate $\mu'=(\beta/\beta')\mu$. In loose words, heating or cooling an optimal learning machine also yields an optimal learning machine, with a representation at higher or lower resolution, depending on whether the temperature is increased or decreased, respectively\footnote{It is important to stress that the number of states with the same value of the Hamiltonian $\mathcal{H}[\bs]$ do not change with $\beta'$. Likewise, the number $W_c$ of states in class $c$ is independent of the temperature. The slope of the linear relation $E_c=E_0+[\log W_c]/\mu'$ instead changes, because it depends on the resolution.}. The same result can also be read through the lens of large deviation theory. 
If $p(\bs|\beta)$ is an optimal learning machine at resolution $H[\bs]$, the distribution that describes large deviations at a different resolution $H'[\bs]$ is also an optimal learning machine given by $p(\bs|\beta')$, where $\beta'$ is adjusted so that the expected value of $-\log p(\bs|\beta')$ equals $H'[\bs]$. \add{In loose words, the manifold of optimal learning machines is invariant under compression (i.e. changes in resolution).} 

\subsection{Relation with maximally informative samples}. 
\label{sec:correspondance}

The exponential density of states $W(E)=W_0 e^{\mu E}$ of optimal learning machines and statistical criticality $m_k\sim k^{-\mu-1}$ of maximally informative samples are different facets of the same optimality principle. Indeed, a sample of $N$ observations of the states of an optimal learning machine correspond to $N$ independent draws from $p(\bs)$. The number $km_k$ of states observed $k$ times should be proportional to the probability $W(E)e^{-E}\Delta E$ of states with energy in an interval $\Delta E\propto\Delta[-\log (k/N)]\propto 1/k$, around $E\simeq E_0 -\log k$. Hence $W(E)=W_0 e^{\mu E}$ implies $m_k\sim k^{-1-\mu}$. 

Also, the relevance $H[E]$ defined for learning machines is closely related to the relevance $\hat H[k]$ defined on a sample, as shown in~\cite{Odilon}. \add{Generally, a direct calculation of $H[E]$ is impractical, because it requires \blue{computing} the density of states $W(E)$, \blue{a formidable task except for simple machines}. Also,} the coding cost $E_{\bs}$ is in general a continuous variable, so its entropy is infinite strictly speaking. Yet upon discretising the energy spectrum in small intervals of size $\Delta$, one expects~\cite{CoverThomas} that if $\Delta$ is small enough, the entropy of the discretised variable 
\[
E_{\bs}^{(\Delta)}=\Delta e\,,\forall \bs:~e\le \frac{E_{\bs}}{\Delta}<e+1,~~~(e=0,\pm 1,\pm 2,\ldots)
\]
depends on $\Delta$ as $H[E^{(\Delta)}]\simeq h[E]-\log\Delta$, where 
\[
h[E]=-\int_{-\infty}^\infty \!dE p(E)\log p(E)
\]
is the differential entropy~\cite{CoverThomas}, and
\[
p(E)=\frac{1}{|\mathcal{S}|}\sum_{\bs\in\mathcal{S}}\delta(E-E_{\bs})\,.
\]
is the distribution density of energy levels. For an intermediate value of $\Delta$, we can estimate $H[E^{(\Delta)}]$ from a sample of $N$ independent draws of the energy $E_{\bs}$. First we note that, with a change of variables $f=e^{-E}$, the differential entropies of $f$ and $E$ stand in the relation
\begin{equation}
h[E]=-\int_0^1\! df p(f)\log\left[p(f)\left|\frac{df}{dE}\right|\right]=h[f]+H[\bs]
\end{equation}
where $H[\bs]=\E{E}$ is the average energy. The same relation holds for the variables at any precision $\Delta$, i.e. $H[E^{(\Delta)}]=H[f^{(\Delta)}]+H[\bs]$. Both quantities on the right hand side of this equation can be estimated in a sample, using the empirical distribution $\hat f_{\bs}=k_{\bs}/N$~\cite{Odilon}. Specifically, $H[f^{(\Delta)}]\approx \hat H[k]$ can be estimated by the relevance of the sample, and $H[\bs] \approx\hat H[\bs]$.
Taken together, these relations imply that, at the precision $\Delta$ afforded by a sample of $N$ points, we have 
\begin{equation}
\label{HEsample1}
H[E^{(\Delta)}]\approx\hat H[k]+\hat H[\bs]\,.
\end{equation}
We remark that both $\hat H[k]$ and $\hat H[\bs]$ are biased estimates of the entropy, specially in the under-sampling regime~\cite{entropy_est,NemenmanCoinc}. Yet, even if these equation do not provide an accurate quantitative estimate of $H[E^{\Delta}]$, they are sufficient to establishing whether a learning machine is close to a state of maximal relevance. This can be done  comparing the values computed from a sample $\hat \bs$ drawn from $p(\bs)$, with the theoretical maximal value of $\hat H[k]+\hat H[\bs]$ that can be achieved in a sample of $N$ points\add{, as we'll discuss later (see Fig.~\ref{fig_RBM})}. 

\subsection{Representing data as typical samples}

Thus far, our focus has been on the properties of the internal representation $p(\bs)$ of the learning machine. Let us now relate the properties we have discussed to the properties of the data $\hat x$ that the learning machine is supposed to learn. Each point $\vec x$ of the training dataset is assumed to be an independent draw from an unknown distribution $\bp(\vec x)$ -- the generative model -- that the learning machine is approximating as $p(\vec x)=\sum_{\bs}p(\vec x|\bs)p(\bs)$. \add{In general,} the training data is very high-dimensional, i.e. $\vec x\in \mathbb{R}^d$ with $d\gg 1$, but it is characterised by a significant structure of statistical dependencies. This manifests in the fact that the data\add{'s relevant variation spans} a low dimensional manifold of {\em intrinsic dimension} $d_{\rm int}\ll d$. \add{For example, Ansuini {\em et al.}~\cite{ansuini2019intrinsic} estimate that the MNIST dataset of handwritten digits spans a $d_{\rm int}\approx 13$ dimensional manifold, in spite of the fact that each data point $\vec x$ is a $d= 784$ dimensional vector}. 

\add{Here we review the arguments of Cubero {\em et al.}~\cite{statcrit} that show how this situation generically leads to an exponential density of states $W(E)=W_0 e^E$ (i.e. $\mu=1$). First we observe that the} learning machine approximates the generative process dividing it in two steps: each $\vec x$ is generated  {\em i)} by drawing an internal state $\bs$ from $p(\bs)$ and then  {\em ii)} by drawing an output $\vec x$ from $p(\vec x|\bs)$. The \add{relevant variation in} the structure of the training data is captured by $p(\bs)$, whereas $p(\vec x|\bs)$ generates ``typical'' $\bs$-type outputs $\vec x$. In other words, two draws from $p(\vec x|\bs)$ are expected to differ only by uninteresting details whereas a draw from $p(\vec x|\bs)$ typically differs significantly from a draw from $p(\vec x|\bs')$ for $\bs'\neq \bs$. 
This means that $\vec x$ conditional \blue{on} $\bs$ can be considered as a vector of weakly interacting random variables.

\add{Weakly dependent random variables generally satisfy the {\em Asymptotic Equipartition Property} (AEP), which states that the probability of typical outcomes is inversely proportional to their number\footnote{Let us briefly remind the statement of the AEP. The AEP is a direct consequence of the law of large numbers. For the simplest case of a vector $\vec x\in\mathbb{R}^d$ where each component $x_a$ is independently drawn from a distribution $p(x)$, the logarithm of the probability satisfies
\[
\frac 1 d \log p(\vec x)=\frac 1 d \sum_{a=1}^d\log p(x_a)\simeq \sum_x p(x)\log p(x)=-H[x]
\]
asymptotically, when $d\to\infty$. Hence, \add{for any $\epsilon>0$,} with probability very close to one a vector $\vec x$ drawn from $p(\vec x)$ belongs to the set of typical points 
\[
\mathcal{A}=\left\{\vec x:~\left|-\frac 1 d \log p(\vec x)-H[x]\right|<\epsilon\right\},
\]
\add{asymptotically as $d\to\infty$.}
Since all points in $\mathcal{A}$ have the same probability $p(\vec x)\simeq e^{-dH[x]}$, and $\sum_{\vec x\in\mathcal{A}} p(\vec x)\simeq 1$, then the number of typical points must equal the inverse of the probabilities of the typical points, i.e. $|\mathcal{A}|\sim e^{dH[x]}$.}~\cite{CoverThomas}. Specifically, if }
$p(\vec x|\bs)$ satisfied the AEP, a data point $\vec x$ drawn from $p(\vec x|\bs)$ belongs to the set of $\bs$-typical points
\[
\mathcal{A}_{\bs}=\left\{\vec x:~\left|-\log p(\vec x|\bs)-h_s\right|<\epsilon\right\},\qquad h_s=-\sum_{\vec x}p(\vec x|\bs) \log p(\vec x|\bs)
\]
with probability close to one. As a consequence, the number of $\bs$-typical points 
is inversely proportional to their probability, i.e. $|\mathcal{A}_{\bs}|\sim 1/p(\vec x|\bs)\simeq e^{h_{\bs}}$. 
Let us assume that also the distribution
\[
p(\vec x|E)=\sum_{\bs:~E_{\bs}=E}p(\vec x|\bs)p(\bs)
\]
satisfies the AEP. Then a draw from $p(\vec x|E)$ is with high probability an $E$-typical point that belongs to the set
\[
\mathcal{A}_{E}=\left\{\vec x:~\left|-\log p(\vec x|E)-h_E\right|<\epsilon\right\},\qquad h_E=-\sum_{\vec x}p(\vec x|E) \log p(\vec x|E)\,,
\]
and the number of $E$-typical points 
is inversely proportional to their probability, i.e. $|\mathcal{A}_{E}|\sim 1/p(\vec x|E)\simeq e^{h_{E}}$. 

An $E$-typical point is also $\bs$-typical for some $\bs$ such that $E_{\bs}=E$. This means that, for this $\vec x$, 
\begin{equation}
\label{pAEP}
p(\vec x|E)\approx p(\vec x|\bs)p(\bs)
\end{equation}
At the same time the number of $E$-typical samples must equal the number of $\bs$-typical samples, times the number $W(E)$ of states $\bs$ with $E_{\bs}=E$, i.e.
\begin{equation}
\label{qAEP}
|\mathcal{A}_{E}|\approx |\mathcal{A}_{\bs}|W(E)\,.
\end{equation}
If the left hand sides and the first factor of the right hand sides of Eqs.~(\ref{pAEP},\ref{qAEP}) are inversely proportional to each other\add{ (i.e. $|\mathcal{A}_{E}| p(\vec x|E)\simeq |\mathcal{A}_{\bs}| p(\vec x|\bs)\simeq 1$)}, then $p(\bs)=e^{-E}$ has to be inversely proportional to $W(E)$, i.e. $W(E)=W_0 e^E$. 

This derivation clarifies the peculiarity of the $\mu=1$ case\footnote{\add{As explained earlier, large deviation theory allows us to explore representations with $\mu\neq 1$}.}. This characterises optimal learning machines which interpret the training data as representative of a typical set, and sub-divides the output space during training into (approximately) non-overlapping $\bs$-typical sets. This picture is reminiscent of the tradeoff between separation and concentration discussed in Ref.~\cite{zarka2020separation}. 

\add{In summary, the inverse proportionality between the probability $p(\bs)=e^{-E}$ and the degeneracy $W(E)$ is a consequence of the AEP. It also holds for models of weakly interacting variables such as maximum entropy models of statistical mechanics. The key point is whether this relation holds on a narrow range of $E$, as in statistical mechanics, or whether it holds on a broader range. This is precisely what the relevance $H[E]$ quantifies. Odilon {\em et al.}~\cite{Odilon} study spin models with pairwise interactions such as those in Fig.~\ref{FigSpin} and they show that $H[E]=h_E\log n$ grows with the logarithm of the number of spins ($n$), asymptotically as $n\to\infty$. The proportionality constant is $h_E=1/2$ for a mean field ferromagnet away from the critical point, whereas at the critical point $h_E=3/4$. This, together with the assumption that $H[E]$ quantifies learning performance, agrees with the observation that critical models have superior learning capabilities~\cite{bertschinger2004real,roli2018dynamical,beggs2008criticality}. The proportionality constant $h_E$ can attain larger values, for spin models with different architectures (e.g. $h_E=5/4$ or $h_E\approx 1.5$, see Fig.~\ref{FigSpin}).} 

\add{This brings us to the key question of what induces a broad distribution of $E$ in learning machines. We shall address this question in the next Section. In brief, what distinguishes qualitatively a learning machine from a model of inanimate matter described by statistical mechanics, is that the data with which the machine is trained is characterised by a rich structure, which is generically described by a significant variation of hidden features. This variation is what induces a broad distribution of coding costs $E$, as also suggested in Refs.~\cite{ACL,MSN}. As a final remark, we note that $W(E)=W_0 e^E$ implies an uniform distribution of coding costs $p(E)={\rm const}$. In other words, coding costs satisfy a principle of maximal ignorance (or entropy).}


\subsection{The relevance and hidden features}

As discussed above, the structure of statistical dependence of the data manifests in the fact that the points $\vec x$ in the training set span a manifold whose intrinsic dimensionality $d_{\rm int}$ is much lower than that of $\vec x$. This makes it possible to describe the  structure of statistical dependencies of the data in terms of {\em hidden features}, than can be thought of as a set of coordinates $\bphi(\vec x)=(\bphi_1(\vec x),\ldots,\bphi_{d_{\rm int}}(\vec x))$ that describes the variation in the dataset along a low dimensional manifold. 
As for $\bp$, the backslash indicates that  the hidden features $\bphi$ are unknown. The aim of statistical learning is to find a statistical model $p(\bs)$ over a discrete variable $\bs$, and a mapping $p(\vec x|\bs)$ from $\bs$ to $\vec x$, such that the marginal distribution $p(\vec x)=\sum_{\bs}p(\vec x|\bs)p(\bs)$ approximates well the unknown $\bp(\vec x)$. Duranthon {\em et al.}~\cite{Odilon} argue that, if the learning machine captures correctly the structure of the data $\vec x$ with which it has been trained, then it must {\em extract} features $\phi$ that approximate well the hidden features $\bphi$. The {\em extracted} features $\phi(\bs)$ are defined on the states of the hidden layers, so they are in principle accessible\footnote{\add{The features defined on $\bs$ can, in principle, correspond to  features $\phi(\vec x)=\sum_{\bs}\phi(\bs)p(\bs|\vec x)$ in the input space $\vec x$.}}. Whatever they might be, Duranthon {\em et al.}~\cite{Odilon} argue that $E_{\bs}$ must be a function of $\phi(\bs)$, because states $\bs$ and $\bs'$ with the same value of $\phi(\bs)=\phi(\bs')$ should differ only by irrelevant details, so they should have the same probability\footnote{Put differently, the distribution over states with the same values of $\phi$ should encode a state of maximal ignorance, i.e. a maximum entropy distribution, \add{i.e. $p(\bs)$ and hence $E_{\bs}$ should be a constant over these states}.}. The data processing inequality \cite{CoverThomas} then implies that 
\begin{equation}
\label{dpi}
I(\bs,\phi)\ge I(\bs,E)=H[E],
\end{equation}
where the last equality comes from the fact that $I(\bs,E)=H[E]-H[E|\bs]$ and $H[E|\bs]=0$. Therefore, $H[E]$ provides a lower bound to the information $I(\bs,\phi)$ that the internal state of the machine contains on the extracted features.

The inequality (\ref{dpi}) provides a rationale for the principle of maximal relevance Eq.~(\ref{maxrel}), because it implies that statistical models with a high value of $H[E]$ are natural candidates for learning machines that efficiently extract information from data with a rich structure. 
Notice that the extracted feature $\phi$ are hardly accessible in practice, let alone the hidden ones $\bphi$. The left hand side of the inequality~(\ref{dpi}) is not easily computable. The distribution $p(\bs)$ instead can be computed for learning machines trained on structured data, and hence $H[E]$ can be estimated\add{, as shown in the next Section}.

The inequality~(\ref{dpi}) allows us to give an alternative \add{formal} definition of optimal learning machines, as those \add{models that achieve} the most compressed representation of the data, while extracting at least $H[E]$ bits of information on the features, i.e.
\begin{equation}
\label{optcompr}
\min_{p(\bs):~I(\bs,\phi)\ge H[E]}H[\bs].
\end{equation}
The relevance $H[E]$ as a function of $H[\bs]$ generally features a maximum (see e.g. Fig.~\ref{FigSpin}). The principle in Eq.~(\ref{optcompr}) only reproduces the left part of this curve, i.e. the part where $H[E]$ increases with $H[\bs]$. In optimal learning machines this corresponds to the lossy compression regime $\mu\le 1$.

\subsection{Real versus Optimal Learning Machines}

The hypothesis that real learning machines approximate the ideal limit of maximal relevance has been tested in Refs.~\cite{Odilon,SMJ}. Song {\em et al.}~\cite{SMJ} show that the internal representation in different layers of deep learning machines exhibit statistical criticality as predicted by the maximum relevance hypothesis. The density of states $W(E)$ for real machines is computationally unaccessible, because the number of states increases exponentially with the number $n$ of hidden units. Yet one can sample the internal states clamping\footnote{\add{A {\em clamped} state $\bs(\vec x)={\rm arg}\max_{\bs} p(\vec x,\bs)$ is the most likely internal state that corresponds to a given input $\vec x$.}} the visible units to the $N$ inputs of the training set. The sample of $N$ internal \add{(clamped)} states obtained in this way are a projection of the training set on the hidden layer(s). The principle of maximal relevance predicts that {\em i)} the relevance should attain values close to the maximal one at the corresponding resolution, and that {\em ii)} the number of states observed $k$ times should exhibit statistical criticality, i.e. $m_k\sim k^{-\mu-1}$, with an exponent $\mu$ which is related to the slope of the $\hat H[k]$-$\hat H[\bs]$ curve. These features should occur when data has a rich structure, but not when the training dataset is structureless. Fig.~\ref{fig_RBM}(right) reports data from Song {\em et al.}~\cite{SMJ} that support both these predictions, for Deep Belief Networks (DBN). The theoretical prediction of the exponent $\mu$ is in good agreement with the observed scaling behaviour $m_k\sim k^{-\mu-1}$ for $\mu\approx 1$, but it overestimates (underestimates) it for shallow (deep) layers. Song {\em et al.}~\cite{SMJ} show that this qualitative agreement with the predictions of the maximum relevance principle extends to several architectures, including variational auto-encoders, convolutional networks and multi-layer perceptrons \add{(see also~\cite{SongThesis})}. The same was shown by Duranthon {\em et al.}~\cite{Odilon} for Restricted Boltzmann Machines (RBM) with a varying number of hidden units, as shown in Fig.~\ref{fig_RBM}(left). When learning machines are trained with structureless data, the internal representation does not converge to states of maximal relevance (see e.g. right panel of Fig.~\ref{fig_RBM}).

\begin{figure}[ht]
\centering
\includegraphics[width=\textwidth,angle=0]{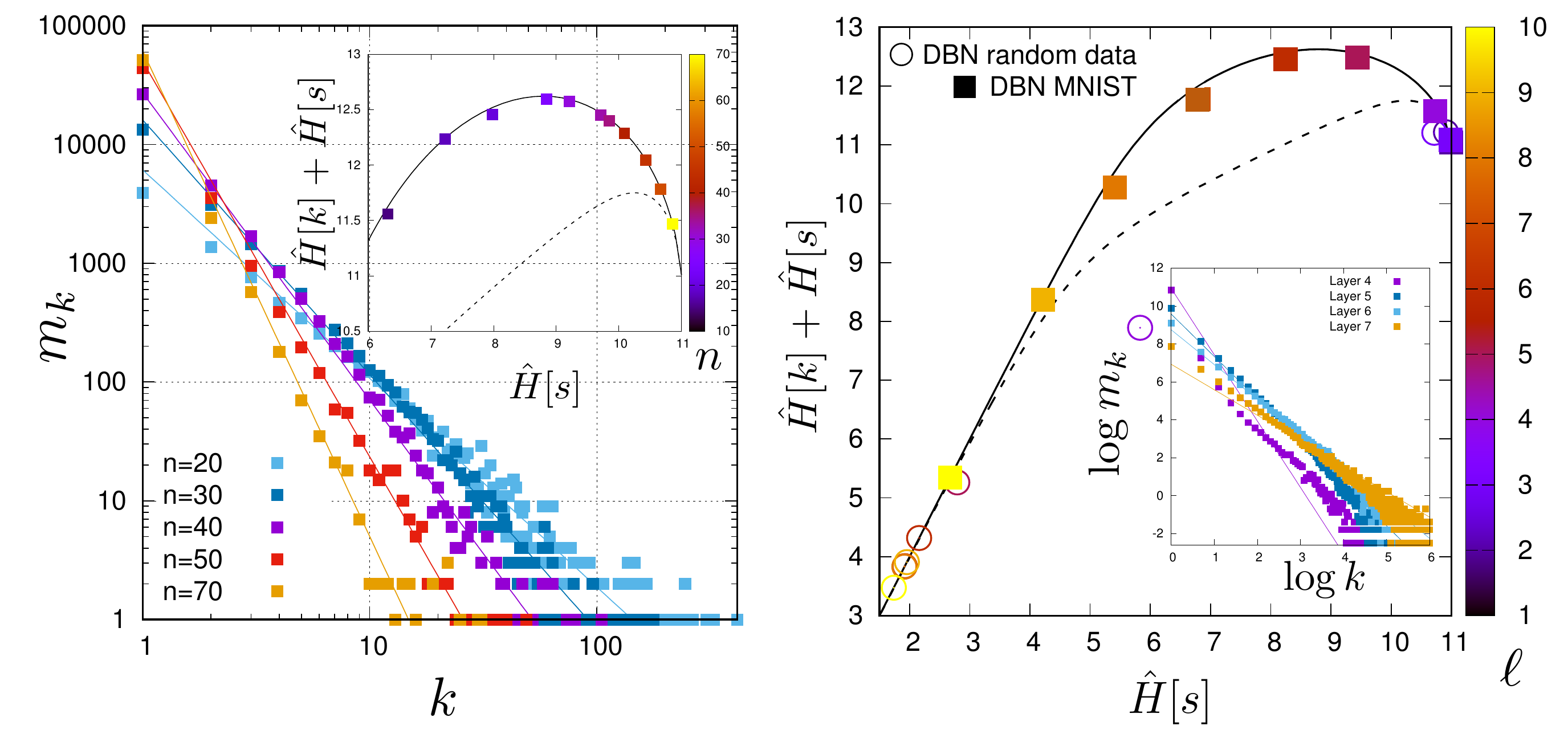}
\caption{\label{fig_RBM} {\bf Left:} Statistical criticality of the internal representation for RBMs with $n=20,30,40,50$ and $70$,  trained on the reduced MNIST dataset studied in Ref.~\cite{Odilon}. The lines correspond to the behaviour $m_k\sim k^{-\mu-1}$ where the exponent is derived from the slope $1-\mu$ of the curve $H[E]\approx \hat H[k]+\hat H[\bs]$ as a function of $H[\bs]\approx\hat H[\bs]$, shown in the inset (for $n=15,18,20,25,30,33,35,40,45,50$ and $70$ hidden units). The points approach closely the (full) line of maximal relevance.
Both $m_k$ and the values of $\hat Hp[k]$ and $\hat H[\bs]$ are computed from a sample of ``clamped'' internal states, that are sampled fixing the visible units to the data used in training. {\bf Right:} Relevance - resolution plot for samples of clamped states for the different layers of a DBN with 10 layers trained on the MNIST dataset (filled squares) and for a reshuffled MNIST dataset (open circles). The  data is the same used in Ref.~\cite{SMJ}, to which we refer for more details. The inset shows the frequency distributions for different layers and their comparison with the theoretically predicted behaviour $m_k\sim k^{-\mu-1}$.}
\end{figure}

Duranthon {\em et al.}~\cite{Odilon} also explore different architectures and provide evidence in support of the hypothesis that, in a given learning task, the maximal values of the likelihood are achieved by models with maximal relevance. Interestingly, they show that the relevance for Gaussian learning machines~\cite{karakida2016dynamical} does not vary during training. Indeed the distribution of energy levels does not broadens during training, but it merely shifts maintaining the same shape. This is consistent with the fact that Gaussian learning machines do not learn anything on the generative model $\bp$ (beyond its parameters), because the shape of 
the learned distribution $p(\bs)$ remains a Gaussian, irrespective of the generative model $\bp$, throughout the learning process.
Instead in learning machines such as RBM learning is associated with a remarkable broadening of the spectrum of energy levels of the internal representation, in agreement with the maximum relevance hypothesis.

Finally, Duranthon {\em et al.}~\cite{Odilon} give evidence of the fact that the features that confer superior learning performance (i.e. higher relevance) to a statistical model may be associated to sub-extensive features of the model that are not accessible to standard statistical mechanics approaches. Also, a superior learning performance is not necessarily related to the existence of a critical point separating two different phases, but when a critical point exists, learning performance improves when the model is tuned to its critical point.

\section{The statistical mechanics of learning machines}
\label{sec:peculiar}

\add{Why should systems that learn be characterised by} 
an exponential distribution of energies $E_{\bs}=-\log p(\bs)$? Which thermodynamic properties does distinguish them from systems that describe inanimate matter? In order to address these questions, following Refs.~\cite{MMR,xie2021random,marsili2019peculiar}, we shall first formulate the problem in terms of a generic optimisation problem, and then study the statistical mechanics properties of ensembles of solutions of such a problem where the objective function is assumed to be randomly drawn from a distribution, in the spirit of Random Energy Models~\cite{REM}. 

\subsection{A generic class of optimisation problems}

\begin{figure}[h]
\centering
\includegraphics[width=0.9\textwidth,angle=0]{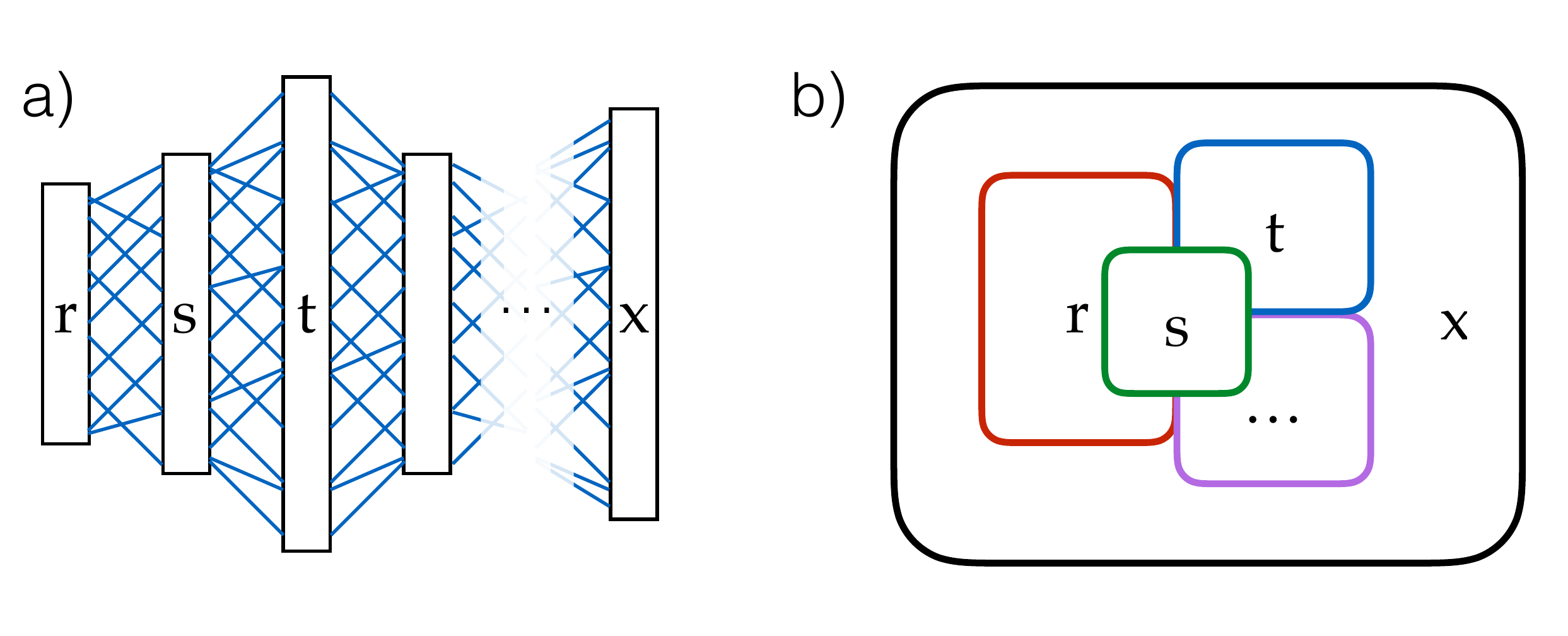}
\caption{\label{fig_OLMs} A sketch of a learning machine (left) and of a complex system interacting with its environment.}
\end{figure}

Let us consider a generic system with an architecture such as those shown in Fig.~\ref{fig_OLMs}. On the left we have the typical architecture of a deep neural network (see e.g.~\cite{roudi2015learning} for a brief review on these architectures). Once trained on a dataset of inputs, this delivers a generative model
\begin{equation}
\label{joingen}
p(\bs,\bt,\ldots,\bz,\vec x)= p(\vec x|\bz)\cdots p(\bt|\bs)p(\bs),
\end{equation}
that relates the internal states \add{$\bs,\bt,\ldots,\bz$} of the different layers to the input $\vec x$. In Eq.~(\ref{joingen}) 
\begin{equation}
p(\bs)=\sum_{\br,\ldots}p(\bs|\br)p(\br|\ldots)\cdots
\end{equation}
is the marginal distribution of layer $\bs$, with respect to deeper layers $\br,\ldots$. The relation between internal states and inputs can be studied by considering the most likely {\em clamped} states $\bs^*(\vec x),\bt^*(\vec x),\ldots$ that the network associates to an input $\vec x$. These are the solution of the maximisation problem 
\begin{eqnarray}
\bs^* (\vec x)& = & {\rm arg}\max_{\bs}\left\{\log p(\bs)+\max_{\bt}\left[\log p(\bt|\bs)+\max_{\bu}\left(\ldots+
\max_{\bz}\log p(\vec x|\bz)p(\bz|\ldots)\right)\right]\right\} \\
 & = & {\rm arg}\max_{\bs}\left\{u_{\bs}+\max_{\bt}v_{\bt|\bs}(\vec x)\right\}\,,
 \label{optmax}
\end{eqnarray}
where $u_{\bs}=\log p(\bs)$.
For a random input $\vec x$ drawn from an unknown distribution $\bp(\vec x)$, the term 
\begin{equation}
\label{leftsyst}
 v_{\bt|\bs}(\vec x) = \max_{\bu\ldots,\bz}\log p(\bt,\bu,\ldots,\bz,\vec x|\bs) 
\end{equation}
is a random function of $\bs$ and $\bt$, because it depends on $\vec x$. When the maximisation on $\bs$ in Eq.~(\ref{optmax}) is dominated by the first term $u_{\bs}$, the solution $\bs^*$ depends weakly on $\vec x$ and we expect that with high probability $\bs^*$ 
coincides with the most probable internal state
\[
\bs_0={\rm arg}\max_{\bs} \log p({\bs}) \equiv {\rm arg}\max_{\bs}u_{\bs}
\]
irrespective of $\vec x$. When it is instead dominated by the second term, $\bs^*$ can take any possible value irrespective of $p(\bs)$. These extremes are not representative of how an efficient learning machine is expected to work. \add{In this case, we expect that the distribution of clamped states
\begin{equation}
\label{pclamp}
q(\bs)=P\{\bs^*=\bs\}
\end{equation}
to be as close as possible to the distribution $p(\bs)$ learned by the machine. It is important to appreciate the difference between these two distributions: $p(\bs)$ is the distribution by which the trained machine can generate a new data point $\vec x$, by propagating a state $\bs$ randomly drawn from $p(\bs)$ to the visible layer. $q(\bs)$ is instead the distribution induced by the data in the internal layer. Therefore, the condition $q(\bs)=p(\bs)$ is morally equivalent to the requirement that the machine approximates well the distribution of the data, i.e. $p(\vec x)\approx \bp(\vec x)$. This is only possible if statistical dependencies propagate from the visible to deep layers.}

Likewise, consider a system with the architecture of Fig.~\ref{fig_OLMs}(right) where the internal variables $\bs$ (e.g. gene expression) interact with different sets of variables $\br,\bt,\ldots,\bz$ (e.g. metabolism, protein synthesis, transport, etc)
that ultimately interact with the environment $\vec x$ \add{(e.g. nutrients, toxins, etc)}. The most likely state of the internal variables $\bs$ is the solution of the optimisation problem~(\ref{optmax}) with $u_{\bs}=\log p(\bs)$ and 
\begin{equation}
\label{rightsys}
 v_{\bt|\bs}(\vec x) = \log p(\bt|\bs)+\max_{\br}\left[\log p(\br|\bs,\bt)+\max_{\bu}\left(\ldots+
\max_{\bz}\log p(\bz|\vec x)\right)\right].
\end{equation}
In a stochastic environment $\vec x$, $v_{\bt|\bs}$ is a random function of $\bs$ and $\bt$. As in the example above, the relation $\bs^*(\vec x)$ between the clamped internal state and the environment in a system that responds efficiently to its environment, like a cell, should neither be constant nor too variable. For this to happen, such a system needs to strike a balance between the two terms in Eq.~(\ref{optmax}). 

Following Ref.~\cite{marsili2019peculiar}, we observe that in both cases, we are led to study the optimisation problem of an objective function 
\[
U(\bs,\bt)=u_{\bs}+v_{\bt|\bs}, 
\]
where $v_{\bs|\bt}$ is a random function. The system $\bt$ plays the role of a proximate environment of the system. {Marsili {\em et al.}~\cite{MMR} offer a further interpretation of this problem, where $\bs$ are the known variables and $\bt$ are unknown unknowns. Then $u_{\bs}$ is the part of the objective function of a complex system which is known, whereas $v_{\bt|\bs}$ represents the interaction with unknown variables $\bt$, and it is itself unknown.} 

\subsection{\add{A random energy approach to learning}}

\add{As in other complex systems -- ranging from heavy ions~\cite{wigner1993characteristic} and ecologies~\cite{may1972will}, to disordered materials~\cite{mezard1987spin} and computer science~\cite{monasson1999determining,zdeborova2016statistical} -- much insight can be gained by studying the typical properties of ensembles of random systems. 
This is because the collective properties of systems of many interacting degrees of freedom often only depend on the statistical properties of the resulting energy landscape. In these circumstances, an approach similar to that of the Random Energy Model~\cite{REM} for spin glasses, may be very helpful. For a learning machine the disorder is induced by the data with which the machine is trained, much in the same way as in glasses and spin glasses it is induced by the random spatial arrangement of particles.}

\subsubsection{\add{How deep can statistical dependencies reach?}}

As in Ref.~\cite{marsili2019peculiar}, we focus on the special case where $\bs=(s_1,\ldots, s_{n_s})$ and $\bt=(t_1,\ldots, t_{n_t})$ are strings of $n_s$ and $n_t$ binary variables, respectively. We assume that, for each value of $\bs,\bt$, $v_{\bs|\bt}$ is drawn independently from the distribution\footnote{For simplicity we restrict to the case where $v_{\bt|\bs}\ge 0$. Strictly speaking this is inconsistent with the definition Eq.~(\ref{leftsyst}). Yet all the derivation generalise to the case where $v_{\bt|\bs}\to v_{\bt|\bs}+v_0$ is shifted by an arbitrary constant.}
\begin{equation}
\label{stretched}
P(v_{\bs|\bt}>x)=e^{-(x/\Delta_t)^{\gamma_t}},\qquad (x\ge 0)
\end{equation}
where $\Delta_t >0$ is a constant that sets the scale of $v_{\bt|\bs}$. 
\add{The value of $\Delta_t$ determines the size of $v_{\bt|\bs}$, and hence which of the two terms dominates the optimisation problem in Eq.~(\ref{optmax}). If $\Delta_t$ is small,} the behaviour of the $\bs$-system is \add{largely independent of the environment, and we expect that $\bs^*$  coincides with the optimum $\bs_0={\rm arg}\max_{\bs} u_{\bs}$ with high probability. If instead $\Delta_t$ is large, $\bs^*$ is} dominated by the environment $\bt$, and the chances that $\bs^*=\bs_0$ will be negligible. 

These two extremes can be distinguished by the value of the entropy of the distribution \add{of the state $\bt$ of the environment, for a given state $\bs$ of the system. Assuming that all variables $\bu,\ldots,\bz$ are clamped to their maximal values, the distribution of $\bt$ is given by
\begin{equation}
P\{\bt|\bs\}=\frac 1 Z_{\bs} e^{v_{\bt|\bs}}\,.
\end{equation}
For $n_t$ large, it is possible to draw on results from random energy models~\cite{REM,xie2021random,marsili2019peculiar}.
As a function of $\Delta_t$, the entropy $H[\bt|\bs]$ of this distribution exhibits a phase transition at
\begin{equation}
\label{deltacrit}
\Delta_t^*=\gamma_t\left(n_t\log 2\right)^{1-1/\gamma_t}.
\end{equation}
For $\Delta_t<\Delta_t^*$ the entropy
\begin{equation}
H[\bt|\bs]\simeq \left[1-\left(\frac{\Delta_t}{\Delta_t^*}\right)^{\gamma_t/(\gamma_t-1)}\right] n_t\log 2
\end{equation}
is extensive, i.e. proportional to $n_t$. For $\Delta_t>\Delta_t^*$, instead, the distribution of $\bt$ is peaked on those few values for which $v_{\bt|\bs}$ is maximal, for the specific value of $\bs$, and $H[\bt|\bs]$ is finite.}


The 
intermediate maximisation of $v_{\bt|\bs}$ over $\bt$ in Eq.~(\ref{optmax}) can be carried out explicitly using extreme value theory~\cite{Galambos}, with the result
\[
\max_{\bt}v_{\bt|\bs}\simeq a_m+b_m\eta_{\bs}
\]
where $\eta_{\bs}$ follows a Gumbel distribution\footnote{\add{Note that, for large $x$, the Gumbel distribution behaves as $P(\eta_{\bs}\le x)\simeq e^{-x}$, which is the same behaviour as in Eq.~(\ref{stretched}) with $\gamma_t=1$. In loose words, in this simplified setting, extreme value theory mandates that the statistics of the interaction term between $\bs$ and the environment is universal and that it follows an exponential law.}} $P(\eta_{\bs}\le x)=e^{-e^{-x}}$, $a_m=\Delta_t (n_t\log 2)^{1/\gamma_t}$ and 
\begin{equation}
\label{eqbm}
b_m=\frac{\Delta_t}{\gamma_t}(n_t\log 2)^{1/\gamma_t -1}=\frac{\Delta_t}{\Delta_t^*}.
\end{equation}
When $m$ is large, these results~\cite{Galambos} lead to the conclusion that the distribution of clamped states $\bs^*$ is given by~\cite{marsili2019peculiar} 
\begin{eqnarray}
q(\bs) & = & P\left\{u_{\bs}+\max_{\bt}v_{\bt|\bs}\ge u_{\bs'}+\max_{\bt}v_{\bt|\bs'},~\forall\bs'\right\} \\
& = & P\left\{\eta_{\bs'}\le \eta_{\bs}+\beta(u_{\bs}- u_{\bs'}),~\forall\bs'\right\} \label{GumbelEVT}\\
 & = & \frac{1}{Z}e^{\beta u_{\bs}},\qquad Z=\sum_{\bs}e^{\beta u_{\bs}}
\label{GibbsEVT}\\
 & = & \frac{1}{Z} [p(\bs)]^\beta.
\label{maxGB}
\end{eqnarray}
Note that Eq.~(\ref{GibbsEVT}) is a Gibbs-Boltzmann distribution\footnote{In order to derive Eq.~(\ref{maxGB}), note that $\eta_{\bs}$ are independent random variables with Gumbel distribution. This makes it possible to compute Eq.~(\ref{GumbelEVT}) explicitly. See~\cite{MMR,marsili2019peculiar} for details.} with an  ``inverse temperature''
\begin{equation}
\label{eqbeta}
\beta=\frac{1}{b_m}=\frac{\Delta_t^*}{\Delta_t}.
\end{equation}
\add{Therefore, statistical dependencies can propagate to layer $\bs$ only if $\beta=1$, i.e. if the parameter $\Delta_t$ is tuned to its critical point $\Delta_t^*$. Since this holds for all layers, we conclude that statistical dependencies can propagate across layers only if all layers are tuned at the critical point. This is a conclusion similar to the one Schoenholz {\em et al.}~\cite{schoenholz2016deep} arrived at, studying the propagation of a Gaussian signal across layers in a random neural network.}

\add{Note that, for a fixed value of $\Delta_t$, as the number $n_t$ of variables in the shallower layer increases, $\beta$ increases for $\gamma_t>1$, whereas it decreases for $\gamma_t<1$. Only for $\gamma_t=1$ we find that $\beta$ is independent of $n_t$. We shall return to this point later.}

\add{In the general case of a system $\bs$ which is in contact with more than one different subsystems $\br,\bt,\ldots$ as in Fig.~\ref{fig_OLMs}(right), the same argument suggests that all proximate ``environments'' $\bt,\br,\ldots$ with which a system is in contact with should be tuned to a critical point, in order for information on the state $\vec x$ of the outer ``environment'' to be transferred efficiently.}

\subsubsection{\add{How does a trained machine differs from an untrained one?}}

\add{The same random energy approach can be applied to the internal state $\bs$ of a learning machine. Specifically, Refs.~\cite{xie2021random,marsili2019peculiar} assume that $u_{\bs}$ is drawn independently from a distribution 
\[
P\{u_{\bs} \le x\}=e^{-(x/\Delta_s)^{\gamma_s}}, 
\]
independently for each $\bs$. For a given value of $\gamma_s$, the properties of the system depend on the single parameter $\Delta_s$. As a function of the parameter $\Delta_s$, the internal representation $p(\bs)$ undergoes a phase transition, like the one of the Random Energy Model~\cite{REM}, between a disordered phase and a frozen phase, at a critical value $\Delta_s^*=\gamma_s(n_s\log 2)^{1/\gamma_s-1}$, in analogy with Eq.~(\ref{deltacrit}). The properties of the phase transition will be discussed in the next subsection, in the limit $n_s\to\infty$.}

\add{As compared to the behaviour of real learning machines, Ref.~\cite{xie2021random} estimates the parameter $\Delta_s$ by matching the entropy $H[\bs]$ of the internal representation of the machine to that of a Random Energy Model where $n_s$ equals the number of hidden (binary) variables. This shows that, irrespective of the value of $\gamma_s$, well trained learning machines such as RBMs and internal layers of DBNs are described by a value $\Delta_s\approx\Delta_s^*$ which is close to the critical point. By contrast, both untrained machines and machines trained on randomised (structureless) data are best fitted by Random Energy Models with a parameter $\Delta_s$ which is larger than $\Delta_s^*$. This confirms the relation between criticality and learning discussed in previous Sections. The distinguishing feature of a well trained machine is that its energy spectrum is as broad as possible, which occurs in Random Energy Models at the phase transition $\Delta_s^*$.}

\add{Interestingly, the logarithm of the number of states $\bs$ with $-\log p(\bs)=E$ 
\[
S(E)=\log W(E)\simeq n\log 2-(E/\Delta_s)^\gamma
\]
is not {\em a priori} linear, unless $\gamma_s=1$. Yet, when the parameter $\Delta_s$ is adjusted to the critical point as a consequence of learning, an approximately linear dependence of $S(E)\simeq E-E_0$ arises in the range of observable values of $E$. This range  extends to an interval of size $\delta E\sim \sqrt{n}$ for $\gamma_s\not =1$, whereas in the special case $\gamma_s=1$ it extends over a range $\delta E\sim n$. Systems with $\gamma_s=1$ are therefore special, as also shown by the analysis of their thermodynamic properties, to which we now turn, following the discussion of Ref.~\cite{marsili2019peculiar}.}

\subsection{The thermodynamic limit}

Let us now consider the behaviour of the $\bs$-system, \add{as described by the distribution $q(\bs)$ in Eq.~(\ref{GibbsEVT}),} in the thermodynamic limit, when both its size $n_s$ and the size of the environment $n_t$ diverge, with a finite ratio $n_t/n_s=\nu$. \add{As we argued above, well trained learning machines correspond to the case where the parameter $\Delta_t$ is tuned to its critical value. Leaving this case aside for the moment, let us first discuss the case where $\Delta_s$ and $\Delta_t$ are finite, following Ref.~\cite{marsili2019peculiar}.}

The number of states $\bs$ with $u_{\bs}\le u$ obeys a stretched exponential distribution
\begin{equation}
\mathcal{N}(u_{\bs}\le u)\simeq 2^{n_s} e^{-(u/\Delta_s)^{\gamma_s}}
\end{equation}
with exponent $\gamma_s$.
The thermodynamics, as usual, is determined by the tradeoff between the entropy and the ``energy'' term \add{$\beta u$}. The entropy 
\begin{equation}
\label{entropy_exprem}
S(u)= \log \mathcal{N}(u_{\bs}\le u) \simeq n_s\log 2\left[1-(u/u_0)^{\gamma_s}\right]
\end{equation}
is extensive in $n_s$, for values of $u$ smaller than the maximum $u_0=\max_{\bs}u_{\bs}\simeq (n_s\log 2)^{1/\gamma_s}\Delta_s$. The energy term \add{in $\log q(\bs)$} instead scales as
\begin{equation}
\beta u_0=\gamma_t\frac{\Delta_s}{\Delta_t}\nu^{1-1/\gamma_t}\left(n_s\log 2\right)^{1+\frac{1}{\gamma_s}-\frac{1}{\gamma_t}}. 
\end{equation}
Therefore for $\gamma_s<\gamma_t$\add{, in the limit $n_s\to\infty$,} the distribution $q(\bs)$ (see Eq.~\ref{maxGB}) is dominated by states with maximal $u_{\bs}$ because $\beta u_0\gg S(u)$, and $H[\bs^*]$ is finite. For $\gamma_s>\gamma_t$ the opposite happens: $q(\bs)$ is dominated by the entropy and by states with small values of $u_{\bs}$, and therefore $H[\bs^*]\simeq n_s\log 2$.

\begin{figure}[ht]
\centering
\includegraphics[width=0.9\textwidth,angle=0]{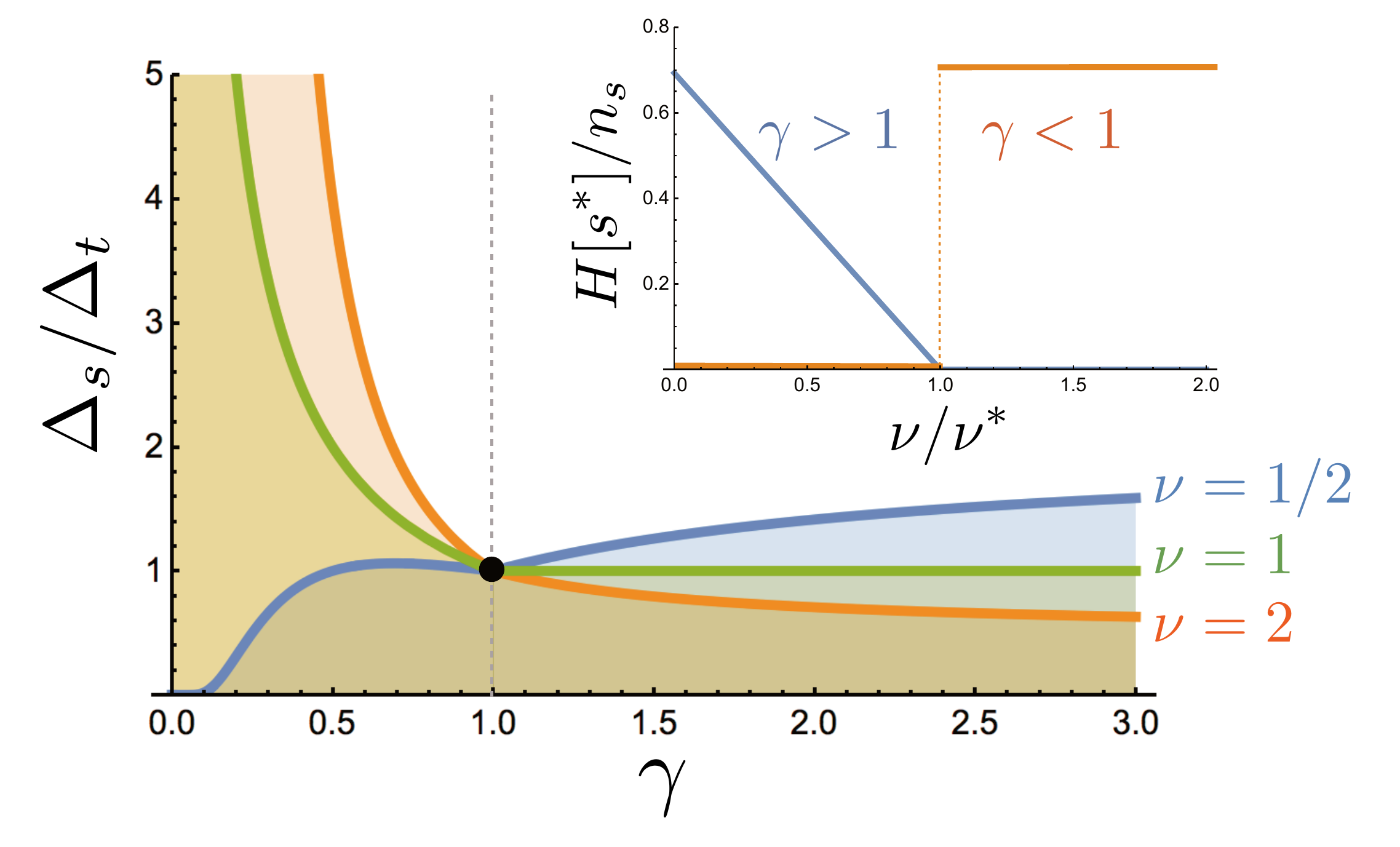}
\caption{\label{fig_ExpREM}
Phase diagram of the random optimisation problem, as a function of the three main parameters, $\gamma$ and $\Delta_s/\Delta_t$. Different lines correspond to different values of the ratio $\nu=n_t/n_s$ between the size of the environment and the size of the sub-system ($\nu=1/2, 1$ and $2$, from top to bottom for $\gamma>1$). The shaded regions below the lines correspond to the disordered (weak interaction) phase, in the three cases. The point at $\gamma=\Delta_s/\Delta_t=1$ denotes the point where Zipf's law occurs. 
The inset shows the behaviour of the entropy $H[\bs^*]$ of the clamped distribution per variable as a function of $\nu$ across the transition. The phase transition is continuous for $\gamma> 1$ and discontinuous for $\gamma<1$.}
\end{figure}

A non-trivial thermodynamic behaviour attains only when $\gamma_s=\gamma_t=\gamma$. The behaviour of the system, 
as a function of $\nu$, $\Delta_s/\Delta_t$ and $\gamma$ is summarised in Fig.~\ref{fig_ExpREM}. As discussed in Ref.~\cite{marsili2019peculiar}, the main determinant in the thermodynamic limit is the convexity of the entropy $S(u)$ in Eq.~(\ref{entropy_exprem}). 
 \begin{description}
  \item[For $\gamma>1$] the entropy is concave, as in generic physical systems. The celebrated Random Energy Model~\cite{REM}, for example, corresponds to $\gamma=2$ and it belongs to this region. The partition function is dominated by states with a value of $u$ such that the slope of the entropy equals $-\beta$, i.e. $\frac{dS}{du}=-\beta$. \add{As a function of $\nu$, the system undergoes a phase transition at a critical value
\begin{equation}
\label{nucrit}
\nu^*=\left(\frac{\Delta_s}{\Delta_t}\right)^{\gamma/(1-\gamma)}\qquad (\gamma>1)
\end{equation}
with a similar phenomenology to the one of the Random Energy Model~\cite{REM}. For $\nu<\nu^*$} the distribution of $\bs^*$ extends over an extensive number of states and the entropy
\begin{equation}
H[\bs^*]=\left(1-\frac{\nu}{\nu^*}\right)n_s\log 2,\qquad (\nu<\nu^*,~\gamma>1)
\end{equation}
is proportional to the system size $n_s$. For $\nu>\nu^*$ the system enters a localised phase, where the entropy $H[\bs^*]$ attains a finite value, as a signal that the distribution $q(\bs)$ becomes sharply peaked around the state $\bs_0$ of maximal $u_{\bs}$. The transition at $\nu^*$ is continuous. The system enters the localised phase as $\nu$ increases, as shown in the inset of Fig.~\ref{fig_ExpREM}. 
  \item[For $\gamma<1$] the entropy $S(u)$ is convex. This gives rise to a starkly different behaviour.
  A consequence of convexity is that, for all values of $\beta$, the partition function is dominated either by states with $u_{\bs}\approx 0$ or by states with $u_{\bs}\simeq u_0$. \add{As a function of $\nu$, the system undergoes a sharp phase transition at 
\begin{equation}
\nu^*= \left(\gamma\frac{\Delta_s}{\Delta_t}\right)^{\gamma/(1-\gamma)}, \qquad (\gamma<1)
\end{equation}
where the entropy suddenly jumps from $H[\bs^*]\simeq n\log 2$ for $\nu>\nu^*$ to a finite value for $\nu<\nu^*$. 
For $\nu>\nu^*$} the distribution $q(\bs)$ is asymptotically uniform on all $2^{n_s}$ states, whereas for \add{$\nu<\nu^*$} it sharply peaks on $\bs_0$. Note that, contrary to the case $\gamma>1$, the transition to a localised state occurs when the relative size of the environment $\nu$ increases (see  inset of Fig.~\ref{fig_ExpREM}).
\item[The case $\gamma=1$] is special, because the entropy $S(u)$ is linear in $u$. The thermodynamic behaviour is independent of $\nu$. The system undergoes a phase transition at $\Delta_s=\Delta_t$ between an extended ($\Delta_s<\Delta_t$) and a localised phase ($\Delta_s>\Delta_t$). 
The entropy $H[\bs^*]$ of the clamped distribution decreases as $\Delta_s$ increases in a way that is sharper and sharper as $n_s$ increases, as shown in Fig.~\ref{figExpREMgamma1}(left).  
\end{description}

\begin{figure}[ht]
\centering
\includegraphics[width=\textwidth,angle=0]{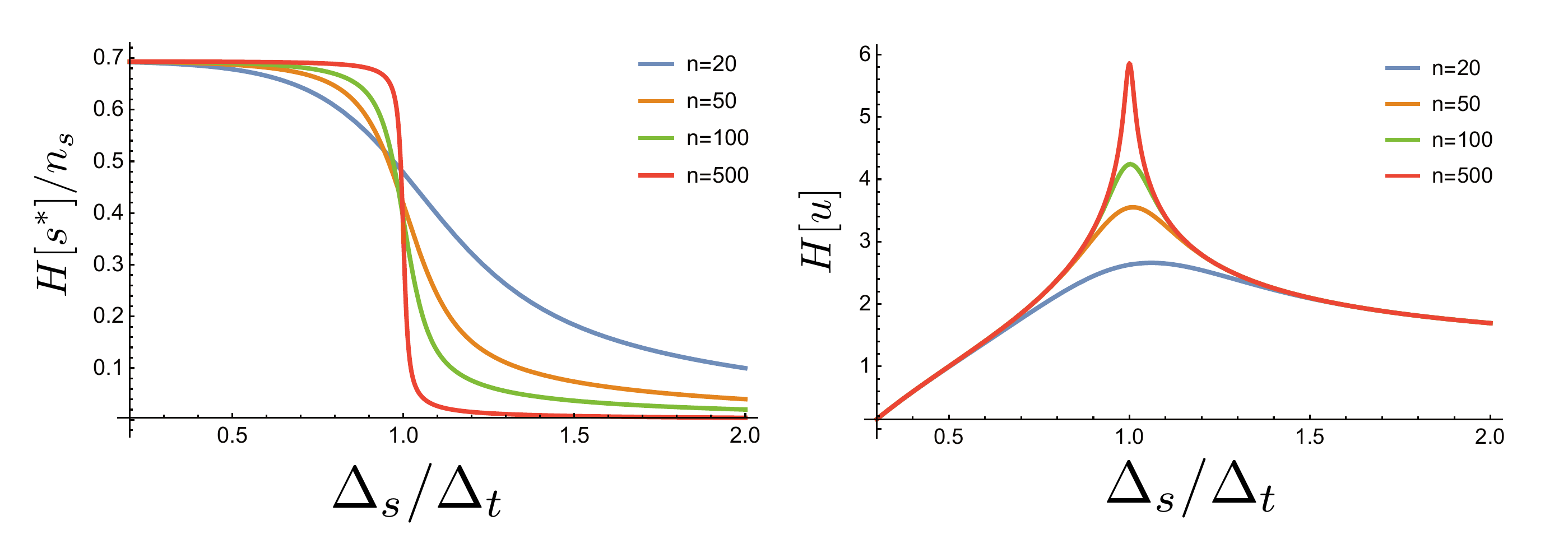}
\caption{\label{figExpREMgamma1} Phase transition at $\Delta_s/\delta_t=1$ for $\gamma=1$. Left: behaviour of the entropy $H[s]/n_s$ for different values of $n_s=20,50,100$ and $500$, as a function of $\Delta_s/\Delta_t$. Right: relevance $H[u]$ vs $\Delta_s/\Delta_t$ for the same values of $n_s$.}
\end{figure}

\add{With the identification $E=E_0-\beta u$, where $E_0$ is a constant, it is easy to see that the optimal learning machines discussed in Section~\ref{sec:OLMprop}} correspond to the case $\gamma=1$, with $\mu=\Delta_t/\Delta_s$. 
The phase transition point corresponds to the point $\mu=1$ of the optimal most compressed lossless representation of the data in optimal learning machines, and to Zipf's law in maximally informative samples. The analog of the relevance is given by the entropy $H[u]$ which, as shown in Fig.~\ref{figExpREMgamma1}(right), exhibits a singularity when $\Delta_s=\Delta_t$. 

\add{As applied to a multi-layer learning machine, the invariance of the distribution of $\bs^*$ on $\nu$ is suggestive. It implies that, as long as the distribution of both $u_{\bs}$ (deeper layer) and $v_{\bt|\bs}$ (shallower layers) are exponential, the relative size of the layers is irrelevant. It is tempting to relate this property to the requirement that,} in a classification task, we should expect that the internal representations in sufficiently deep layers should be the same, irrespective of whether the machine is trained with high resolution or with low resolution data (e.g. digital pictures). \add{If $n_t$ increases with the dimensionality $d$ of the input $\vec x$, this requirement is met only when $\gamma=1$, within the {simplified picture} discussed here.}

\add{Finally, the case $\gamma=1$ is also representative of the situation where $\Delta_s$ and $\Delta_t$ are tuned at their respective critical points for generic values of $\gamma_s,\gamma_t> 1$. Indeed, as discussed earlier, at the critical point the density of states is approximately exponential in this case.}

\bigskip

In summary, optimal learning machines ($\gamma=1$) enjoy very peculiar statistical mechanics properties. They sit at the boundary between systems with properties that are similar to those of generic physical systems ($\gamma>1$) and systems with unphysical properties ($\gamma<1$). 

\section{Conclusions}
\label{sec:conclusions}

\begin{quote}
These words I wrote in such a way that a stranger does not know. \\
You too, by way of generosity, read them in way that you know.\\
\centerline{\em The Divan of Hafez}
\end{quote}

\blue{The main objectives of this review are to point out that the concept of relevance can be quantitatively and universally defined and is a key ingredient in learning}, specially in the under-sampling regime. \blue{The under-sampling regime, as we refer to it here, is the regime in which learning or inference should be performed with high-dimensional but scarce, relative to this dimension, data and when the generative model is largely unknown.} We provide evidence that the principle of maximal relevance provides an unifying foundation for understanding efficient representations both in data and in learning machines. This principle, for example, \add{provides a rationale for} the ubiquitous appearance of heavy-tailed distributions in \blue{natural} systems~\cite{newman2005power,munoz2018colloquium}. \blue{As shown in Section \ref{sec:stat-crit} power law distributions arise when the sampled variables are maximally relevant, and consequently, as shown in this review, maximally informative} about their generative process. The principle of maximal relevance, \blue{as discussed in Section \ref{sec:maxrel-crit}}, also sheds light on the origin of criticality, by identifying the order parameter of the transition with the compression rate (the resolution), and by clarifying the nature of the phase transition in terms of the mode collapse phenomenon. As applied to learning machines, the principle of maximal relevance \blue{was discussed in Section \ref{sec:learning}, and it} offers a generic argument for the observation~\cite{langton1990computation,bertschinger2004real} that their performance is optimal} when they're tuned to the critical point. \add{We argue that criticality is necessary for the efficient propagation of information across different subsystems}. We devote this concluding Section to discuss avenues for application or further extension of the results discussed so far.

\bigskip

A lot of progress has been done in understanding statistical learning in the high-dimensional limit\blue{; see e.g. \cite{wainwright2019high}}. Statistical mechanics provides a fairly detailed account of the statistical properties of systems such as Hopfield models of associative memory \cite{Hopfield}, neural networks in the teacher-student paradigm (see e.g.~\cite{engel2001statistical}) or network reconstruction task in Boltzmann learning machines (see e.g.~\cite{roudi2009statistical,nguyen2017inverse}). Several attempts have been made to characterise the statistical mechanics properties of more complex learning machines, such as RBMs or multi-layer perceptrons~\cite{tubiana2017emergence,decelle,zecchina,hennig,biroli}. Yet a satisfactory understanding is still elusive and the spectacular performance of learning machines such as deep neural networks still remains puzzling.

One reason for this failure is that early attempts~\cite{engel2001statistical,roudi2009statistical,nguyen2017inverse} have  focused on learning tasks from synthetic data or in the teacher-student scenario which are rather atypical as compared to real data. \add{Only recently these approach have been generalised to data with a rich structure~\cite{mezard,goldt2019modelling,gherardi2020,zdeborova2020understanding}, such as the one sampled from real systems}. 
\add{Furthermore, both in supervised or unsupervised settings, learning is defined by encoding the task (regression, classification, clustering, etc.) into an objective function~\cite{barberBRML2012}, turning learning into an optimisation problem (e.g. maximum likelihood or error minimisation). Statistical mechanics approaches necessarily} depend on the objective function assumed, on the architecture and on the statistical model of the data~\cite{tubiana2017emergence,bulso2021restricted}. This makes it hard to generalise conclusions beyond the specific setting considered. 


Furthermore, all these approaches are problematic in the under-sampling regime. \add{When data is high-dimensional and scarce, practical progress is possible only under strong assumptions (linearity, Gaussianity, pairwise dependencies) that limit computational complexity and/or grant analytic tractability~\cite{barberBRML2012}. This may distort statistical inference in uncontrollable ways, especially in the under-sampling regime, by, for instance, hiding effects from high-order interactions (see e.g.~\cite{riechers2021fraudulent}). The concept of relevance may allow to develop methods of featureless inference where no assumption on the generative model or on the structure of dependencies is used.}

\add{We believe that an understanding of learning in the under-sampling regime} cannot be based on assumptions on the data, on the task or on the models. Rather, it should rely on principles of information theory. In this spirit, we propose the maximisation of the relevance as a general principle for statistical learning.

This \add{principle} offers precise guidelines for the design of efficient learning machines. In models such as RBM training involves both the internal representation $p(\bs)$ and the output layer $p(\vec x|\bs)$ that projects the internal state on the output $\vec x$. Typically the machine is initialised in a state where $p(\bs)$ is very far from a distribution of maximal relevance. This appears inefficient and suggests that \blue{other schemes may be more efficient.} One such scheme is where the internal model $p(\bs)$ is fixed at the outset as a distribution of maximal relevance at a preassigned resolution, and \blue{learning is, instead, focuses on the way this maximal relevance distribution interacts with the distribution of the data in the output layer.} Architectures where the internal representation is fixed are already used in Extreme Learning Machines~\cite{kasun2013representational} recurrent neural networks and in reservoir computing (see~\cite{principe2015universal} for a review), and they grant considerable speed-up in training because only the output layer needs to be learned. 
This suggests that learning can count on a huge degeneracy of models from which to draw the internal representation of a dataset. 
\add{There is indeed mounting evidence~\cite{mei2019generalization,zecchina} that the success of deep learning machines is due to the fact that massive over-parametrisation endows their energy landscape of a dense set of ``flat'' optima that are easily accessible to algorithms~\cite{zecchina}. In addition, advances in transfer learning~\cite{pan2009survey} show that the internal representation $p(\bs)$ learned on a particular task can be used to learn a different dataset, showing that the internal representation enjoys a certain degree of task independence.} 

\add{To what extent $p(\bs)$ can be independent of the data \blue{used for learning and the task to be learnt} is a key question. Wolpert~\cite{wolpert2021important} argues that the success of learning algorithms is ultimately due to the fact that the tasks are drawn from a very biased distribution. Indeed, otherwise the no-free-lunch theorems on learning~\cite{wolpert1997no} would imply that an algorithm that performs well on a task has no guarantee to performs better than any other algorithm on a randomly chosen task. We speculate that the distribution over tasks alluded to in~\cite{wolpert2021important}
should be skewed towards problems where the data is intrinsically relevant. Whether maximal relevance is enough to identify such distribution, and hence the internal representation $p(\bs)$, or not is an interesting open question.}

\add{An approach where $p(\bs)$ is fixed at the outset} seems rather natural and it would have several further advantages. First the possibility to devise fast learning algorithms as in architectures where only the output layer is learned~\cite{principe2015universal}. Second, in an approach of this type the resolution $H[\bs]$ can be fixed at the outset and does need to be tuned by varying the number of hidden nodes or by resorting to {\em ad-hoc} regularisation schemes. In these machines the resolution could be varied continuously by exploring large deviations of $H[\bs]$, as discussed in Section~\ref{sec:OLMprop}. 
For example, the optimal tradeoff between resolution and relevance can be used to tune learning machines to optimal generative performance, as claimed in~\cite{SMJ}. 
Furthermore, different data could be learned on the same ``universal'' internal representation
$p(\bs)$\add{, thereby allowing to establish relations between different datasets\footnote{\add{If  the output layers $p(\vec x|\bs)$ and $p(\vec y|\bs)$ are learned with the same representation $p(\bs)$ for two different datasets $\vec x$ and $\vec y$, it is possible to compute a joint distribution $p(\vec x,\vec y)=\sum_{\bs}p(\vec x|\bs)p(\vec y|\bs)p(\bs)$ by marginalising over the hidden layer $\bs$. The associations obtained in these way are reminiscent of the phenomenon of synesthesia~\cite{ward2013synesthesia} in neuroscience, whereby a perceptual stimulus (a letter or a taste) evokes the experience of a concurrent percept (color or shape).}}}.

\add{We conjecture that designs inspired by the principle of maximal relevance might also be efficient in the sense of requiring minimal thermodynamic costs of learning. The last decade has witnessed remarkable advances in stochastic thermodynamics and its relation to information processing~\cite{parrondo2015thermodynamics}, learning~\cite{goldt2019stochastic} and computing~\cite{wolpertk}. These have shown that information processing requires driving systems out of equilibrium, and the efficiency with which it can be done is limited by fundamental bounds. There is mounting evidence that thermodynamic efficiency coincides with optimal information processing. For example, Boyd et al.~\cite{boyd} show that models  that provide the most accurate description of the environment also allow for maximal work extraction when used as information engines. Touzo {\em et al.}~\cite{touzo2020optimal} have shown that, in a cyclic engine that exploits a measurement, maximal work extraction coincides with optimal coding of the measured quantity, and the protocol that achieves this requires driving the engine to a non-equilibrium critical state of maximal relevance. 
Part of the thermodynamic cost of learning comes from adjusting the internal energy levels of the machine. These costs are saved if the internal state distribution $p(\bs)$ is fixed throughout, as in the architectures described above. Taken together, these pieces of evidence hint at the conjecture that optimal learning machines which learn from maximally informative samples should operate at minimal thermodynamic costs.}

\add{Under the assumption that evolution should promote efficient architectures of information processing that impose a lower burden on the energy budget of an organism, \blue{the conjectures mentioned above} would also be of relevance for understanding living systems. Even the most \blue{routine tasks performed by a living system}, such as searching for food, involves a host of statistical problems. 
Each of these problems likely needs to be solved in a regime where the data is barely sufficient to reach a conclusion \blue{with high certainty}, because a timely response may be far more important than a highly statistically significant one. 
This suggests that living systems have likely evolved to solve statistical problems in the under-sampling regime.} 
The hypothesis that the basis of statistical learning in living systems relies on an efficient representation with maximal relevance, provides a guideline for statistical inference in high-dimensional data in biology. \add{Indeed, the concept of relevance allows us to identify what are meaningful features in high-dimensional datasets, without the need of knowing {\em a priori} what they are meaningful for. Hence,} the precise characterisation of most informative samples given by the concept of relevance can be exploited to devise methods for extracting relevant variables in high-dimensional data. Along this line, Grigolon {\em et al.}~\cite{Grigolon} have used the concept of relevance for identifying relevant residues in protein sequences and Cubero {\em et al.}~\cite{RyanMSR} have applied the same concept to distinguish informative neurons responsible for spatial cognition in rats from uninformative ones (see Section~\ref{sec:neuro}). While these first attempts are encouraging, more research is needed to fully exploit the insights that a relevance based approach can unveil. 

\section*{\add{Acknowledgments}}

\add{We're greatly indebted to our collaborators, Ryan Cubero, Odilon Duranthon, Ariel Haimovici, Silvio Franz, Silvia Grigolon,  Jungyo Jo, Iacopo Mastromatteo, Juyong Song and Rongrong (Emily) Xie with whom parts of these ideas have been developed. We also acknowledge very useful discussions with Vijay Balasubramanian, Jean Barbier, Claudia Battistin, Antonio Celani, Benjamin Dunn, Saptarshi Ghosh, Peter Latham, Marc Mezard, Thierry Mora, Kamiar Rahnama Rad, Guido Sanguinetti, Miguel Virasoro, and Riccardo Zecchina} \blue{Y. R. has been financially supported by funding from the Kavli Foundation and from the Norwegian Research Council, Centre of Excellence scheme (Centre for Neural Computation, grant number 223262). }

\appendix

\section{Appendix: Parametric inference}
\label{app:param}

In order to derive Eq.~(\ref{eq:MI}), let us start by recalling that the posterior $p(\theta|\hat \bs)$ is given by Bayes rule
\begin{equation}
\label{eqpost}
p(\theta|\hat\bs)=\frac{f(\hat \bs|\theta)p_0(\theta)}{p(\hat\bs)}
\end{equation}
where, recalling that $\hat\bs$ are $N$ independent draws $\bs^{(i)}$ from $f(\theta|\bs)$, 
\begin{equation}
f(\hat \bs|\theta)=\prod_{i=1}^N f(\bs^{(i)}|\theta)
\end{equation}
is the likelihood, and 
\begin{equation}
\label{eqevid}
p(\hat \bs)=\int d\theta f(\hat \bs|\theta)p_0(\theta)
\end{equation}
is the evidence. For $N\to\infty$ integrals on $\theta$ in Eqs.(\ref{eqevid},\ref{eqdkl}) are dominated by the region where $\theta\approx\hat\theta$ where
\begin{equation}
\hat\theta={\rm arg}\max_{\theta}f(\hat \bs|\theta)
\end{equation}
is the maximum likelihood estimator of the parameters. This depends on the data $\hat \bs$ but we shall omit this dependence to keep the notation as light as possible. To leading order, the log-likelihood can be approximated by power expansion around $\hat\theta$, as
\begin{equation}
\label{eqsaddle}
\log f(\hat \bs|\theta)= \log f(\hat \bs|\hat\theta)-\frac{N}{2}\sum_{i,j=1}^d(\theta_i-\hat\theta_i)L_{i,j}(\hat\theta)(\theta_j-\hat\theta_j)+\ldots
\end{equation}
where $\ldots$ corresponds to sub-leading terms, and
\begin{equation}
L_{i,j}(\theta)=-\frac{\partial^2}{\partial\theta_i\partial\theta_j}\frac 1 N \sum_{i=1}^N \log f(\bs^{(i)}|\theta)
\end{equation}
is minus the Hessian of the log-likelihood \add{per data point}. Since the expression above is an average over a sample of $N$ independent draws, it converges under very general conditions, to a finite value. In addition, since $\hat\theta$ is a maximum, the matrix $\hat L(\hat\theta)$ is positive definite. Using the approximation~(\ref{eqsaddle}), Eq.~(\ref{eqevid}) can be evaluated, to leading order, as a gaussian integral 
\begin{equation}
\label{approxevid}
p(\hat \bs)\simeq \left(\frac{2\pi}{N}\right)^{d/2}\frac{p_0(\hat\theta)}{\sqrt{{\rm det}\hat L(\hat\theta)}}f(\hat\bs|\hat\theta).
\end{equation}
Combining this with Eq.~(\ref{eqpost}), we find that the posterior is, to leading order, a Gaussian distribution
\begin{equation}
p(\theta|\hat\bs)\simeq \left(\frac{N}{2\pi}\right)^{d/2}\sqrt{{\rm det}\,\hat L(\hat\theta)}e^{-\frac{N}{2}\sum_{i,j} 
(\theta_i-\hat\theta_i)L_{i,j}(\hat\theta)(\theta_j-\hat\theta_j)}\,
\end{equation}
Inserting this in the expression for the Kullback-Leibler divergence, leads to 
\begin{eqnarray}
D_{KL}\left[p(\theta|\hat\bs)||p_0(\theta)\right] & = &  \int d\theta p(\theta|\hat\bs)\log\frac{p(\theta|\hat\bs)}{p_0(\theta)} \\
 & \simeq & \frac d 2 \log\frac{N}{2\pi} +\frac 1 2 \log {\rm det}\,\hat L(\hat\theta)-\log p_0(\hat\theta) \\
 & & -\frac{N}{2}
 \int d\theta p(\theta|\hat\bs)\sum_{i,j} 
(\theta_i-\hat\theta_i)L_{i,j}(\hat\theta)(\theta_j-\hat\theta_j).\label{eqdkl}
\end{eqnarray}
Within the same Gaussian approximation, the last integral in Eq.~(\ref{eqevid}) can be easily computed, with the result that the last term in Eq.~(\ref{eqevid}) equals $d/2$. This yields Eq.~(\ref{eq:MI}).

As for Eq.~(\ref{eq:BMS}), the evidence of model $f$ coincides with $p(\hat\bs)$. Hence Eq.~(\ref{approxevid}) leads directly to Eq.~(\ref{eq:BMS}).



\bibliographystyle{elsarticle-num} 
\bibliography{RelevanceReview.bib}





\end{document}